\documentclass[lettersize,journal]{IEEEtran} 
\pdfoutput=1
\usepackage{amsmath,amsfonts}
\usepackage{algorithmic}
\usepackage{algorithm}
\usepackage{array}
\usepackage[caption=false,font=normalsize,labelfont=sf,textfont=sf]{subfig}
\usepackage{textcomp}
\usepackage{stfloats}
\usepackage{url}
\usepackage{verbatim}
\usepackage{graphicx}
\usepackage{cite}

\usepackage{hyperref}
\usepackage{booktabs}       
\usepackage{nicefrac}       
\usepackage{microtype}      
\usepackage{xcolor}         
\usepackage{graphicx}
\usepackage{wrapfig,lipsum,booktabs}
\usepackage{multirow}
\usepackage{xcolor}         

\newtheorem{definition}{Definition}
\renewcommand{\eqref}[1]{Eq.~(\textup{\ref{#1}})}
\usepackage{bm}
\usepackage{cleveref}
\crefname{table}{Table}{Tables}
\crefname{section}{Section}{Sections}
\crefname{appendix}{Appendix}{Appendixes}
\crefname{algorithm}{Algorithm}{Algorithms}
\crefname{figure}{Fig.}{Figs.}
\crefname{theorem}{Theorem}{}
\crefname{lemma}{Lemma}{}
\crefname{definition}{Definition}{}

\begin{document}

\title{Noise May Contain Transferable Knowledge: Understanding Semi-supervised Heterogeneous Domain Adaptation from an Empirical Perspective}

\author{Yuan Yao, Xiaopu Zhang, Yu Zhang,~\IEEEmembership{Member,~IEEE}, Jian Jin, and Qiang Yang,~\IEEEmembership{Fellow,~IEEE}

\thanks{Yuan Yao is with the Beijing Teleinfo Technology Company Ltd., China Academy of Information and Communications Technology, Beijing 100095, China. (e-mail: yaoyuan.hitsz@gmail.com)}

\thanks{Xiaopu Zhang is with the Department of Research and Development, Inspur Computer Technology Co., Ltd., Beijing 100095, China (e-mail: zhangxiaopu@inspur.com)} 

\thanks{Yu Zhang is with the Department of Computer Science and Engineering, Southern University of Science and Technology, Shenzhen 518055, China. (e-mail: yu.zhang.ust@gmail.com)}

\thanks{Jian Jin is with the Research Institute of Industrial Internet of Things, China Academy of Information and Communications Technology, Beijing 100095, China. (e-mail: jin.jian@caict.ac.cn)}


\thanks{Qiang Yang is with the Department of Computer Science and Engineering, Hong Kong University of Science and Technology, Hong Kong, and also with WeBank, Shenzhen 518052, China. (e-mail: qyang@cse.ust.hk)}

\thanks{
Corresponding authors: Yu Zhang and Jian Jin.}
}



\maketitle

\begin{abstract}

Semi-supervised heterogeneous domain adaptation (SHDA) addresses learning across domains with distinct feature representations and distributions, where source samples are labeled while most target samples are unlabeled, with only a small fraction labeled. Moreover, there is no one-to-one correspondence between source and target samples. Although various SHDA methods have been developed to tackle this problem, the nature of the knowledge transferred across heterogeneous domains remains unclear. This paper delves into this question from an empirical perspective. We conduct extensive experiments on about 330 SHDA tasks, employing two supervised learning methods and seven representative SHDA methods. Surprisingly, our observations indicate that both the category and feature information of source samples do not significantly impact the performance of the target domain. Additionally, noise drawn from simple distributions, when used as source samples, may contain transferable knowledge. Based on this insight, we perform a series of experiments to uncover the underlying principles of transferable knowledge in SHDA. Specifically, we design a unified Knowledge Transfer Framework (KTF) for SHDA. 
Based on the KTF, we find that the transferable knowledge in SHDA primarily stems from the transferability and discriminability of the source domain. Consequently, ensuring those properties in source samples, regardless of their origin (\textit{e.g.}, image, text, noise), can enhance the effectiveness of knowledge transfer in SHDA tasks. The codes and datasets are available at \href{https://github.com/yyyaoyuan/SHDA}{https://github.com/yyyaoyuan/SHDA}.
\end{abstract}


\section{Introduction}

In recent years, supervised learning techniques have undergone significant advancements with sufficient high-quality labeled samples \cite{Krizhevsky2012Imagenet,He2016Deep,Vaswani2017Attention,Lecun2015Deep}. In practice, however, it is often prohibitive to collect abundant high-quality labeled samples due to concerns about privacy, confidentiality, copyright, \textit{etc}. To mitigate this challenge, 
domain adaptation (DA) methods \cite{Pan2010A,Csurka-2017A,Zhuang2020A,Yang2020Transfer} have been proposed to improve the learning performance in a label-insufficient target domain by drawing upon knowledge from a related label-sufficient source domain. Those methods have achieved remarkable progress in various practical applications \cite{Zhang2019A,Ge2020Mutual,Peng2020Federated,Liu2023Discovering,Hoyer2023MIC,Oza2023Unsupervised}. In general, most existing DA methods \cite{Long2013Adaptation,Long2016Deep,Long2018Transferable,Xu2022CDTrans,Chen2019Transferability,Zhang2019Bridging,Rangwani2022A} assume that the original feature representation of source samples is identical to that of target ones. Accordingly, they cannot be directly utilized to handle the \textit{heterogeneous} scenarios, where source and target samples are characterized by distinct feature representations. However, those heterogeneous scenarios are common in many practical applications \cite{Day2017A,Bao2023A}, such as cross-modal image recognition \cite{Yao2019Heterogeneous,Fang2023Semi-Supervised} and cross-lingual text categorization \cite{Zhou2019Multi,Yao2020Discriminative,Wang2020Prototype-Matching}.

\begin{figure}
\centering
{\includegraphics[width=0.98\columnwidth]{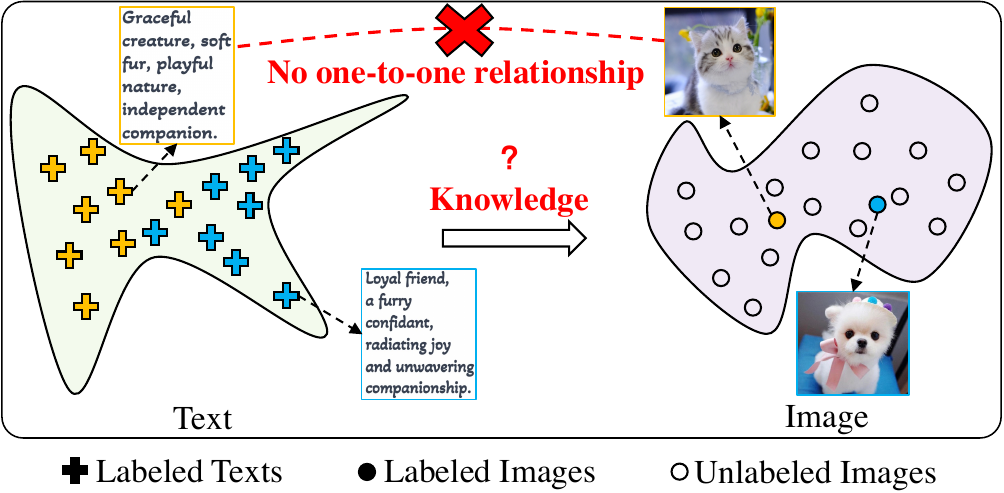}}
\caption{Example scenario of SHDA with a textual source domain and a visual target domain. Here, all texts are labeled, but most images remain unlabeled, with only a small number having labels. Also, there is no one-to-one relationship between texts and images. We do not know what knowledge is transferred across heterogeneous domains.}
\label{fig:Question_unique}
\vspace{-4ex}
\end{figure}

To tackle those scenarios, researchers have formulated an important but challenging problem, \textit{i.e.}, \textit{semi-supervised heterogeneous domain adaptation} (SHDA) \cite{Day2017A,Bao2023A}. 
As illustrated in \cref{fig:Question_unique}, under the SHDA setting, source and target samples originate from different feature spaces, such as text and image. Also, source samples are labeled, while the target domain has limited labeled samples and a substantial amount of unlabeled ones. In addition, there is no one-to-one correspondence, \textit{i.e.}, pair information, between source and target samples. Numerous SHDA methods have been developed \cite{Yao2019Heterogeneous,Li2020Simultaneous,Wang2020Prototype-Matching,Gu2022Keypoint-Guided,Fang2023Semi-Supervised}, resulting in improved transfer performance across heterogeneous domains. 
Since samples from the two domains could be very dissimilar due to the heterogeneous feature spaces, we pose a question: ``\textit{What is the transferable knowledge in SHDA}?'' This is an essential issue of SHDA, and however, it has not been well-explored.

\begin{figure}
\centering
{\includegraphics[width=0.98\columnwidth]{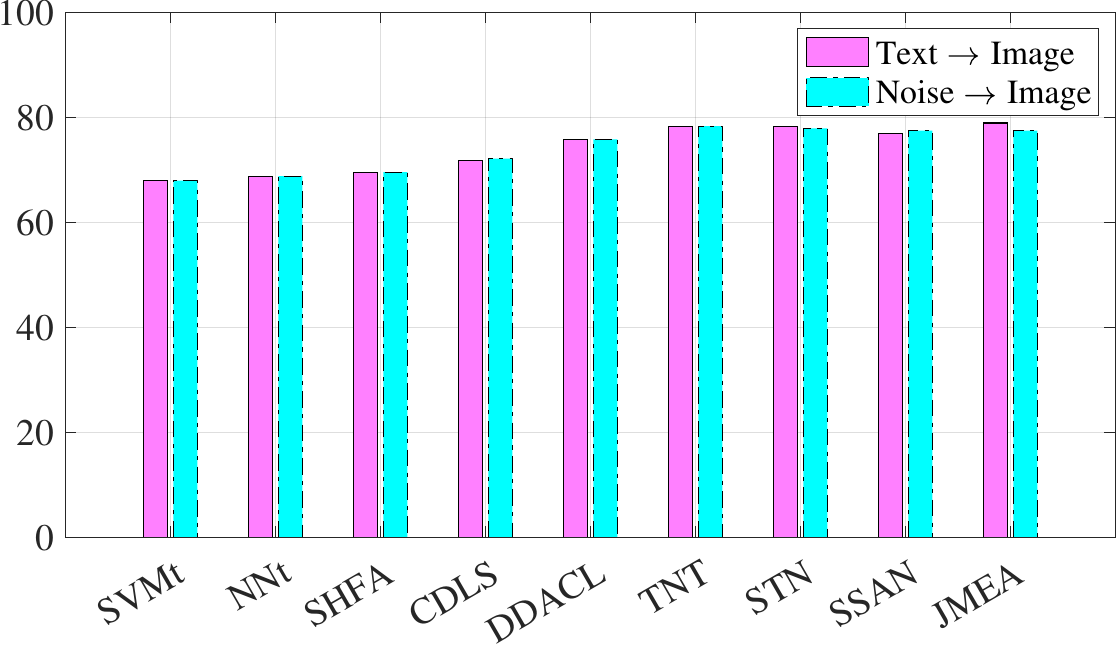}}
\caption{Experimental results on the NUS-WIDE+ImageNet-8 dataset \cite{Chua2009NUS-WIDE,Deng2009ImageNet}, which demonstrates that noise may contain transferable knowledge. Here, Text $\rightarrow$ Image is a vanilla SHDA task, whilst Noise $\rightarrow$ Image is a specialized SHDA task with pure noise as the source sample. In addition, SVMt and NNt are two supervised learning methods, whereas SHFA, CDLS, DDACL, TNT, STN, SSAN, and JMEA are seven SHDA methods.}
\label{fig:Introduction}
\vspace{-4ex}
\end{figure}

To explore the above problem in depth, we perform comprehensive experiments across nearly 330 SHDA tasks using two supervised learning methods and seven typical SHDA methods, including SVMt \cite{Chang2011LIBSVM}, NNt \cite{Abadi2016TensorFlow}, SHFA \cite{Li2014Learning}, CDLS \cite{Tsai2016Learning}, DDACL \cite{Yao2020Discriminative}, TNT \cite{Chen2016Transfer}, STN \cite{Yao2019Heterogeneous}, SSAN \cite{Li2020Simultaneous}, and JMEA \cite{Fang2023Semi-Supervised}.
Specifically, we first investigate how the category and feature information of source samples influence the performance of the target domain. To our surprise, this seemingly significant information is not dominant in affecting the performance of the target domain. Accordingly, we first hypothesize that noise drawn from simple distributions, \textit{e.g.}, Gaussian distribution, as source samples may contain transferable knowledge. Then, we conduct extensive experiments to verify this hypothesis, exemplified by the results on the NUS-WIDE+ImageNet-8 dataset \cite{Chua2009NUS-WIDE,Deng2009ImageNet} shown in \cref{fig:Introduction}. Here, Text $\rightarrow$ Image is a vanilla SHDA task, while Noise $\rightarrow$ Image is a specialized SHDA task with pure noise as source samples. Our findings reveal that all the methods demonstrate comparable performance on both tasks. Based on this observation, we empirically confirm through extensive experiments that noise may indeed contain transferable knowledge, which can thus be utilized as source samples to improve the performance of the target domain.

Building on the pivotal observation above, we synthesize various noise domains to conduct a series of experiments aimed at uncovering the mystery of transferable knowledge in SHDA. Concretely, we first develop a unified Knowledge Transfer Framework (KTF) and then perform large-scale experiments by creating various noise domains. Based on the KTF, we analyze the correlation between the transferability/discriminability of source samples and the performance improvement ratio in the target domain \cite{Wei2018Transfer}. \textit{As a result, we find that the core of transferable knowledge mainly lies in the transferability and discriminability of the source domain}. Consequently, regardless of the origin of source samples (\textit{e.g.}, image, text, and noise), maintaining their transferability and discriminability is crucial for ensuring effective knowledge transfer in SHDA tasks. 

We highlight the contributions of this paper as follows. 
\begin{itemize}
\item To the best of our knowledge, we are the first to empirically investigate the transferable knowledge in SHDA. 
\item We observe that noise drawn from simple distributions as source samples may contain transferable knowledge, which has the potential to inspire more intriguing research. 
\item Our observations indicate that the essence of transferable knowledge in SHDA primarily lies in the transferability and discriminability of the source domain, regardless of its origin (\textit{e.g.}, image, text, and noise).
\item We open-source the codes and datasets used in this paper at \href{https://github.com/yyyaoyuan/SHDA}{https://github.com/yyyaoyuan/SHDA}, including seven typical SHDA methods and several popular datasets, which, to our humble knowledge, is the first relatively comprehensive SHDA open-source repository.
\end{itemize}

The remaining parts of this paper are organized as follows. In Section \ref{section:overview}, we first provide an overview of SHDA. Then, Section \ref{section:setup} offers the detailed experimental setups. Next, we perform extensive experiments in Sections \ref{section:smss}-\ref{section:smsn} to explore the transferable knowledge in SHDA.
Subsequently, in Section \ref{section:discussion}, we present several insightful discussions. Finally, we make conclusions in Section \ref{section:conclusion}.

\section{Overview}
\label{section:overview}

In this section, we begin by defining SHDA, followed by a concise review. Finally, we summarize the pipeline of SHDA.

\begin{table}[htpb]
  \centering
  \caption{Notations.}
    \begin{tabular}{ll}
    \toprule
    Notation & Description \\
    \midrule
    $\mathcal{X}_s$ / $\mathcal{X}_t$ & Source/Target feature space \\
    $\mathcal{D}_s$ / $\mathcal{D}_t$ & Source/Target domain \\
    $\mathcal{D}_l$ / $\mathcal{D}_u$ & Labeled/Unlabeled target domain \\
    $\mathbf{x}_i^s$ / $\mathbf{x}_i^l$ / $\mathbf{x}_i^u$ & the $i$-th sample in $\mathcal{D}_s$ / $\mathcal{D}_l$ / $\mathcal{D}_u$ \\
    $\mathbf{y}_i^s$ / $\mathbf{y}_i^l$ & One-hot label of $\mathbf{x}_i^s$ / $\mathbf{x}_i^l$\\
    $n_s$ / $n_l$ / $n_u$ & Number of samples in $\mathcal{D}_s$ / $\mathcal{D}_l$ / $\mathcal{D}_u$ \\
    $C$     & Number of categories \\
    \bottomrule
    \end{tabular}%
  \label{tab:notations}%
  \vspace{-4ex}
\end{table}%

\subsection{Notations and Definition}

Let $\mathcal{X}_s \subset \mathbb{R}^{d_s}$ and $\mathcal{X}_t \subset \mathbb{R}^{d_t}$ be the source and target feature spaces, respectively.
The source domain is denoted by $\mathcal{D}_s=\{(\mathbf{x}^s_i, \mathbf{y}^s_i)\}_{i=1}^{n_s}$, where $\mathbf{x}^s_i \in \mathcal{X}_s$ is the $i$-th source sample, and $\mathbf{y}^s_i$ is its corresponding one-hot label over $C$ categories. Similarly, we denote the target domain by $\mathcal{D}_t = \mathcal{D}_l \cup \mathcal{D}_u = \{(\mathbf{x}^l_i, \mathbf{y}^l_i)\}_{i=1}^{n_l} \cup \{\mathbf{x}^u_i\}_{i=1}^{n_u}$, where $\mathbf{x}^l_i$ $(\mathbf{x}^u_i) \in \mathcal{X}_t$ is the $i$-th labeled (unlabeled) target sample, and $\mathbf{y}^l_i$ is its associated one-hot label for $\mathbf{x}^l_i$ among the $C$ categories. 
Based on those notations as summarized in \cref{tab:notations}, the SHDA task is defined as follows. 

\begin{definition} (SHDA). Under the SHDA setting, a source domain $\mathcal{D}_s$ and a target domain $\mathcal{D}_t$ are given, with samples drawn from distinct distributions. Also, source and target samples share the same categories, but there is no one-to-one correspondence between them. Moreover, $\mathcal{X}_s \neq \mathcal{X}_t$, $n_s \gg n_l$, and $n_u \gg n_l$. The goal is to train a high-quality model using samples from both $\mathcal{D}_s$ and $\mathcal{D}_t$ and then apply the trained model to classify samples in $\mathcal{D}_u$.
\end{definition}

\subsection{Overview}

Existing SHDA methods can be roughly categorized into two approaches, \textit{i.e.}, the shallow projection approach and deep projection approach. 
In the following, we provide a review for those two approaches.

\subsubsection{Shallow Projection Approach}

Most existing SHDA methods fall into this approach, primarily utilizing the classifier adaptation and distribution alignment mechanisms for domain adaptation. Specifically, HFA \cite{Duan2012Learning}, SHFA \cite{Li2014Learning}, and MMDT \cite{Hoffman2013Efficient,Hoffman2014Asymmetric} employ the classifier adaptation mechanism, which uses all samples from both domains to learn a domain-shared classifier, aligning the discriminative structures of both domains. For instance, MMDT projects target samples into the source domain by training a domain-shared support vector machine on labeled cross-domain samples. HFA and SHFA first augment the projected source and target samples with the original features and then learn a support vector machine shared between domains. LS-UP \cite{Tsai2016Heterogeneous}, PA \cite{Li2018Heterogeneous}, SGW \cite{Yan2018Semi-Supervised}, and KPG \cite{Gu2022Keypoint-Guided} adopt the distribution alignment mechanism, which learns optimal projections by reducing the distributional divergence between domains. For example, PA first learns a common subspace by dictionary-sharing coding and then alleviates the distributional divergence between domains. Recently, KPG regards labeled cross-domain samples as key samples to guide the correct matching in optimal transport. SCP-ECOC \cite{Xiao2015Semi-supervised}, SDASL \cite{Yao2015Semi-supervised}, G-JDA \cite{Hsieh2016Recognizing}, CDLS \cite{Tsai2016Learning}, SSKMDA \cite{Xiao2015Feature}, DDACL \cite{Yao2020Discriminative}, and KHDA \cite{Fang2023Semi-Supervised} take into account both the classifier adaptation and distribution alignment strategies. For example, G-JDA and CDLS perform the distribution alignment and classifier adaptation strategies in an iterative manner, and DDACL learns a domain-shared classifier by both reducing the distributional discrepancy and enlarging the prediction discriminability.

\subsubsection{Deep Projection Approach}

With the advancement of deep learning techniques \cite{Lecun2015Deep}, some studies have utilized them to address the SHDA problem.
Specifically, DTN \cite{Shu2015Weakly-Shared} reduces the divergence of the parameters in the last layers between the source and target projection networks. TNT \cite{Chen2016Transfer} simultaneously considers feature projection, sample categorization, and domain adaptation with deep neural networks. Deep-MCA \cite{Li2019Heterogeneous} utilizes a deep neural network to complete the heterogeneous feature matrix and find a better measure function for distribution alignment across domains. STN \cite{Yao2019Heterogeneous} adopts the soft-labels of unlabeled target samples to align the conditional distributions between domains, and builds non-linear source and target projection networks. SSAN \cite{Li2020Simultaneous} considers the implicit semantic correlation and explicit semantic alignment mechanisms in a heterogeneous transfer network. PMGN \cite{Wang2020Prototype-Matching} constructs an end-to-end graph prototypical network to learn the domain-invariant category prototype representations, which not only mitigates the distributional divergence but also enhances the prediction discriminability. Recently, JMEA \cite{Fang2023Semi-Supervised} jointly trains a transferable classifier and a semi-supervised classifier to screen high-confidence pseudo-labels for unlabeled target samples.

\begin{figure}[!tbph]
\centering
{\includegraphics[width=0.98\columnwidth]{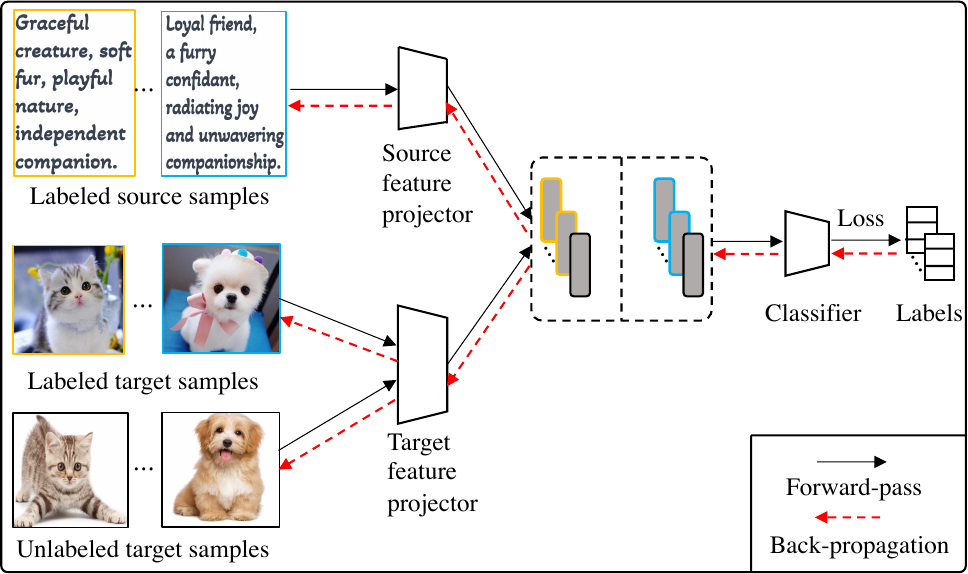}}
\caption{In general, the SHDA pipeline integrates the classification adaptation and distribution alignment mechanisms to jointly learn the source and target feature projectors, along with the classifier, from scratch in a semi-supervised manner. Notably, the feature projectors are unique to each domain.}
\label{fig:Pipeline}
\vspace{-4ex}
\end{figure}

\subsection{Pipeline}

In summary, most SHDA methods generally follow the pipeline illustrated in \cref{fig:Pipeline}. Specifically, they employ classification adaptation and distribution alignment mechanisms to jointly learn the source and target feature projectors, along with the classifier, from scratch in a semi-supervised fashion.
Note that the feature projectors are unique to each domain. 
Overall, \textit{SHDA methods excel at adapting to samples originating from distinct feature spaces, enabling effective domain adaptation even when the source and target domains differ significantly in their feature representations}. This characteristic enhances their generality compared to \textit{homogeneous} domain adaptation methods \cite{Long2013Adaptation, Long2016Deep, Long2018Transferable, Xu2022CDTrans, Chen2019Transferability, Zhang2019Bridging, Rangwani2022A}, while also prompting our interest in investigating the fundamental issues of SHDA.

\section{Setups for Empirical Studies}
\label{section:setup}

In this section, we introduce the experimental setup in detail, including datasets, baselines, and the evaluation metric.

\subsection{Datasets}

Following \cite{Yao2019Heterogeneous,Li2020Simultaneous,Wang2020Prototype-Matching}, we adopt three widely-used SHDA datasets, including \textbf{Office+Caltech-10} \cite{Saenko2010Adapting,Griffin2007Caltech-256}, \textbf{Multilingual Reuters Collection} \cite{Amini2009Learning}, and \textbf{NUS-WIDE+ImageNet-8} \cite{Chua2009NUS-WIDE,Deng2009ImageNet}. 

The \textbf{Office+Caltech-10} dataset comprises two sub-datasets, \textit{i.e.}, Office \cite{Saenko2010Adapting}, and Caltech-256 \cite{Griffin2007Caltech-256}. The Office dataset includes 4,652 images across 31 classes collected from three different domains: Amazon (\textbf{A}), Webcam (\textbf{W}), and DSLR (\textbf{D}). The Caltech-256 (\textbf{C}) dataset consists of 30,607 images of 256 objects. We select 10 shared categories from those two datasets to create the Office+Caltech-10 dataset. In addition, we represent each image using two kinds of features: 800-dimensional $SURF$ ($S_{800}$) \cite{Bay2006Surf} and 4096-dimensional $DeCAF_6$ ($D_{4096}$) \cite{Donahue2014DeCAF}. In the following experiments, we designate \textbf{A}, \textbf{C}, and \textbf{W} as source domains and \textbf{C}, \textbf{W}, and \textbf{D} as target ones.
For the source domain, we treat all images as labeled samples. As for the target domain, we randomly choose three images in each category as labeled samples, and the remaining images are regarded as unlabeled samples.

The \textbf{Multilingual Reuters Collection} dataset \cite{Amini2009Learning} includes 111,740 articles, which are classified into six categories and written in five distinct languages: English (\textbf{E}), French (\textbf{F}), German (\textbf{G}), Italian (\textbf{I}), and Spanish (\textbf{S}). We utilize the bag-of-words representation with term frequency-inverse document frequency features to represent each article. Subsequently, by following \cite{Tsai2016Learning,Yao2019Heterogeneous,Yao2020Discriminative,Li2020Simultaneous}, the principal component analysis \cite{jolliffe2016principal} is performed to reduce the dimensionalities of features to 1131, 1230, 1417, 1041, and 807 for \textbf{E}, \textbf{F}, \textbf{G}, \textbf{I}, and \textbf{S}, respectively. In subsequent experiments, we designate the \textbf{S} domain as the target domain, while the remaining domains are treated as the source domains. For the source domain, we randomly select 100 articles per category as labeled samples. As for the target domain, we randomly pick up five and 500 articles in each category as labeled and unlabeled samples, respectively.

\begin{table*}[t]
  \centering
  \tabcolsep=0.14cm
  \caption{Baselines utilized in the paper.}
    \begin{tabular}{llll}
    \toprule
    Method & Type  & URL for Code & Publication \\
    \midrule
    SVMt \cite{Chang2011LIBSVM}  & Supervised Learning & \url{https://www.csie.ntu.edu.tw/~cjlin/libsvm/}
    & ACM TIST 2011 \\
    NNt \cite{Abadi2016TensorFlow}  & Supervised Learning & \url{https://github.com/tensorflow/tensorflow}     & OSDI 2016 \\
    SHFA \cite{Li2014Learning}  & Shallow Projection SHDA & \href{https://github.com/wenli-vision/SHFA\_release}{https://github.com/wenli-vision/SHFA\_release} & TPAMI 2014 \\
    CDLS \cite{Tsai2016Learning}  & Shallow Projection SHDA & \href{https://github.com/yaohungt/Cross-Domain-Landmarks-Selection-CDLS-/tree/master}{https://github.com/yaohungt/Cross-Domain-Landmarks-Selection-CDLS-/tree/master} & CVPR 2016 \\
    DDACL \cite{Yao2020Discriminative} & Shallow Projection SHDA & \href{https://github.com/yyyaoyuan/DDA}{https://github.com/yyyaoyuan/DDA} & Pattern Recognition 2020 \\
    TNT \cite{Chen2016Transfer}   & Deep Projection SHDA & \href{https://github.com/wyharveychen/TransferNeuralTrees}{https://github.com/wyharveychen/TransferNeuralTrees} & ECCV 2016 \\
    STN \cite{Yao2019Heterogeneous}   & Deep Projection SHDA & \href{https://github.com/yyyaoyuan/STN}{https://github.com/yyyaoyuan/STN} & ACM MM 2019 \\
    SSAN \cite{Li2020Simultaneous}  & Deep Projection SHDA & \href{https://github.com/BIT-DA/SSAN}{https://github.com/BIT-DA/SSAN} & ACM MM 2020 \\
    JMEA \cite{Fang2023Semi-Supervised}  & Deep Projection SHDA & \href{https://github.com/fang-zhen/Semi-supervised-Heterogeneous-Domain-Adaptation}{https://github.com/fang-zhen/Semi-supervised-Heterogeneous-Domain-Adaptation} & TPAMI 2023 \\
    \bottomrule
    \end{tabular}%
  \label{tab:baselines}%
  \vspace{-4ex}
\end{table*}%

The \textbf{NUS-WIDE+ImageNet-8} dataset contains the NUS-WIDE \cite{Chua2009NUS-WIDE} and ImageNet \cite{Deng2009ImageNet} datasets. The former comprises 269,648 images along with their corresponding tags from Flickr, and the latter consists of 3.2 million images and 5,247 synsets. By following \cite{Chen2016Transfer,Yao2019Heterogeneous,Li2020Simultaneous}, we select eight overlapping categories from those two datasets to build the NUS-WIDE+ImageNet-8 dataset. Also, we utilize the tag from NUS-WIDE and the image from ImageNet as the \textbf{Text} and \textbf{Image} domains, respectively. Furthermore, in line with \cite{Chen2016Transfer,Yao2019Heterogeneous,Li2020Simultaneous}, we adopt a five-layer neural network to extract the 64-dimensional features for representing texts from the \textbf{Text} domain. Also, we employ the $D_{4096}$ features to characterize the images from the \textbf{Image} domain. In the following experiments, we randomly sample 100 texts from each category within the \textbf{Text} domain to serve as labeled source samples. From the \textbf{Image} domain, three images per category are randomly selected as labeled target samples, whereas the remaining images are treated as unlabeled target samples.

\subsection{Baselines}

In the experiments, we utilize nine baselines, including SVMt \cite{Chang2011LIBSVM}, NNt \cite{Abadi2016TensorFlow}, SHFA \cite{Li2014Learning}, CDLS \cite{Tsai2016Learning}, DDACL \cite{Yao2020Discriminative}, TNT \cite{Chen2016Transfer}, STN \cite{Yao2019Heterogeneous}, SSAN \cite{Li2020Simultaneous}, and JMEA \cite{Fang2023Semi-Supervised}. 
Here, SVMt and NNt are two supervised learning methods, while SHFA, CDLS, DDACL, TNT, STN, SSAN, and JMEA are SHDA methods. Among those SHDA methods, SHFA, CDLS, and DDACL belong to the shallow projection approach, while TNT, STN, SSAN, and JMEA belong to the deep projection approach.
For clarity, we summarize all the baselines in \cref{tab:baselines} and list their details below.

\noindent\textbf{SVMt} \cite{Chang2011LIBSVM}. It solely utilizes labeled target samples to learn a support vector machine. We utilize LIBSVM \cite{Chang2011LIBSVM} to implement SVMt, and the regularization parameter $C$ (\textit{i.e.}, Eq.~(1) in \cite{Chang2011LIBSVM}) is set to 1.

\noindent\textbf{NNt} \cite{Abadi2016TensorFlow}. It employs labeled target samples to train a simple neural network. We implement NNt using the TensorFlow framework \cite{Abadi2016TensorFlow} with the following objective function as
\begin{equation} \label{NNt}
\min_{f, g_t} \frac{1}{n_l} \sum_{i = 1}^{n_l} \mathcal{L}_{ce} \big( \mathbf{y}_i^l, f (g_t(\mathbf{x}_i^l)) \big)
+ \tau \big( \left\| g_t \right\|^2 + \left\| f \right\|^2 \big),
\end{equation}
where $\mathcal{L}_{ce} (\cdot, \cdot)$ is the cross-entropy loss function, $g_t (\cdot)$ is a single-layer fully connected network with the Leaky ReLU activation function \cite{Maas2013Rectifier}, and $f(\cdot)$ is a softmax classifier. We optimize \eqref{NNt} by utilizing the Adam optimizer \cite{Kingma2015Adam} with a learning rate of 0.01, and empirically set $\tau = 0.001$. Also, the dimensionality of hidden layer representations is set to 256, and the number of iterations is specified as 100.

\noindent\textbf{SHFA} \cite{Li2014Learning}. It first augments projected source and target samples with original ones and then learns a support vector machine in a semi-supervised manner. For all tasks, we employ the default parameter settings described in Section 4.1 of \cite{Li2014Learning}, and the parameter $\lambda$ (\textit{i.e.}, Eq~(16) in \cite{Li2014Learning}) is empirically fixed to 1.

\noindent\textbf{CDLS} \cite{Tsai2016Learning}. It identifies representative cross-domain samples during distribution alignment. The recommended parameter settings detailed in Section 4.1 of \cite{Tsai2016Learning} are used on all tasks.

\noindent\textbf{DDACL} \cite{Yao2020Discriminative}. It learns a softmax classifier by both aligning the distributions across domains and enlarging the discriminability of cross-domain samples. As described in Section 5.1 of \cite{Yao2020Discriminative}, we utilize the default parameter settings for all tasks, and the parameter $\tau$ (\textit{i.e.}, Eq~(12) in \cite{Yao2020Discriminative}) is empirically set to $0.001$.

\noindent\textbf{TNT} \cite{Chen2016Transfer}. It jointly performs feature projection, sample categorization, and distribution alignment in a unified neural network framework. For all tasks, we follow the suggested parameter settings outlined in Section 4.1 of \cite{Chen2016Transfer}.

\noindent\textbf{STN} \cite{Yao2019Heterogeneous}. It adopts soft-labels of unlabeled target samples to reduce the conditional distributional divergence across domains and learns a transferable classifier using labeled cross-domain samples. Following \cite{Yao2019Heterogeneous}, we utilize the default parameter settings on all tasks.

\noindent\textbf{SSAN} \cite{Li2020Simultaneous}. It learns a heterogeneous transfer network by taking the implicit semantic correlation and explicit semantic alignment strategies into consideration. As presented in Section 4.1 in \cite{Li2020Simultaneous}, the recommended settings of parameter are used in all the experiments, and the number of epochs is set to 1000.

\noindent\textbf{JMEA} \cite{Fang2023Semi-Supervised}. It simultaneously learns a transferable classifier and a semi-supervised classifier to acquire high-confident pseudo-labels for unlabeled target samples. For all tasks, we adopt the suggested parameter settings in Section 8.2 of \cite{Fang2023Semi-Supervised} except for the parameter $\rho$ (see Algorithm 2 in \cite{Fang2023Semi-Supervised}). This parameter is empirically fine-tuned to achieve good performance across distinct tasks. Specifically, for tasks from the Office+Caltech-10 and Multilingual Reuters Collection datasets, $\rho$ is set to be $0.0001$. As for tasks from the NUS-WIDE+ImageNet-8 dataset, $\rho$ is set to $0.001$.

Given that the data-loading modules within the open-source codes of various baselines are distinct, we standardize those modules into a unified format. 
This allows for the flexible loading of experimental datasets by editing the corresponding configuration files. Please see details at \href{https://github.com/yyyaoyuan/SHDA}{https://github.com/yyyaoyuan/SHDA}.

\subsection{Evaluation Metric} 

Following \cite{Yao2019Heterogeneous,Li2020Simultaneous,Wang2020Prototype-Matching}, we employ the classification accuracy on unlabeled target samples as the evaluation metric, which is calculated as
\begin{equation}
\text{Accuracy} = \frac{|\mathbf{x}_i^u: \mathbf{x}_i^u \in \mathcal{D}_u \cap \mathbf{\widetilde{y}}_i^u = \mathbf{y}_i^u|}{|\mathbf{x}_i^u: \mathbf{x}_i^u \in \mathcal{D}_u|},
\end{equation}
where $\mathbf{\widetilde{y}}_i^u$ and $\mathbf{y}_i^u$ denote the predicted and ground-truth one-hot labels for unlabeled target sample $\mathbf{x}_i^u$, respectively. Moreover, for a fair comparison, we report the average classification accuracy for each method based on 10 random trials.

\section{Study on the category and feature information of source samples}
\label{section:smss}

As illustrated in \cref{fig:Pipeline}, in the SHDA problem, source and target samples are characterized by completely different types of features. Moreover, there is a lack of paired cross-domain samples to facilitate learning the correspondence between source and target samples. However, even under such challenging circumstances, SHDA methods still yield effective transfer performance. \textit{This inspires us to delve deeper into source samples, each of which comprises both category and feature information}. Accordingly, in this section, we investigate how the category and feature information of source samples influence the performance on the target domain, respectively.
\begin{figure}[t]
\centering
{\includegraphics[width=0.98\columnwidth]{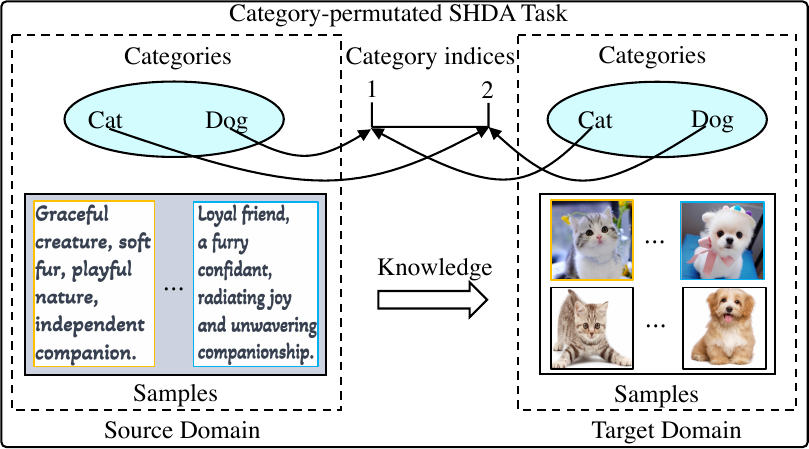}}
\caption{An illustration of the category-permutated SHDA task, where source and target samples have identical categories but with different orders of category indices.}
\label{fig:categoryPermutatedSHDA}
\vspace{-3ex}
\end{figure}
\begin{figure}[t]
\centering
{\includegraphics[width=0.98\columnwidth]{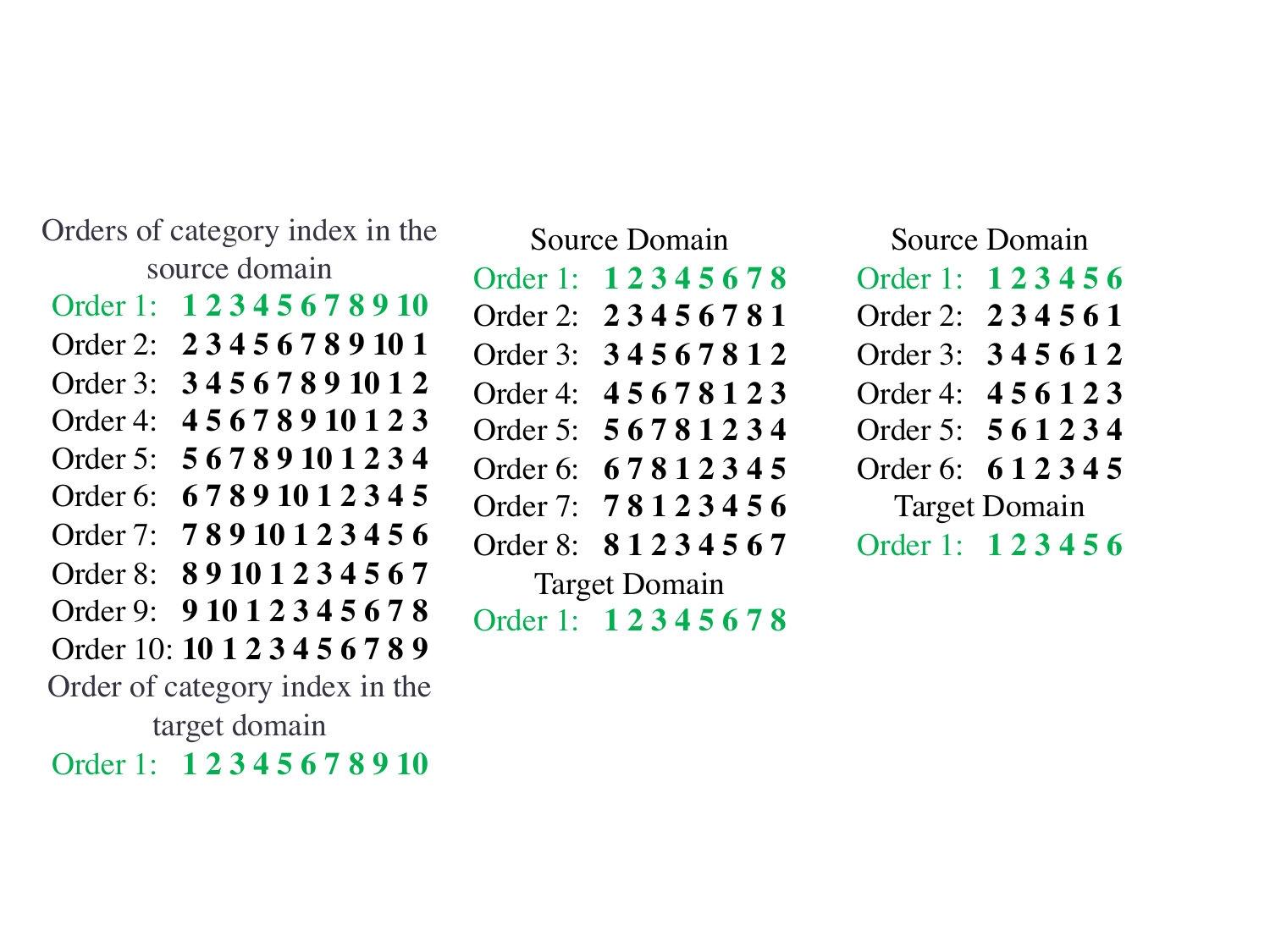}}
\vspace{-2ex}
\caption{The orders of category indices for source and target samples on all datasets. Here, we preserve the order of category indices for target samples while exclusively modifying that of source samples. Consequently, the task is considered as a vanilla SHDA task only when the category indices of both source and target samples are aligned in order 1.}
\label{fig:allOrder}
\vspace{-4ex}
\end{figure}

\subsection{Study on Category Information of Source Samples via Category-permutated SHDA Tasks}
\label{subsection:categoryInformation}

We first study how the category information of source samples affects the performance of the target domain. 
Under the SHDA setting, since source samples are labeled, the primary connection between source and target samples is the presence of a small number of labeled target samples. 
Accordingly, the category information of those labeled target samples plays a vital role in adapting source samples.
Usually each category is mapped to a unique category index for learning and this index is merely a numerical identifier without any semantic meaning.
Thus, in the following experiments, we aim to explore how randomly permutating the category index of all source samples belonging to a specific category to other categories could impact the performance of the target domain.
To this end, we construct a set of \texttt{category-permutated SHDA tasks}.
\cref{fig:categoryPermutatedSHDA} provides an instance of the category-permutated SHDA task, where the source and target domains have the same categories but with different orders of category indices.

According to this setting, we design eight groups of transfer directions: \textbf{A} ($S_{800}$) $\rightarrow$ \textbf{C} ($D_{4096}$), \textbf{C} ($S_{800}$) $\rightarrow$ \textbf{W} ($D_{4096}$), \textbf{W} ($S_{800}$) $\rightarrow$ \textbf{D} ($D_{4096}$), \textbf{Text} $\rightarrow$ \textbf{Image}, \textbf{E} $\rightarrow$ \textbf{S}, \textbf{F} $\rightarrow$ \textbf{S}, \textbf{G} $\rightarrow$ \textbf{S}, and \textbf{I} $\rightarrow$ \textbf{S}. Here, the first three groups are from the Office+Caltech-10 dataset with a total of 10 categories. 
Hence, we construct 10 SHDA tasks for each group by changing the order of category indices for source samples. Specifically, for source samples within the same category, we randomly permutate their category index to correspond to a distinct category.
For instance, if the category index of source samples is $1$, we randomly change it to $5$.
As depicted in \cref{fig:allOrder}, the order $1$ denotes the ground-truth order, while the other orders are permutated, resulting in shifts in the category information. \textit{It is essential to underscore that the order of category indices for target samples strictly adheres to the ground-truth order and remains unchanged throughout all tasks}. Consequently, the SHDA task is considered a vanilla SHDA task only when the order of category indices for source samples adheres to the order $1$; otherwise, it is identified as a category-permutated SHDA task. Similarly, we adopt the same operations to create eight SHDA tasks for the fourth group, and six SHDA tasks for each of the last four groups, based on their respective numbers of categories. 
Thus, we build a total of 62 SHDA tasks. 

\begin{figure*}[t]
\centering
\subfloat[A ($S_{800}$) $\rightarrow$C ($D_{4096}$) \label{fig:ACOrder}] 
{
{\includegraphics[width=0.5\columnwidth]{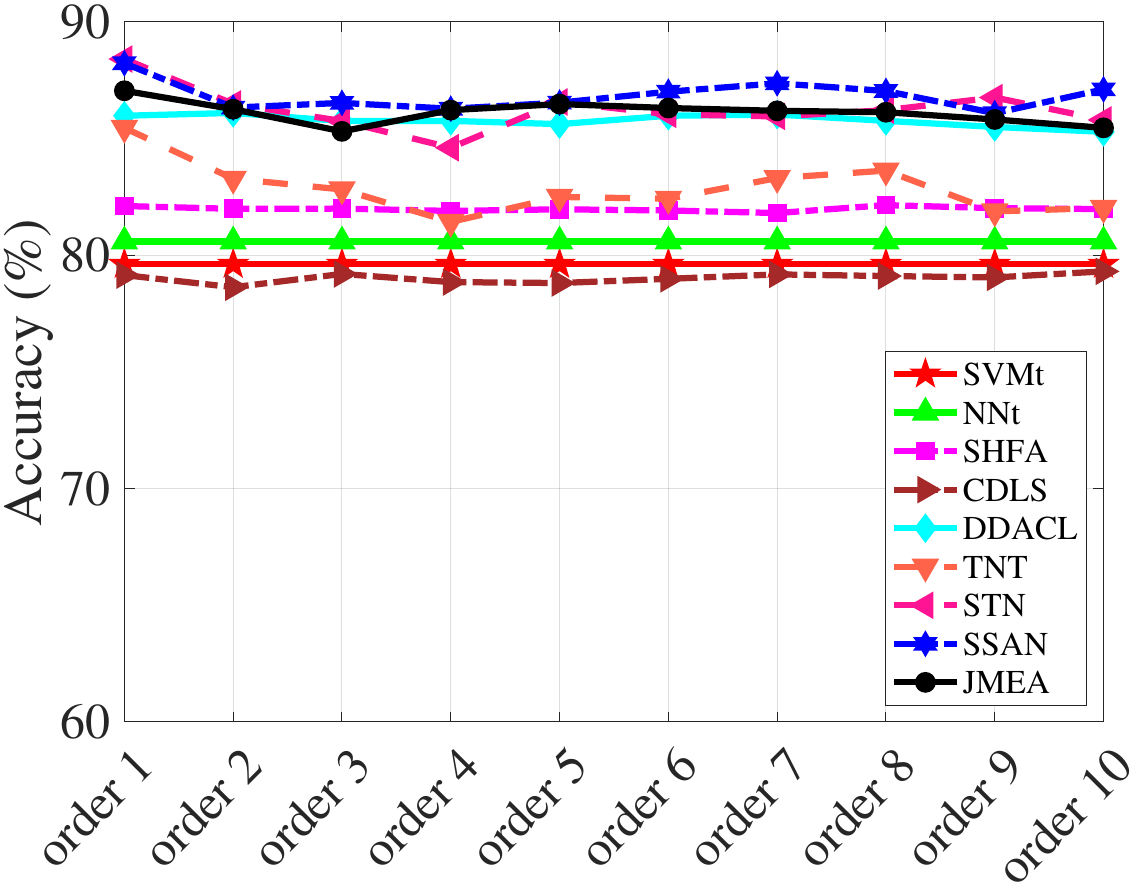}}
}
\hspace{-3.8mm}
\subfloat[C ($S_{800}$) $\rightarrow$ W ($D_{4096}$) \label{fig:CWOrder}] {
{\includegraphics[width=0.5\columnwidth]{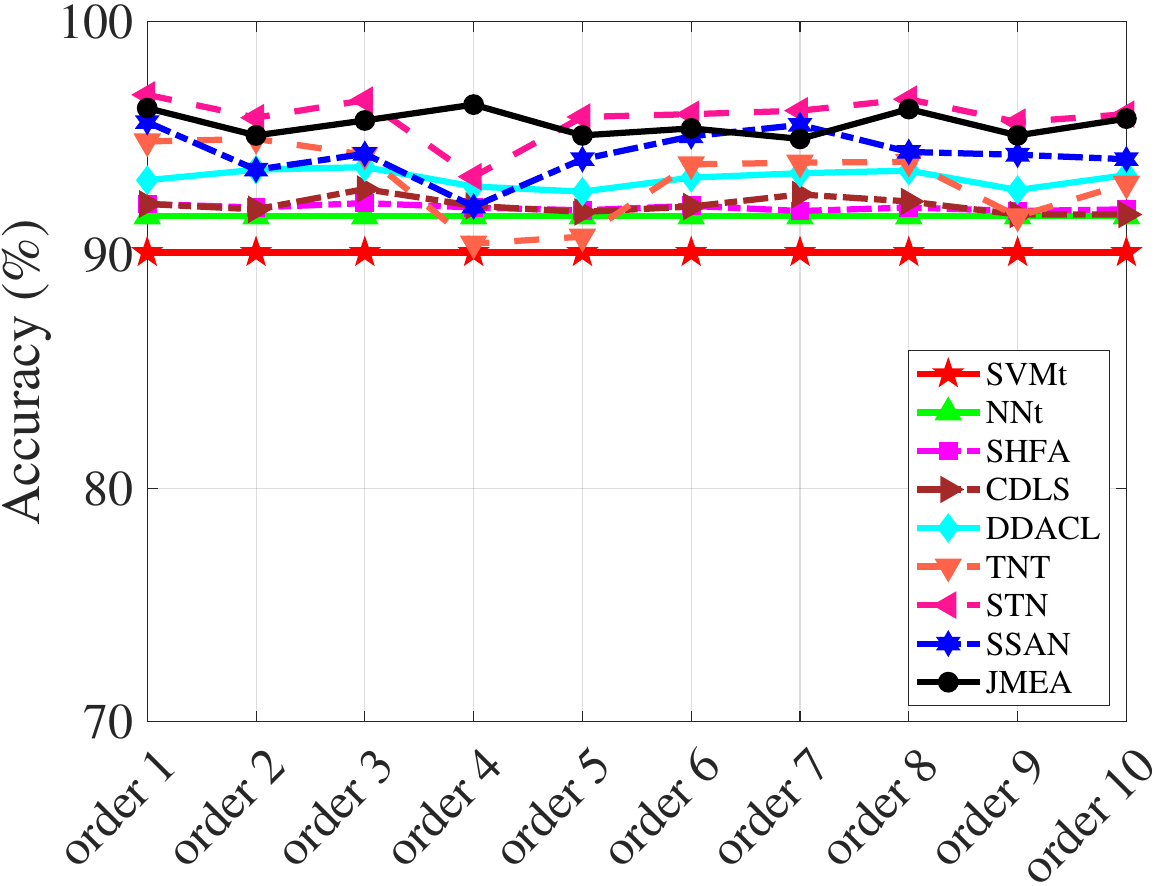}}
}
\hspace{-3.8mm}
\subfloat[W ($S_{800}$) $\rightarrow$D ($D_{4096}$) \label{fig:WDOrder}] {
{\includegraphics[width=0.5\columnwidth]{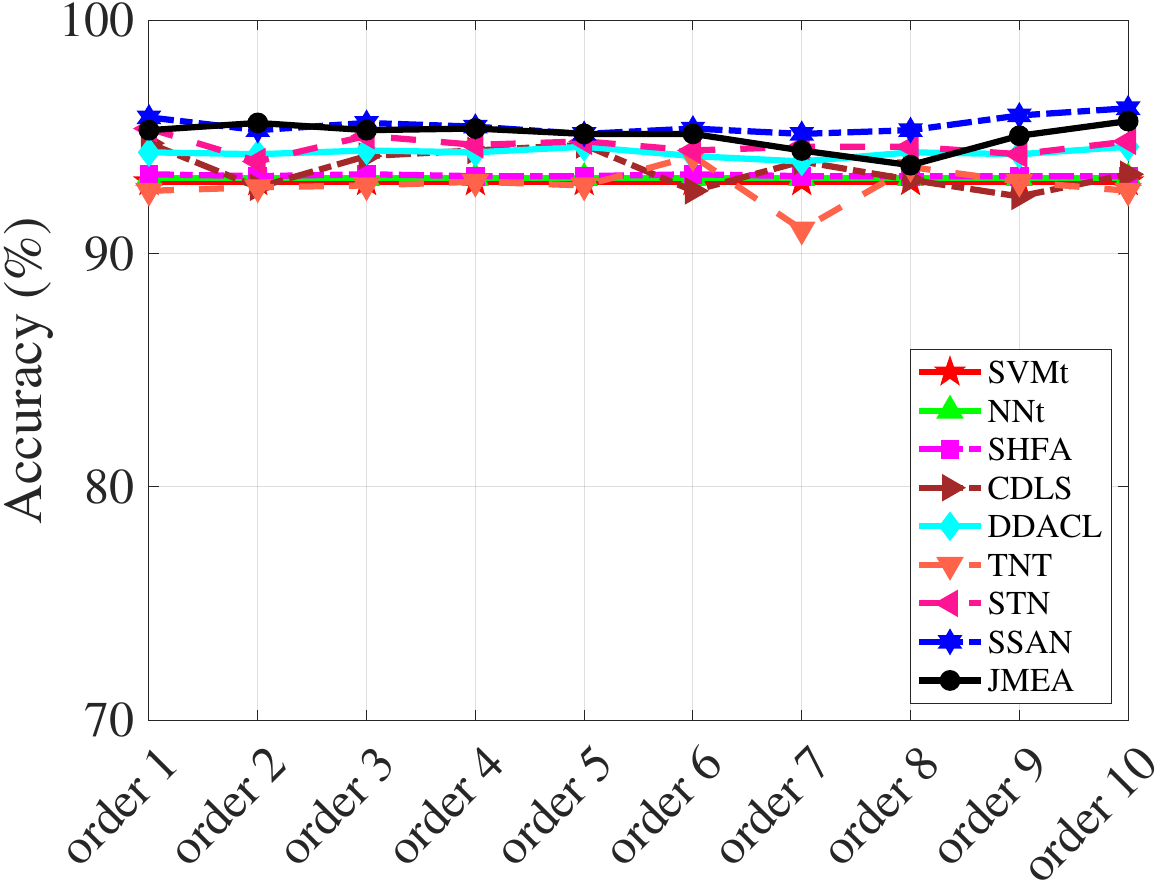}}
}
\hspace{-3.8mm}
\subfloat[Text$\rightarrow$Image \label{fig:TIOrder}] {
{\includegraphics[width=0.5\columnwidth]{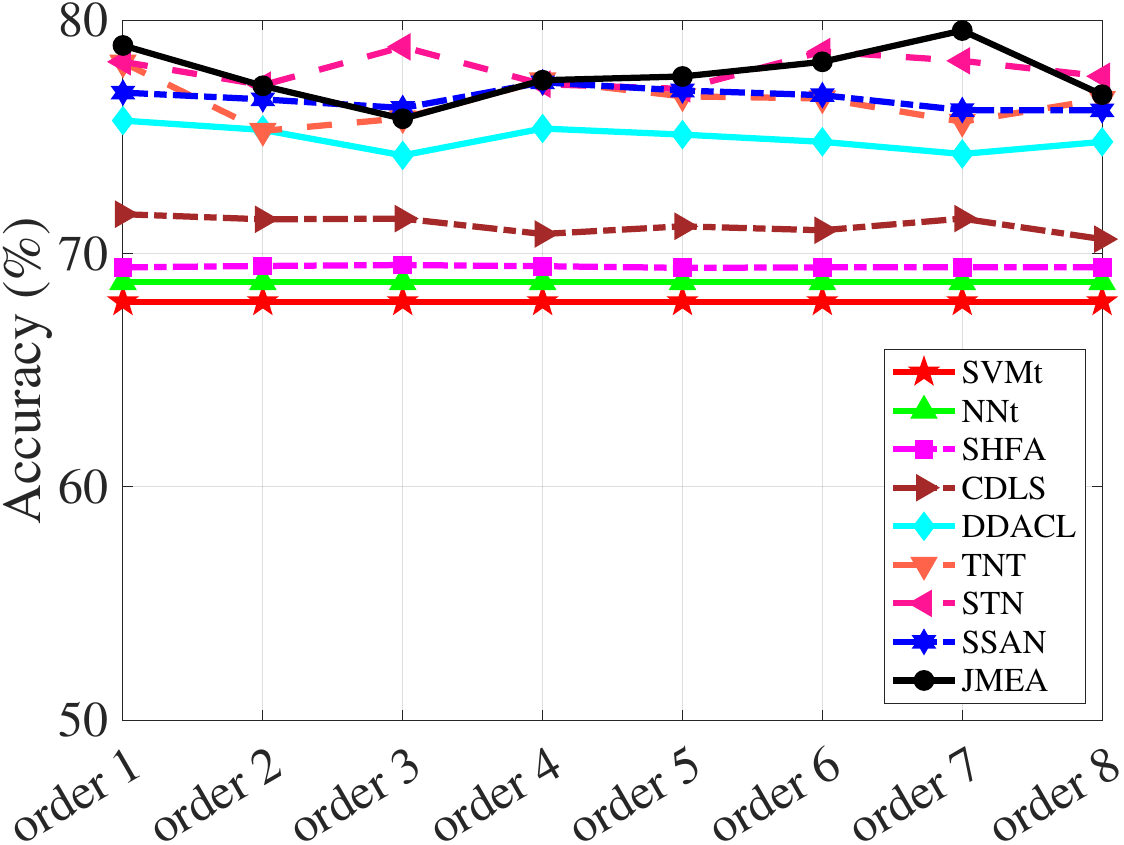}}
}
\hspace{-3.8mm}
\subfloat[E$\rightarrow$S \label{fig:ESOrder}] {
{\includegraphics[width=0.5\columnwidth]{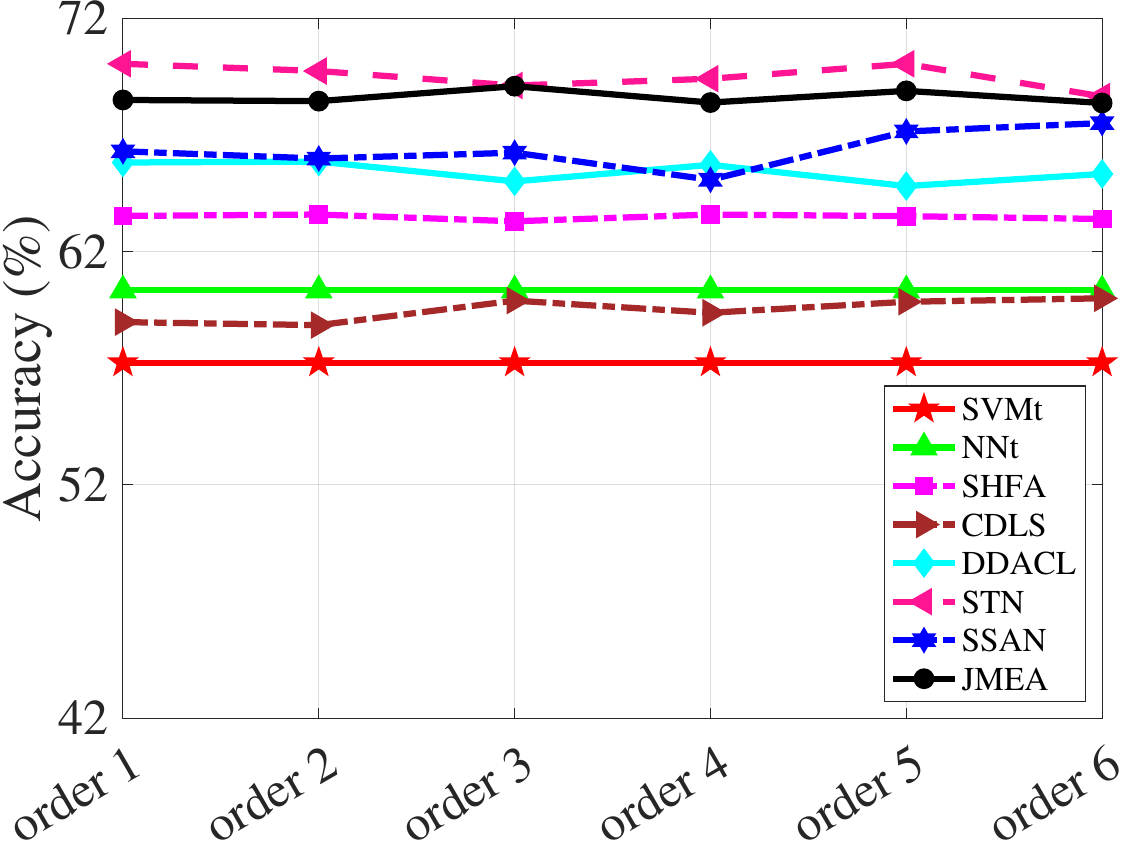}}
}
\hspace{-3.8mm}
\subfloat[F$\rightarrow$S \label{fig:FSOrder}] {
{\includegraphics[width=0.5\columnwidth]{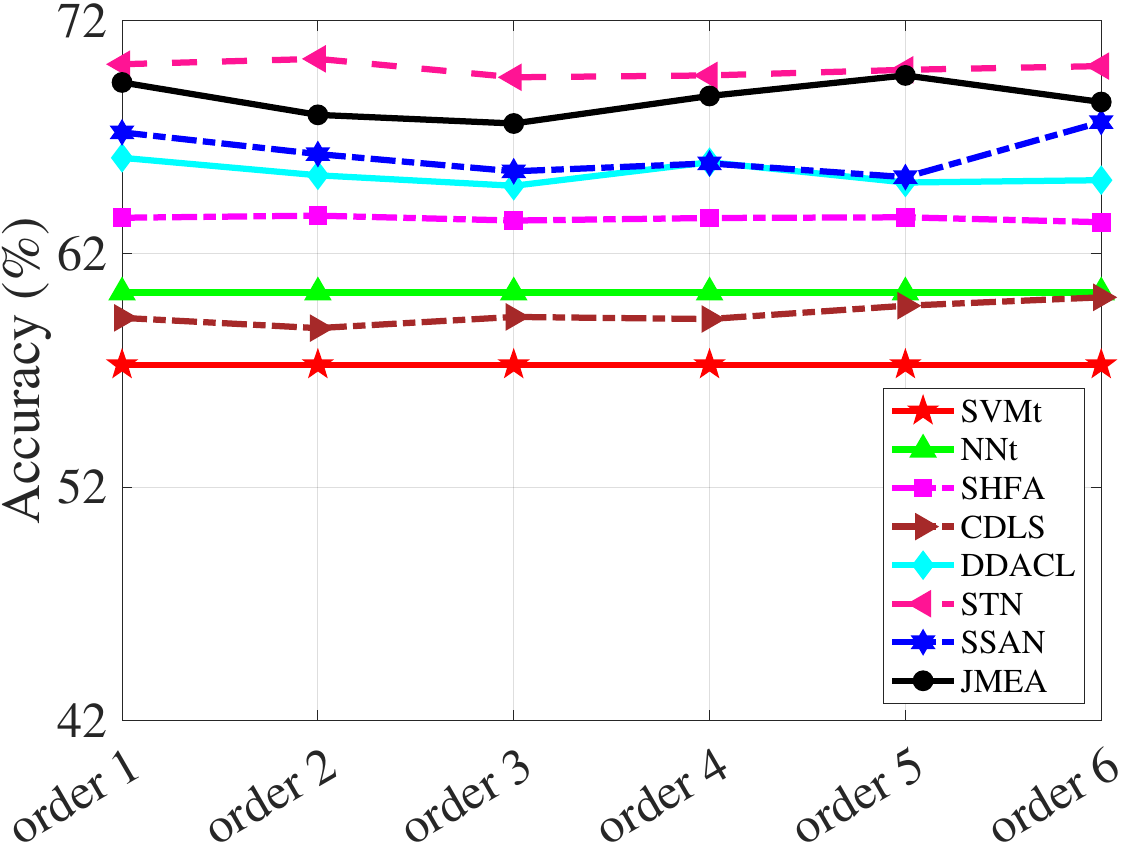}}
}
\hspace{-3.8mm}
\subfloat[G$\rightarrow$S \label{fig:GSOrder}] {
{\includegraphics[width=0.5\columnwidth]{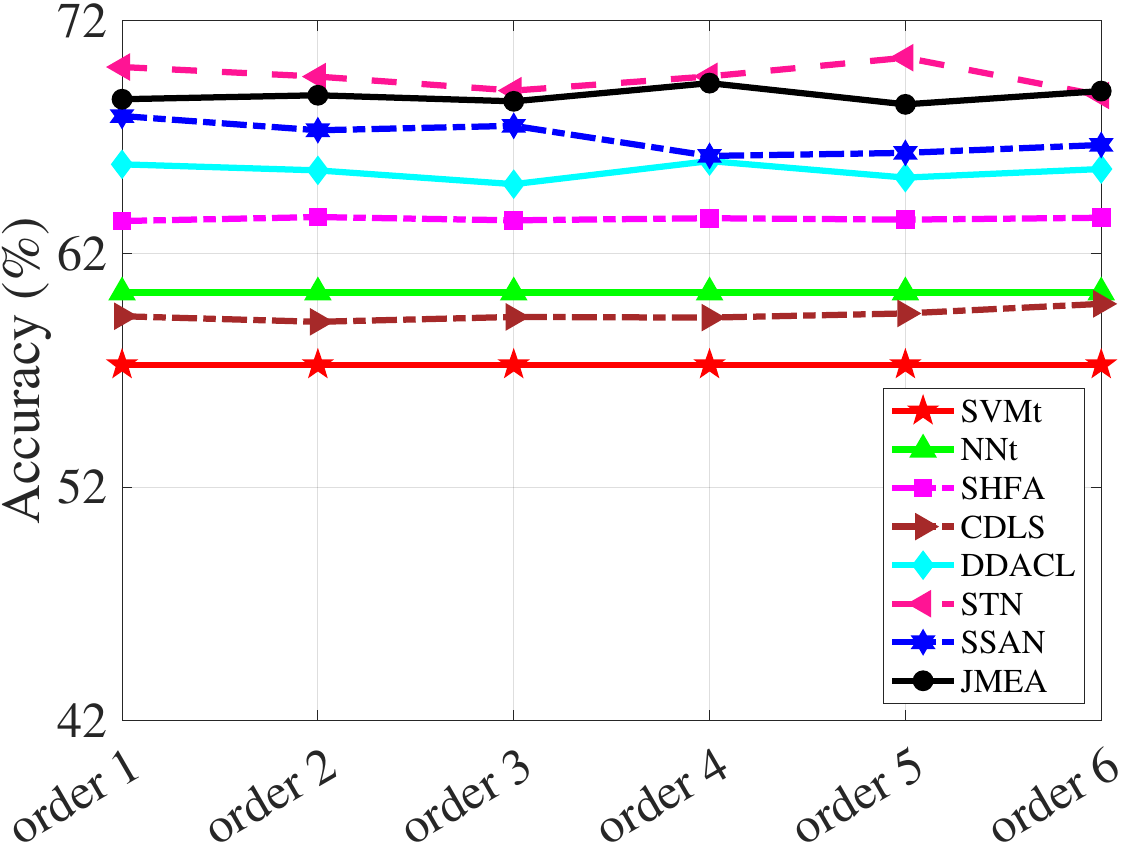}}
}
\hspace{-3.8mm}
\subfloat[I$\rightarrow$S \label{fig:ISOrder}] {
{\includegraphics[width=0.5\columnwidth]{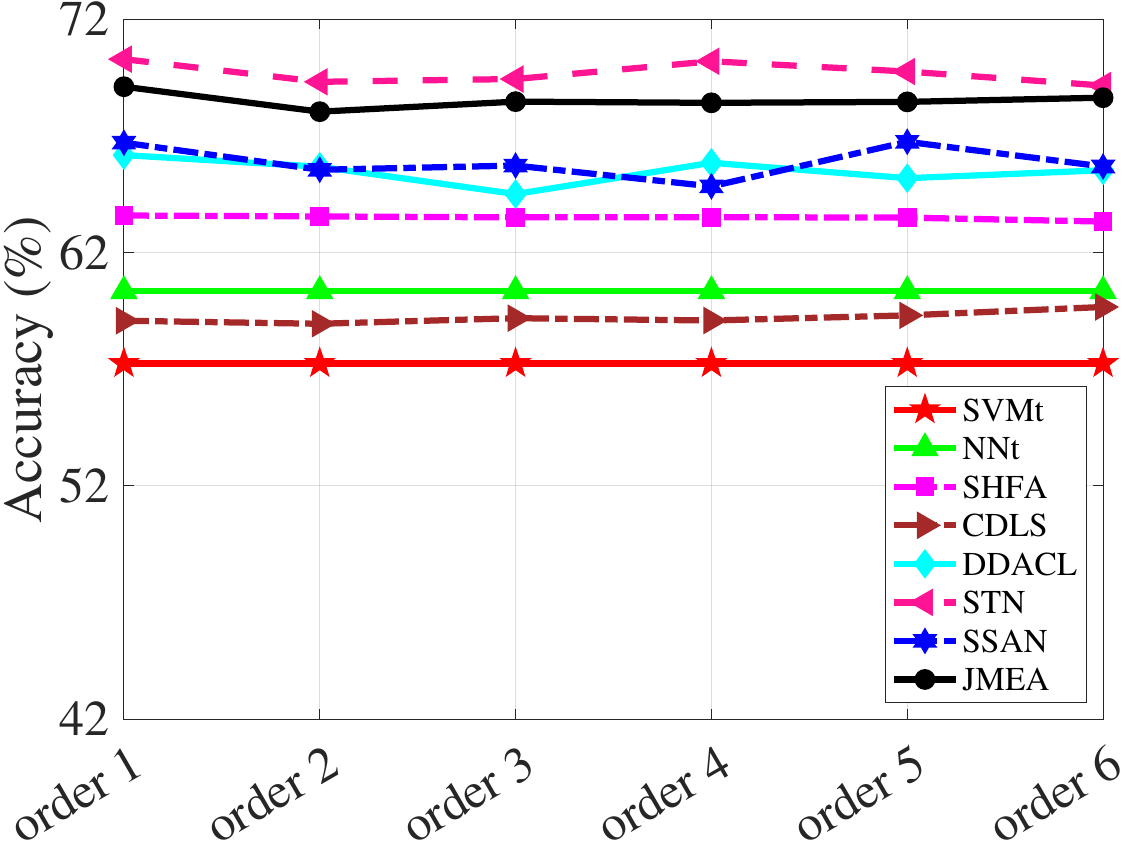}}
}
\caption{Classification accuracies (\%) with distinct orders of category indices for source samples.}
\label{fig:Categoryorder}
\vspace{-5px}
\end{figure*}


\cref{fig:Categoryorder} shows average accuracies of all baselines \textit{w.r.t.} distinct orders of category indices for source samples on all the above 62 tasks. According to the results, we can observe that as the orders of category indices for source samples change, the accuracies of all methods remain almost unchanged. The observation implies that those SHDA methods do not require the actual semantic categories in the source domain to perfectly correspond to those in the target domain. In other words, \textit{those SHDA methods primarily rely on aligning the category indices between source and target samples}. \textit{One important reason is that the source and target feature projectors have completely different architectures and are trained from scratch. Therefore, permutating the category indices of source samples does not significantly affect the training of the target feature projector}. Moreover, in \cref{homogeneousExp} we conduct several additional experiments under the \textit{homogeneous} setting to empirically verify this perspective. 
Overall, all the results indicate that the category information of source samples is not a primary factor influencing the performance of the target domain in SHDA.

\subsection{Study on Feature Information of Source Samples via Cross-dataset SHDA Tasks} 

In the aforementioned experiments, we change the category information of source samples, while their feature information remains unchanged. For instance, in the 10 SHDA tasks with the transfer direction of \textbf{A} ($S_{800}$) $\rightarrow$ \textbf{C} ($D_{4096}$), the feature information of source samples is all $S_{800}$.
Consequently, in the subsequent experiments, our goal is to investigate the impact on the performance of the target domain when utilizing different feature representations for source samples.
 
\begin{figure}[t]
\centering
{\includegraphics[width=0.98\columnwidth]{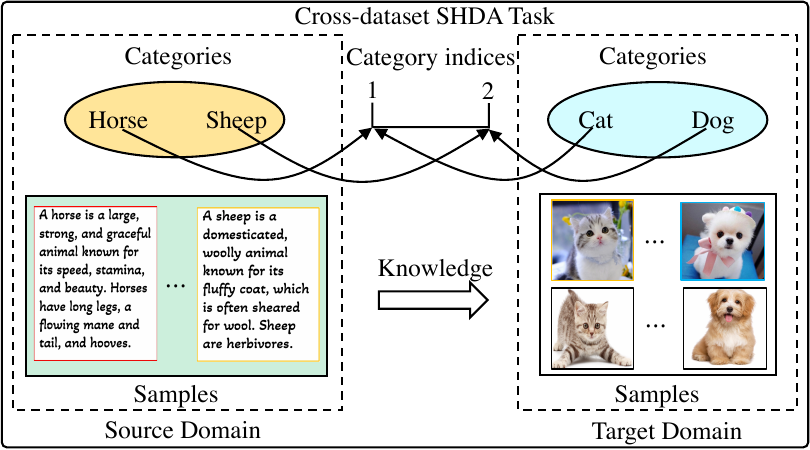}}
\caption{Example illustration of the cross-dataset SHDA task. Here, source and target samples have different categories but are forcibly mapped to the same category indices.}
\label{fig:crossDatasetSHDA}
\vspace{-4ex}
\end{figure}

For this purpose, we design a series of \texttt{cross-dataset SHDA tasks}. An example of the cross-dataset SHDA task is illustrated in \cref{fig:crossDatasetSHDA}, where source and target samples have different categories but are forcibly mapped to the same category indices.
Adhering to the above setting, we treat the domains of \textbf{Image} and \textbf{S} as two target domains, each comprising eight and six categories, respectively. For the former, we choose each source domain from a domain set \{\textbf{Text}, \textbf{A} ($S_{800}$), \textbf{C} ($S_{800}$), \textbf{W} ($S_{800}$), \textbf{A} ($D_{4096}$), \textbf{C} ($D_{4096}$), \textbf{W} ($D_{4096}$)\}. As there are a total of 10 categories in the domains of \textbf{A}, \textbf{C}, and \textbf{W}, we only utilize the samples belonging to the first eight categories as source samples.
Accordingly, source and target samples can be assigned to the same category indices from 1 to 8.
As for the latter, we adopt each domain from a set \{\textbf{E}, \textbf{F}, \textbf{G}, \textbf{I}, \textbf{A} ($S_{800}$), \textbf{C} ($S_{800}$),  \textbf{W} ($S_{800}$), \textbf{A} ($D_{4096}$), \textbf{C} ($D_{4096}$), \textbf{W} ($D_{4096}$), \textbf{Text}\} as the source domain. Analogously, for each domain in \{\textbf{A}, \textbf{C}, \textbf{W}, \textbf{Text}\}, we solely employ the samples associated with the first six categories as source samples. Thus, both source and target samples can be allocated to identical category indices ranging from 1 to 6.
As a result, we establish 18 SHDA tasks in total. Among those tasks, \textbf{Text} $\rightarrow$ \textbf{Image}, \textbf{E} $\rightarrow$ \textbf{S}, \textbf{F} $\rightarrow$ \textbf{S}, \textbf{G} $\rightarrow$ \textbf{S}, and \textbf{I} $\rightarrow$ \textbf{S} are vanilla SHDA tasks, while the rest tasks are cross-dataset SHDA ones.

\begin{figure}[t]
\centering
\subfloat[Target domain: \textbf{Image} \label{fig:targetI}] {
{\includegraphics[width=0.48\columnwidth]{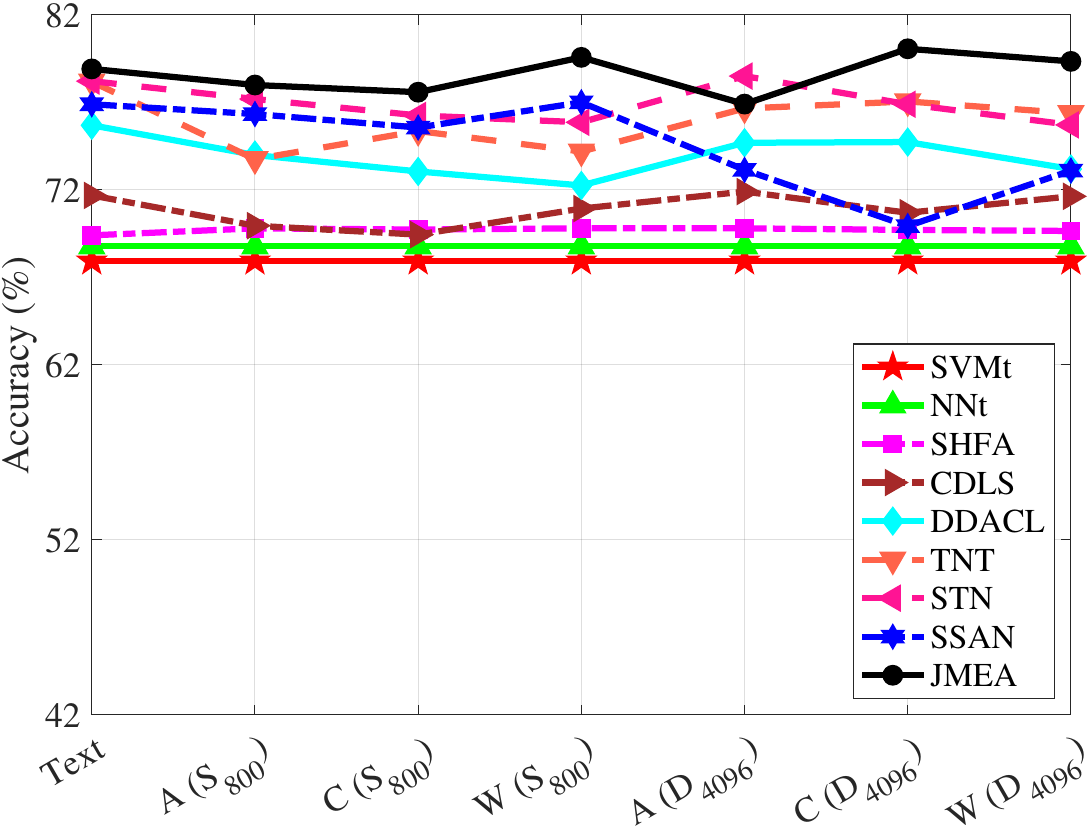}}
}
\hspace{-3.8mm}
\subfloat[Target domain: \textbf{S} \label{fig:targetS}] {
{\includegraphics[width=0.48\columnwidth]{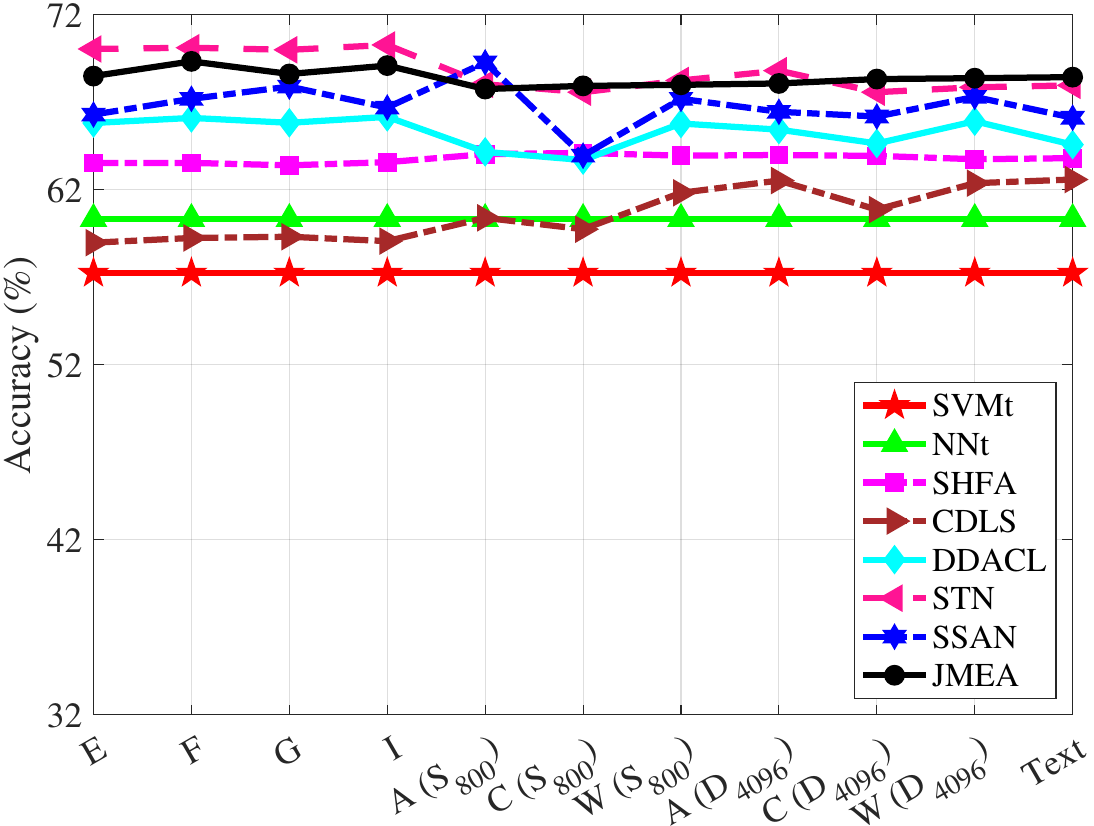}}
}
\caption{Classification accuracies (\%) of different source samples with distinct feature information.}
\label{fig:featuresAll}
\vspace{-3ex}
\end{figure}

\cref{fig:featuresAll} presents the accuracies of all methods \textit{w.r.t.} different source samples with distinct feature information. Based on the results, we can observe that the accuracy curves of most SHDA methods are almost stable. 
This is a surprising observation as it indicates that those SHDA methods can achieve effective knowledge transfer across completely unrelated datasets. \textit{The underlying cause for this phenomenon may be that during domain adaptation, those SHDA methods rely solely on matching category indices to align source samples with target samples, even when source and target samples are completely unrelated. By doing so, the distributions of source and target domains are aligned, thereby enhancing the performance of the target domain}. On the whole, all the results imply that the feature information of source samples is not a dominant factor in affecting the performance of the target domain in SHDA.

\subsection{Summary}

In summary, we make the following important observation.

\noindent\textbf{Observation 1:} \textit{The category and feature information of source samples are not primary factors influencing the performance of the target domain in SHDA}.

\section{Study on Noise as Source Samples} 
\label{snss}

Based on Observation 1, it is evident that the performance of the target domain is not significantly influenced by either the category information or the feature information of the source samples. This observation indicates that the transferable knowledge from the source to the target domain may not inherently rely on the specific semantic categories or detailed feature representations in the source domain. Motivated by this insight, we revisit the necessity of utilizing vanilla source samples in SHDA tasks and propose an innovative hypothesis: \textit{Noise drawn from a random distribution, when used as source samples, may still encapsulate transferable knowledge capable of supporting the adaptation.} Next, we undertake a comprehensive series of experiments to empirically confirm this hypothesis.

\subsection{Study on Source Samples via Noise-injection SHDA Tasks}

\begin{figure}[t]
\centering
{\includegraphics[width=0.98\columnwidth]{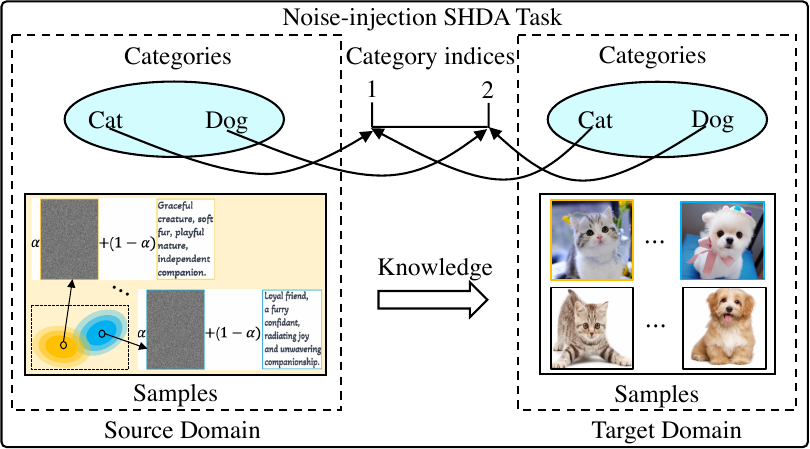}}
\caption{Example illustration of the noise-injection SHDA task. Here, source samples are mixed with distinct ratios of noise.}
\label{fig:noiseInjectionSHDA}
\vspace{-3ex}
\end{figure}

We first investigate the impact of injecting different proportions of noise into source samples on the performance of the target domain. To achieve this, we design several \texttt{noise-injection SHDA tasks}. 
\cref{fig:noiseInjectionSHDA} illustrates an example of noise-injection SHDA tasks, where source samples are mixed with distinct ratios of noise. 
Abiding by this example, we initially select the tasks of \textbf{E} $\rightarrow$ \textbf{S} and \textbf{A} ($S_{800}$) $\rightarrow$ \textbf{C} ($D_{4096}$) as the base tasks. Then, we utilize two different Gaussian mixture distributions to construct two distinct noise domains, \textit{i.e.}, \textbf{NE} and \textbf{NA}. In particular, to establish the \textbf{NE} domain, we directly sample noise from six distinct Gaussian distributions based on the number and dimensionalities of samples in each category of the \textbf{E} domain. Here, each Gaussian distribution is characterized by a unique mean sampled from a standard Gaussian distribution and shares the same covariance, which is set to the identity matrix. This configuration simplifies the setup while ensuring that each distribution represents noise belonging to distinct categories. Similarly, we adopt the same strategy to create the \textbf{NA} domain using 10 different Gaussian distributions. Finally, we inject noise from the \textbf{NE} and \textbf{NA} domains into samples within the \textbf{E} and \textbf{A} ($S_{800}$) domains, respectively. This injection is performed by using varying ratios, \textit{i.e.}, $\textbf{NE}_{\alpha}=\alpha \widetilde{\textbf{NE}} + (1 - \alpha) \widetilde{\textbf{E}}$ and $\textbf{NA}_{\alpha} (S_{800}) = \alpha \widetilde{\textbf{NA}} + (1- \alpha) \widetilde{\textbf{A}} (S_{800})$, where $\alpha$ ranges from 0 to 1 in an increment of 0.2, and $\widetilde{\textbf{X}}$ denotes the sample matrix from the $\textbf{X}$ domain, where samples are arranged in ascending order based on their category index. In principle, the larger the value of $\alpha$, the higher the component of noise. When $\alpha$ equals zero, \textbf{NE}$_{\alpha}$ and \textbf{NA}$_{\alpha}$ ($S_{800}$) become the source domains of \textbf{E} and \textbf{A} ($S_{800}$), respectively. Conversely, when $\alpha$ equals one, \textbf{NE}$_{\alpha}$ and \textbf{NA}$_{\alpha}$ ($S_{800}$) degenerate into the noise domains of 
\textbf{NE} and \textbf{NA}, respectively. Accordingly, we generate 12 noise-injection domains: \textbf{NE}$_{0.0}$, \textbf{NE}$_{0.2}$, \textbf{NE}$_{0.4}$, \textbf{NE}$_{0.6}$, \textbf{NE}$_{0.8}$, \textbf{NE}$_{1.0}$, \textbf{NA}$_{0.0}$ ($S_{800}$), \textbf{NA}$_{0.2}$ ($S_{800}$), \textbf{NA}$_{0.4}$ ($S_{800}$), \textbf{NA}$_{0.6}$ ($S_{800}$), \textbf{NA}$_{0.8}$ ($S_{800}$), and \textbf{NA}$_{1.0}$ ($S_{800}$), where the subscript denotes the value of $\alpha$. As a result, we construct a total of 12 noise-injection SHDA tasks, \textit{i.e.}, 
\textbf{NE}$_{0.0}$ $\rightarrow$ \textbf{S}, \textbf{NE}$_{0.2}$ $\rightarrow$ \textbf{S}, \textbf{NE}$_{0.4}$ $\rightarrow$ \textbf{S}, \textbf{NE}$_{0.6}$ $\rightarrow$ \textbf{S}, \textbf{NE}$_{0.8}$ $\rightarrow$ \textbf{S}, 
\textbf{NE}$_{1.0}$ $\rightarrow$ \textbf{S}, \textbf{NA}$_{0.0}$ ($S_{800}$) $\rightarrow$ \textbf{C} ($D_{4096}$), \textbf{NA}$_{0.2}$ ($S_{800}$) $\rightarrow$ \textbf{C} ($D_{4096}$), 
\textbf{NA}$_{0.4}$ ($S_{800}$) $\rightarrow$ \textbf{C} ($D_{4096}$), 
\textbf{NA}$_{0.6}$ ($S_{800}$) $\rightarrow$ \textbf{C} ($D_{4096}$), 
\textbf{NA}$_{0.8}$ ($S_{800}$) $\rightarrow$ \textbf{C} ($D_{4096}$), and \textbf{NA}$_{1.0}$ ($S_{800}$) $\rightarrow$ \textbf{C} ($D_{4096}$).

\begin{figure}[t]
\centering
\subfloat[Target domain: \textbf{S} \label{fig:mixupTS5}] {
{\includegraphics[width=0.48\columnwidth]{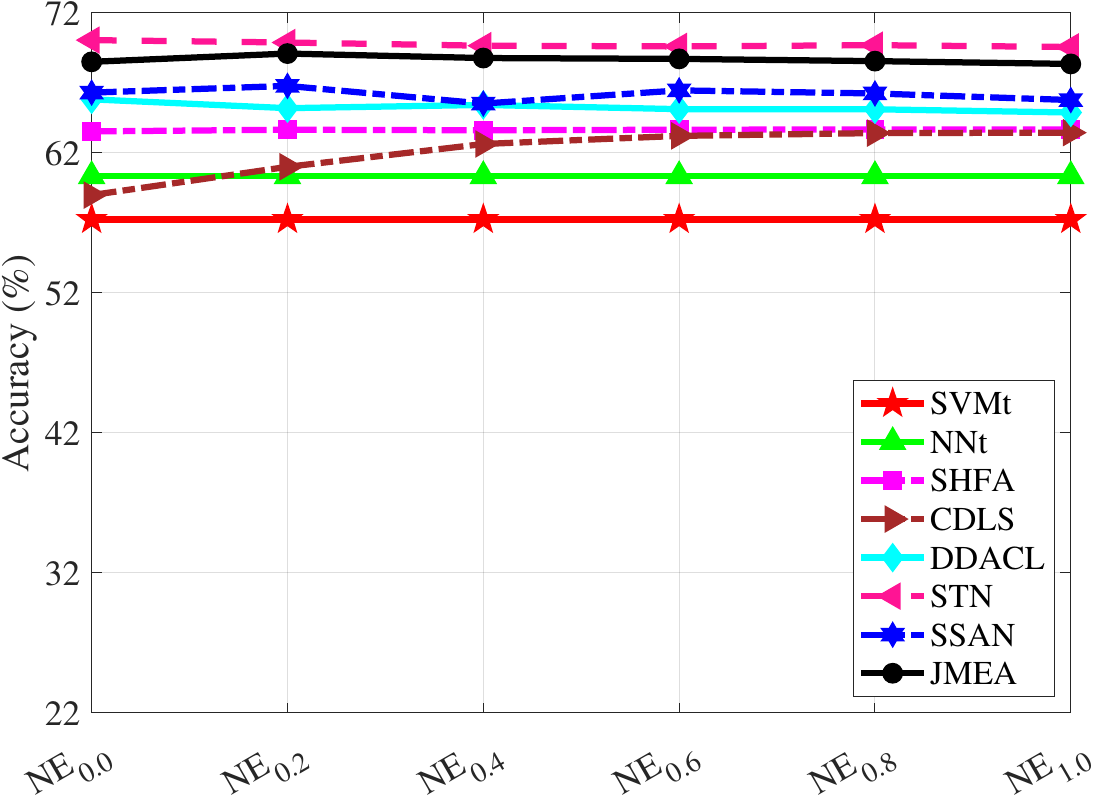}}
}
\hspace{-3.8mm}
\subfloat[Target domain: \textbf{C} ($D_{4096}$) \label{fig:mixupTCD}] {
{\includegraphics[width=0.48\columnwidth]{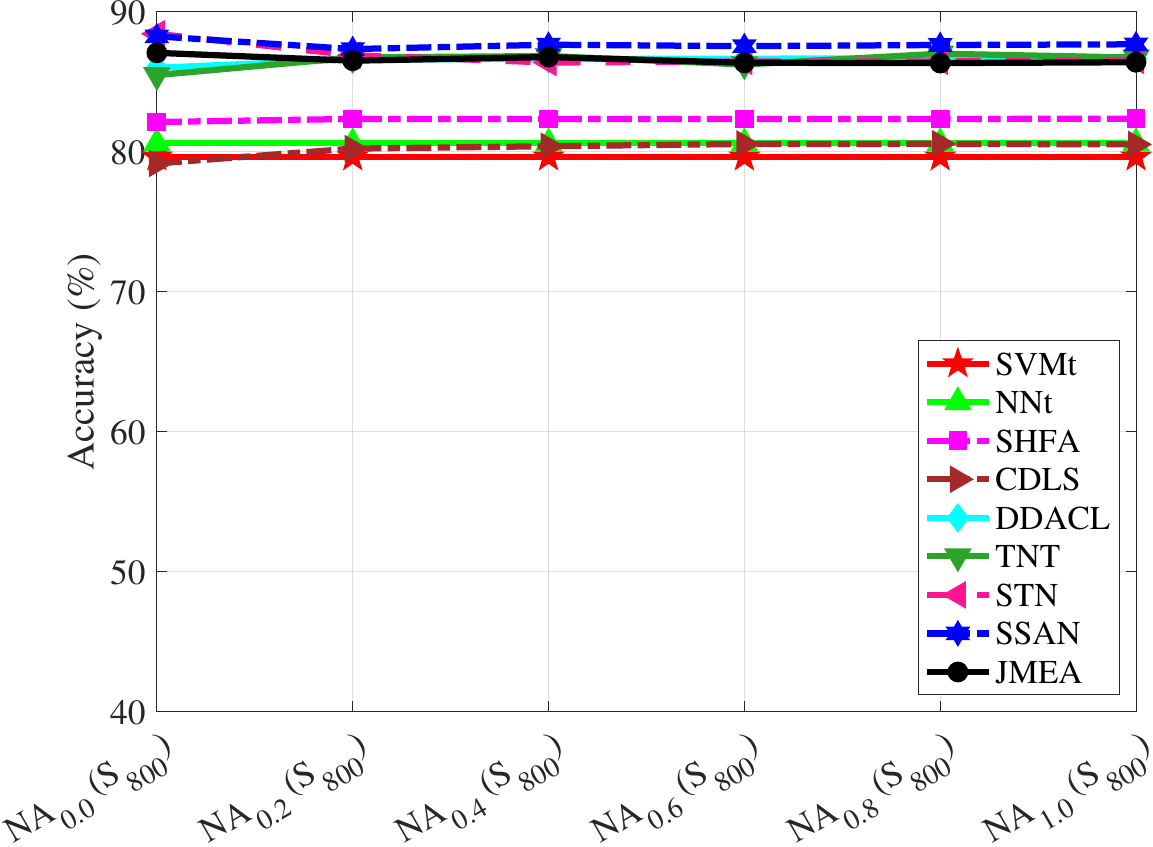}}
}
\caption{Classification accuracies (\%) with different proportions of nosies.}
\label{fig:mixup}
\vspace{-3ex}
\end{figure}

\cref{fig:mixup} shows the accuracies of all methods \textit{w.r.t.} distinct ratios of noise on all the above tasks. From the results, we can see that as the proportion of noise increases, the performance of all methods remains nearly unchanged. Even when the source domains entirely degenerate into noise domains, \textit{i.e.}, $\alpha = 1$, the performance of the target domain is still uncompromised.
The results imply that even if source samples are disturbed by noise, it has no significant impact on the performance of the target domain. Those interesting observations again indicate that the category and feature information of source samples are not primary factors influencing the performance of the target domain. This aligns with the findings from the above experiments in Section \ref{section:smss}.

\subsection{Study on Source Samples via Noise-based SHDA Tasks}

\begin{figure}[t]
\centering
{\includegraphics[width=0.98\columnwidth]{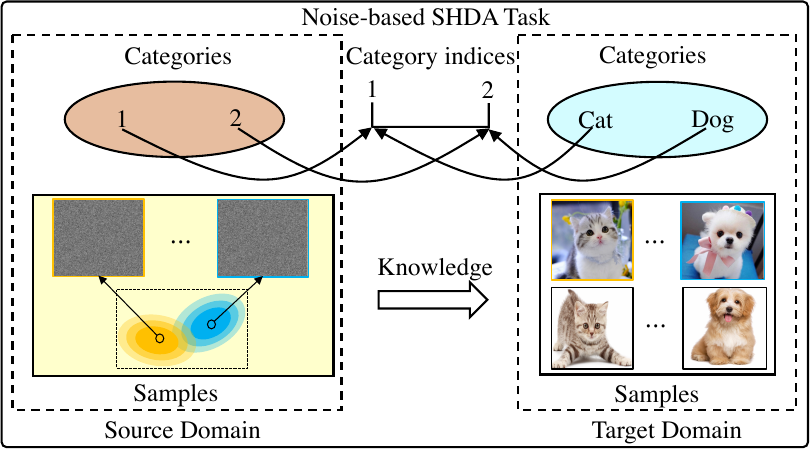}}
\caption{Example illustration of the noise-based SHDA task. Here, source samples consist of noise drawn from a random distribution without any semantic meaning, where the category indices of the target domain are randomly and uniquely assigned to each category of source noise.}
\label{fig:noiseBasedSHDA}
\vspace{-3ex}
\end{figure}

Building upon the above experimental results, we find that using noise drawn from random Gaussian mixture distributions as source samples (we refer to them as \textit{source noise} for brevity) to perform SHDA is feasible and effective. To delve deeper into the influence of source noise on the performance of the target domain, we establish several \texttt{noise-based SHDA tasks} as exemplified in \cref{fig:noiseBasedSHDA}. \textit{Since the source noise lacks semantic meaning, we randomly and uniquely assign the category indices of all categories in the target domain to each category of source noise}. Next, we consider the domains of \textbf{S} and \textbf{C} ($D_{4096}$) as the target domains, respectively, and investigate how the following factors of source noise affect the performance of the target domain: (i) the mean and covariance of source noise; (ii) the number of source noise; (iii) the dimensionality of source noise; and (iv) the distribution of source noise.

\subsubsection{Study on Source Noise with Different Means and Covariances}


To explore the influence of source noise with different means and covariances on the performance of the target domain, we create 10 different noise domains, each derived from a unique Gaussian mixture distribution. For each distribution, $C$ distinct means and variances are generated, where $C = 6$ for the \textbf{S} domain and $C = 10$ for the \textbf{C} ($D_{4096}$) domain. The means are represented as $c \delta \cdot \bm{\mu}_c$ ($c = 1, 2, \dots, C$), while the variances are expressed as $c \delta \cdot \bm{\Sigma}_c$. Each mean $\bm{\mu}_c$ is sampled from a standard Gaussian distribution, and each variance $\bm{\Sigma}_c = \text{PSD} (\frac{\bm{\Sigma} + \bm{\Sigma}^\top}{2})$, where $\bm{\Sigma}$ is a matrix with elements drawn from a standard Gaussian distribution. The operator $\text{PSD} (\cdot)$ sets all negative eigenvalues of its input matrix to zero while retaining non-negative eigenvalues, ensuring that the resulting matrix is positive semi-definite. 
The scaling factor $\delta$ varies from 0.2 to 1.0 in increments of 0.2, producing 5 distinct Gaussian mixture distributions for each of the \textbf{S} and \textbf{C} domains. 
We summarize the norms of the means and covariances for those distributions in \cref{tab:normMeanCovariance}. 
According to the number of categories in those noise domains, we denote them by \textbf{N$_1^6$}, \textbf{N$_2^6$}, \textbf{N$_3^6$}, \textbf{N$_4^6$}, \textbf{N$_5^6$}, \textbf{N$_1^{10}$}, \textbf{N$_2^{10}$}, \textbf{N$_3^{10}$}, \textbf{N$_4^{10}$}, and \textbf{N$_5^{10}$}, respectively. Here, the superscript denotes the total number of categories, while the subscript serves as a unique identifier to distinguish between noise domains with distinct statistical properties. Furthermore, for all noise domains, the number of noise in each category is set to 100, with each noise having a dimensionality of 300. Therefore, we build 10 noise-based SHDA tasks in total, \textit{i.e.}, \textbf{N$_1^6$} $\rightarrow$ \textbf{S}, \textbf{N$_2^6$} $\rightarrow$ \textbf{S}, \textbf{N$_3^6$} $\rightarrow$ \textbf{S}, \textbf{N$_4^6$} $\rightarrow$ \textbf{S}, \textbf{N$_5^6$} $\rightarrow$ \textbf{S}, \textbf{N$_1^{10}$} $\rightarrow$ \textbf{C} ($D_{4096}$), \textbf{N$_2^{10}$} $\rightarrow$ \textbf{C} ($D_{4096}$), \textbf{N$_3^{10}$} $\rightarrow$ \textbf{C} ($D_{4096}$), \textbf{N$_4^{10}$} $\rightarrow$ \textbf{C} ($D_{4096}$), and \textbf{N$_5^{10}$} $\rightarrow$ \textbf{C} ($D_{4096}$).

\begin{table}[htbp]
  \centering
  \caption{The statistics of norms of the means and covariances for the noise domains, where $C$ denotes the total number of categories in each noise domain, and $\bm{\mu}_c$, $\bm{\Sigma}_c$ represent the mean and covariance of category $c$ in each noise domain, respectively.}
  \vspace{-2ex}
    \begin{tabular}{ccc}
    \toprule
    Domain & $\frac{1}{C} \sum_{c=1}^C \| \bm{\mu}_c \|_2$  & $\frac{1}{C} \sum_{c=1}^C \| \bm{\Sigma}_c \|_F$ \\
    \midrule
    \textbf{N}$_1^6$ & 12.62 & 105.34 \\
    \textbf{N}$_2^6$ & 24.44 & 210.32 \\
    \textbf{N}$_3^6$ & 36.16 & 315.39 \\
    \textbf{N}$_4^6$ & 46.74 & 420.53 \\
    \textbf{N}$_5^6$ & 60.43 & 525.05 \\
    \textbf{N}$_1^{10}$ & 19.45 & 164.80 \\
    \textbf{N}$_2^{10}$ & 38.43 & 330.82 \\
    \textbf{N}$_3^{10}$ & 57.24 & 496.16 \\
    \textbf{N}$_4^{10}$ & 77.02 & 661.60 \\
    \textbf{N}$_5^{10}$ & 95.54 & 824.46 \\
    \bottomrule
    \end{tabular}%
  \label{tab:normMeanCovariance}%
  \vspace{-5ex}
\end{table}%

\begin{figure}[t]
\centering
\subfloat[Target domain: \textbf{S} \label{fig:DifferentMeanCovarianceTS5}] {
{\includegraphics[width=0.48\columnwidth]{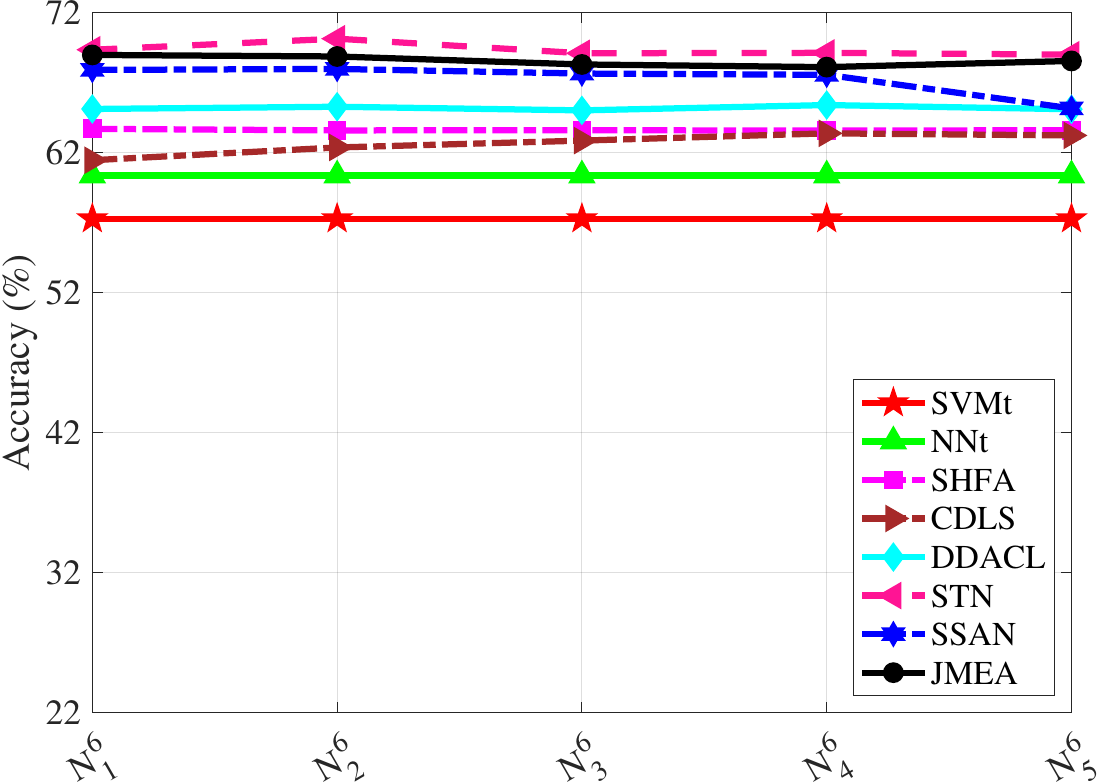}}
}
\hspace{-3.8mm}
\subfloat[Target domain: \textbf{C} ($D_{4096}$) \label{fig:DifferentMeanCovarianceTCD}] {
{\includegraphics[width=0.48\columnwidth]{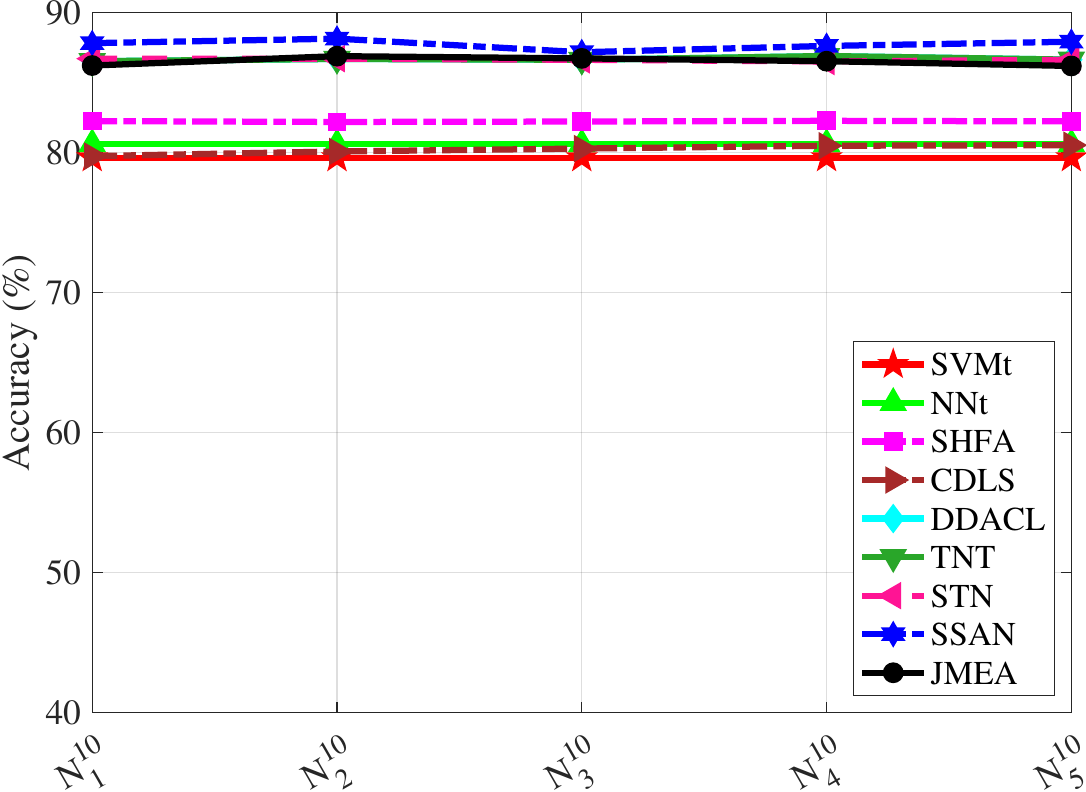}}
}
\caption{Classification accuracies (\%) with various noise domains characterized by distinct means and covariances.}
\label{fig:DifferentMeanCovariance}
\vspace{-2ex}
\end{figure}

\cref{fig:DifferentMeanCovariance} plots the accuracies of all methods \textit{w.r.t.} various source noise characterized by distinct means and covariances. Based on the results, we find that the performance of all methods remains steady despite variations in the means and covariances of source noise. Those results imply that the performance of the target domain is not sensitive to changes in the mean and covariance of source noise.

\subsubsection{Study on Source Noise with Different Sample Numbers}

To evaluate how the number of source noise affects the performance of the target domain, we construct 10 noise domains based on Gaussian mixture distributions, each containing a varying number of source noise. Specifically, for each noise domain, we sample noise directly from $C$ (\textit{i.e.}, $C = 6$ or $10$) distinct Gaussian distributions. Each Gaussian distribution is characterized by a unique mean drawn from the standard Gaussian distribution and shares a common covariance matrix, which is set to the identity matrix. Moreover, the number of source noise per category differs across domains, ranging from 300 to 700 in an increment of 100. Additionally, the dimensionality of noise is consistently fixed at 300 across all noise domains.
Based on the number of noise per category in those noise domains, we denote them by \textbf{NS$_{300}^6$}, \textbf{NS$_{400}^6$}, \textbf{NS$_{500}^6$}, \textbf{NS$_{600}^6$}, \textbf{NS$_{700}^6$},  \textbf{NS$_{300}^{10}$}, \textbf{NS$_{400}^{10}$}, \textbf{NS$_{500}^{10}$}, \textbf{NS$_{600}^{10}$}, and \textbf{NS$_{700}^{10}$}, respectively, where the superscript denotes the total number of categories and the subscript represents the corresponding number of noise per category. As a result, we build a total of 10 noise-based SHDA tasks, \textit{i.e.}, \textbf{NS$_{300}^6$} $\rightarrow$ \textbf{S}, \textbf{NS$_{400}^6$} $\rightarrow$ \textbf{S}, \textbf{NS$_{500}^6$} $\rightarrow$ \textbf{S}, \textbf{NS$_{600}^6$} $\rightarrow$ \textbf{S}, \textbf{NS$_{700}^6$} $\rightarrow$ \textbf{S}, \textbf{NS$_{300}^{10}$} $\rightarrow$ \textbf{C} ($D_{4096}$), \textbf{NS$_{400}^{10}$} $\rightarrow$ \textbf{C} ($D_{4096}$), \textbf{NS$_{500}^{10}$} $\rightarrow$ \textbf{C} ($D_{4096}$), \textbf{NS$_{600}^{10}$} $\rightarrow$ \textbf{C} ($D_{4096}$), and \textbf{NS$_{700}^{10}$} $\rightarrow$ \textbf{C} ($D_{4096}$).

\begin{figure}[t]
\centering
\subfloat[Target domain: \textbf{S} \label{fig:fig:analysisSamplesTS5}] {
{\includegraphics[width=0.48\columnwidth]{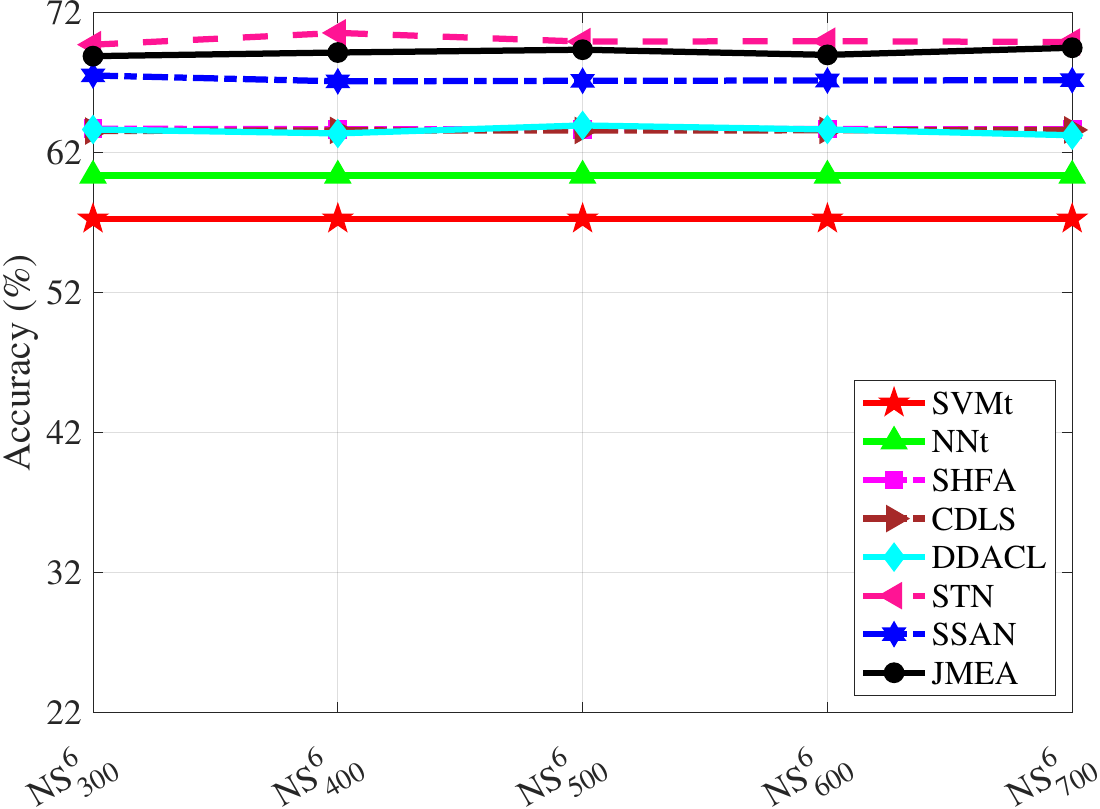}}
}
\hspace{-3.8mm}
\subfloat[Target domain: \textbf{C} ($D_{4096}$) \label{fig:analysisSamplesTCD}] {
{\includegraphics[width=0.48\columnwidth]{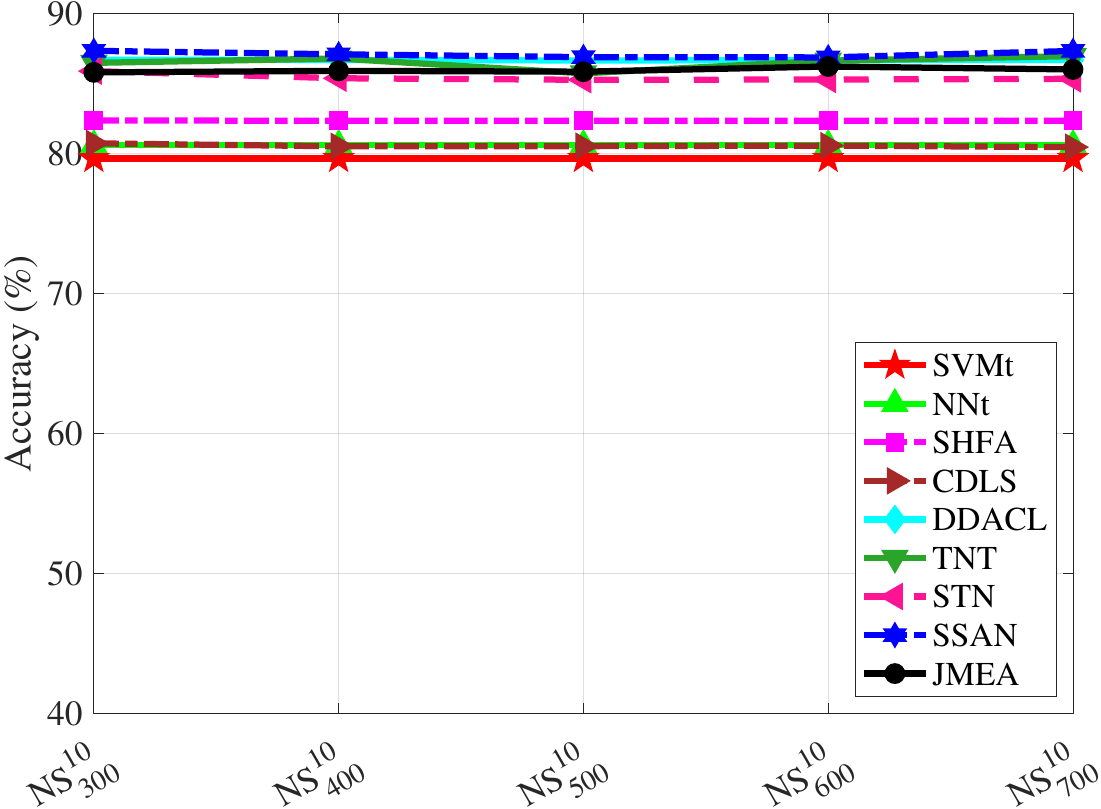}}
}
\caption{Classification accuracies (\%) with different noise domains characterized by distinct sample numbers.}
\label{fig:analysisSamples}
\vspace{-2ex}
\end{figure}

We plot the accuracies of all methods \textit{w.r.t.} the number of source noise in \cref{fig:analysisSamples}. From those results, we find that the performance of all methods is almost constant as the number of samples changes. Those results suggest that the changes in the number of source noise do not have a significant impact on the performance of the target domain.

\subsubsection{Study on Source Noise with Different Dimensionalities} 
To assess how the dimensionality of source noise affects the performance of the target domain, we construct 10 noise domains with different dimensionalities, each sampled from a unique Gaussian distribution. 
In particular, for each noise domain, noise is sampled directly from $C$ (\textit{i.e.}, $C = 6$ or $10$) different Gaussian distributions.  Each of those Gaussian distributions has a unique mean drawn from the standard Gaussian distribution, and all share a common covariance that is set to the identity matrix.
For distinct noise domains, the dimensionalities of noise range from 100 to 500 with an increment of 100. In addition, we fix the number of noise per category to 500 across different noise domains. According to the number of dimensionalities and categories in those noise domains, we name them as \textbf{ND$_{100}^6$}, \textbf{ND$_{200}^6$}, \textbf{ND$_{300}^6$}, \textbf{ND$_{400}^6$}, \textbf{ND$_{500}^6$}, \textbf{ND$_{100}^{10}$}, \textbf{ND$_{200}^{10}$}, \textbf{ND$_{300}^{10}$}, \textbf{ND$_{400}^{10}$}, and \textbf{ND$_{500}^{10}$}, respectively. Consequently, we construct 10 noise-based SHDA tasks in total, \textit{i.e.}, \textbf{ND$_{100}^6$} $\rightarrow$ \textbf{S}, \textbf{ND$_{200}^6$} $\rightarrow$ \textbf{S}, \textbf{ND$_{300}^6$} $\rightarrow$ \textbf{S}, \textbf{ND$_{400}^6$} $\rightarrow$ \textbf{S}, \textbf{ND$_{500}^6$} $\rightarrow$ \textbf{S}, \textbf{ND$_{100}^{10}$} $\rightarrow$ \textbf{C} ($D_{4096}$), \textbf{ND$_{200}^{10}$} $\rightarrow$ \textbf{C} ($D_{4096}$), \textbf{ND$_{300}^{10}$} $\rightarrow$ \textbf{C} ($D_{4096}$), \textbf{ND$_{400}^{10}$} $\rightarrow$ \textbf{C} ($D_{4096}$), and \textbf{ND$_{500}^{10}$} $\rightarrow$ \textbf{C} ($D_{4096}$).

\begin{figure}[t]
\centering
\subfloat[Target domain: \textbf{S} \label{fig:analysisDimensionsTS5}] {
{\includegraphics[width=0.48\columnwidth]{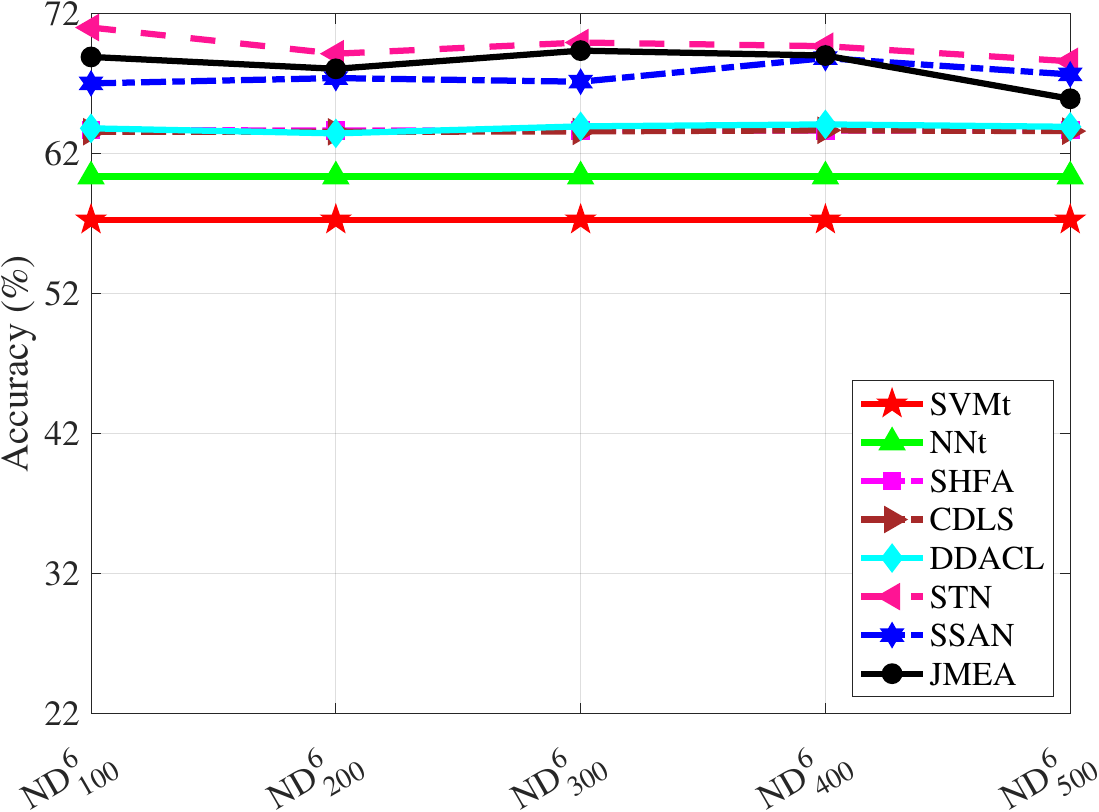}}
}
\hspace{-3.8mm}
\subfloat[Target domain: \textbf{C} ($D_{4096}$) \label{fig:analysisDimensionsTCD}] {
{\includegraphics[width=0.48\columnwidth]{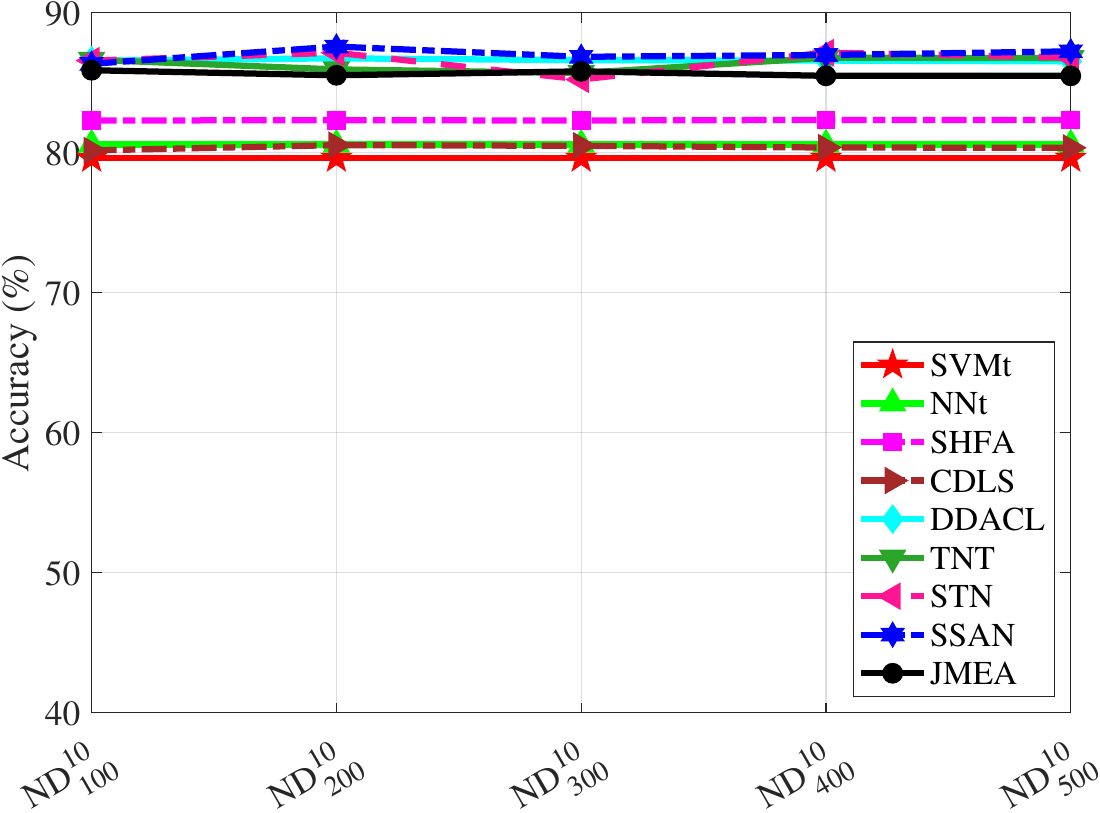}}
}
\caption{Classification accuracies (\%) with different noise domains characterized by distinct dimensionalities.}
\label{fig:analysisDimensions}
\vspace{-2ex}
\end{figure}

The accuracies of all baselines \textit{w.r.t.} the dimensionality of source noise is presented in \cref{fig:analysisDimensions}. We find that when varying the dimensionality of source noise, the performance of all methods remains relatively stable, which indicates that variations in the dimensionality of source noise do not significantly affect the performance of the target domain.

\begin{figure}[t]
\centering
\subfloat[Target domain: \textbf{S} \label{fig:distributionTS5}] {
{\includegraphics[width=0.48\columnwidth]{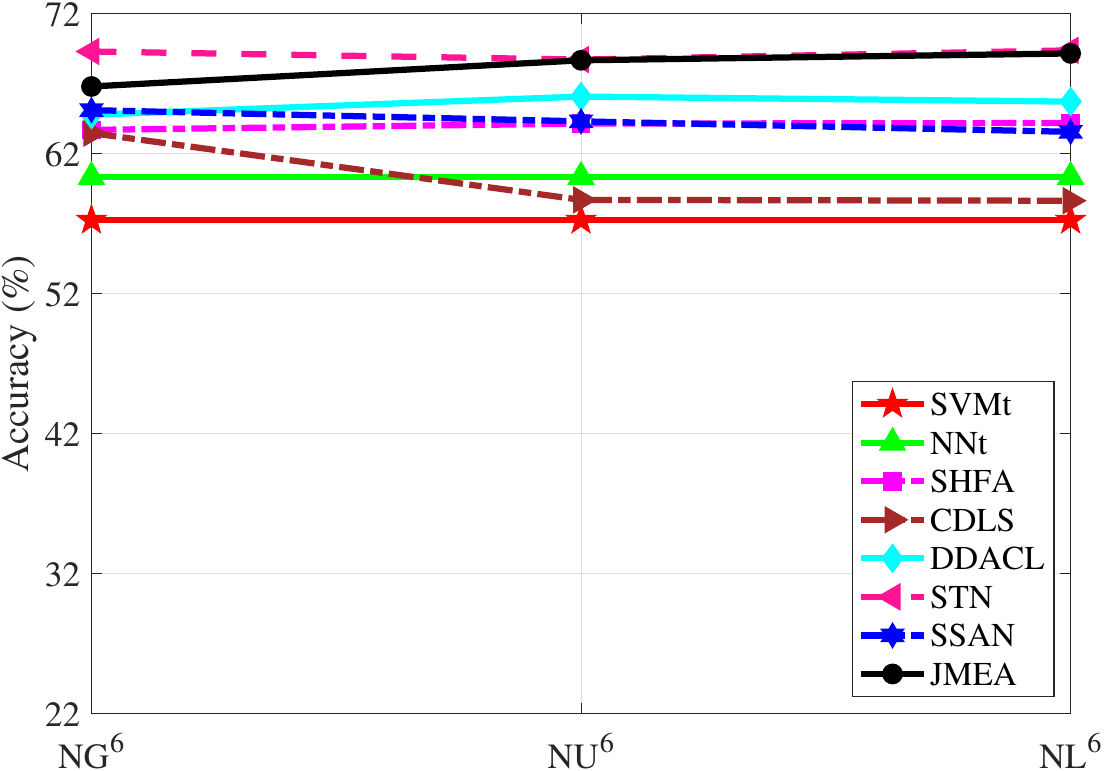}}
}
\hspace{-3.8mm}
\subfloat[Target domain: \textbf{C} ($D_{4096}$) \label{fig:distributionTCD}] {
{\includegraphics[width=0.48\columnwidth]{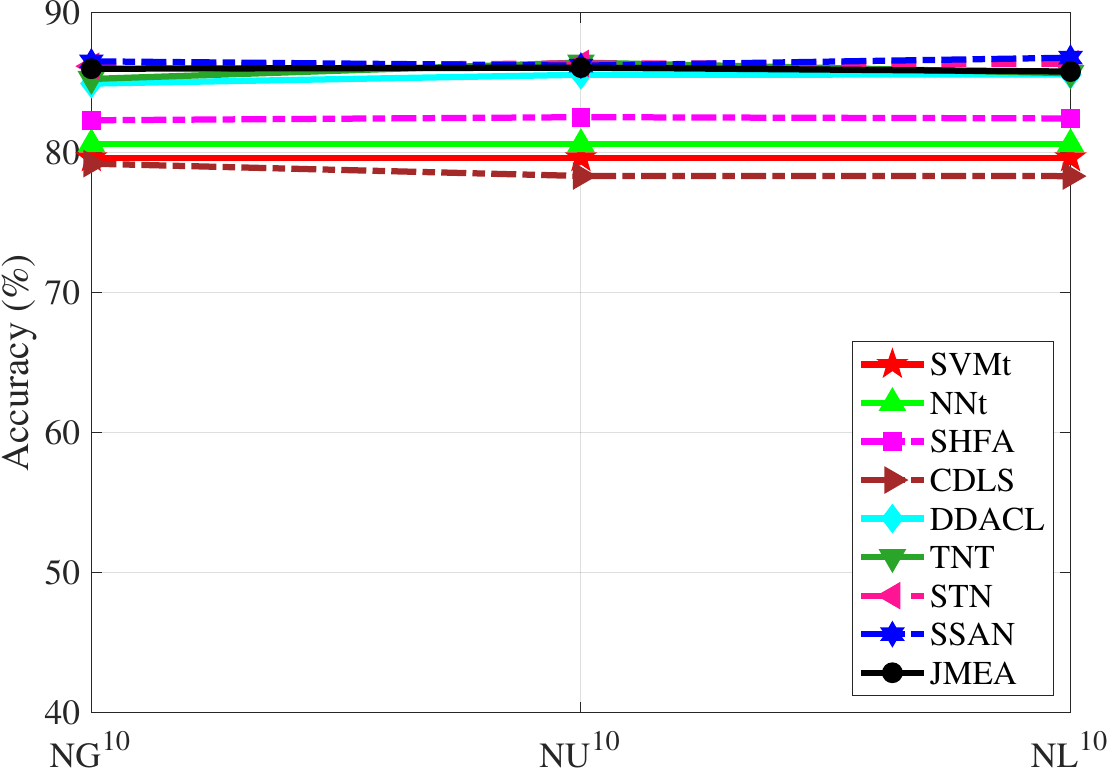}}
}
\caption{Classification accuracies (\%) with different noise domains characterized by distinct types of distributions.}
\label{fig:distributions}
\vspace{-2ex}
\end{figure}

\subsubsection{Study on Source Noise with Different Distributions}
\label{section:distribution}

In the above experiments, all noise domains are drawn from Gaussian distributions. To study the impact of noise domains with different types of distributions on the performance of the target domain, we generate six noise domains using different types of distributions, \textit{i.e.}, \textbf{NG}$^6$, \textbf{NU}$^6$, \textbf{NL}$^6$, \textbf{NG}$^{10}$,  \textbf{NU}$^{10}$, and \textbf{NL}$^{10}$, where the superscript represents the total number of categories. 
Specifically, we utilize the same noise generation process described in the previous section to create the \textbf{NG}$^6$ and \textbf{NG}$^{10}$ domains, respectively. For the construction of the domains of \textbf{NU}$^6$ and \textbf{NU}$^{10}$, we sample noise per category from a Uniform distribution, \textit{i.e.,} $U(-10, 10)$, respectively. 
We build the \textbf{NL}$^6$ and \textbf{NL}$^{10}$ domains by sampling noise within each category from a Laplace distribution, \textit{i.e.}, $L(0, 1)$, respectively. 
Additionally, for a fair comparison, in all noise domains, we fix the number of noise within each category to 100, and the dimensionality of noise is set to 300. 
As a result, we establish six SHDA tasks in total, \textit{i.e.}, \textbf{NG}$^6$ $\rightarrow$ \textbf{S}, \textbf{NU}$^6$ $\rightarrow$ \textbf{S}, \textbf{NL}$^6$ $\rightarrow$ \textbf{S}, \textbf{NG}$^{10}$ $\rightarrow$ \textbf{C} ($D_{4096}$), \textbf{NU}$^{10}$ $\rightarrow$ \textbf{C} ($D_{4096}$), and \textbf{NL}$^{10}$ $\rightarrow$ \textbf{C} ($D_{4096}$).

According to the results plotted in \cref{fig:distributions}, we can observe that using different kinds of distributions has a relatively minor impact on the performance of all methods. Those results suggest that similar phenomena observed with Gaussian distributions could occur with other types of distributions, indicating a relatively general phenomenon.
Those observations once again demonstrate that noise sampled from simple distributions may contain transferable knowledge. 

\subsection{Summary}

All the above experimental results confirm that source noise drawn from simple distributions can yield effective knowledge transfer in SHDA tasks, thereby enhancing the performance of the target domain. This is in line with our expectations and corroborates our hypothesis. Accordingly, we summarize those results into a pivotal observation.

\noindent\textbf{Observation 2:} \textit{Noise drawn from random distributions could contain transferable knowledge for SHDA}.

\section{Study on Transferable Knowledge in SHDA through Source Noise}
\label{section:smsn}

Observation 2 reveals that akin to vanilla source samples, source noise may harbor transferable knowledge. This observation is both surprising and intriguing, prompting us to explore further what knowledge from the source domain is useful for the performance of the target domain in SHDA. Accordingly, in this section, we utilize source noise to delve deeper into the transferable knowledge in SHDA, as it allows us to flexibly construct various source domains.

\subsection{A Unified Knowledge Transfer Framework}

To gain a deeper understanding of transferable knowledge in SHDA, we develop a unified Knowledge Transfer Framework (KTF) to perform large-scale analysis experiments. Specifically, we first construct a common subspace that serves as a shared representation space for both the source and target domains. Within this subspace, we directly generate source noise, eliminating the need to learn a source feature projector. This strategy not only simplifies the analysis but also facilitates a more direct and focused exploration of the transferable knowledge encapsulated within the source noise. 
For simplicity, we denote source noise in the common subspace by $\tilde{\mathcal{D}}_s = \{ (\tilde{\mathbf{x}}_i^s, \mathbf{y}_i^s) \}_{i = 1}^{n_s}$, where $\tilde{\mathbf{x}}^s_i$ is the $i$-th source noise in the common subspace and $\mathbf{y}^s_i$ is its associated one-hot label over $C$ categories. Then, drawing on several typical designs used in most SHDA methods \cite{Hsieh2016Recognizing,Tsai2016Learning,Yao2019Heterogeneous,Yao2020Discriminative,Li2020Simultaneous,Fang2023Semi-Supervised}, we incorporate three crucial factors into KTF. \textbf{(1)} The empirical risk of labeled target samples, \textit{i.e.}, $\mathcal{L}_l$, which characterizes \textit{the discriminability of labeled target samples} with smaller values indicating higher discriminability. 
\textbf{(2)} The empirical risk of source noise, \textit{i.e.}, $\mathcal{L}_s$, which quantifies \textit{the discriminability of source noise} with smaller values signifying higher discriminability.
\textbf{(3)} The distributional divergence between domains, \textit{i.e.}, $\mathcal{L}_{s, t}$, which reflects \textit{the transferability of source noise} with smaller values suggesting stronger transferability. Accordingly, the objective function of KTF is formulated as 
\begin{equation} \label{KTF}
\min_{f, g_t} \mathcal{L}_l + \beta \mathcal{L}_s
+ \mu \mathcal{L}_{s, t} 
+ \tau \big( \left\| g_t \right\|^2 + \left\| f \right\|^2 \big),
\end{equation}
where $\beta$, $\mu$, and $\tau$ are positive trade-off parameters. Recall that $g_t (\cdot)$ is a single-layer fully connected network with the Leaky ReLU activation function \cite{Maas2013Rectifier} to project target samples into the common subspace and $f(\cdot)$ is the domain-shared classifier. As a result, the knowledge from source noise will be transferred into the target domain by optimizing the problem (\ref{KTF}). Next, we elaborate on how to instantiate $\mathcal{L}_l$, $\mathcal{L}_s$, and $\mathcal{L}_{s, t}$, respectively.

The empirical risk of labeled target samples, \textit{i.e.}, $\mathcal{L}_l$, refers to the average loss incurred by a classifier when trained on labeled target samples. To achieve this, we utilize the softmax classifier to instantiate $f (\cdot)$ and cross-entropy loss $\mathcal{L}_{ce} (\cdot, \cdot)$ to instantiate $\mathcal{L}_l$. As a result, we formulate $\mathcal{L}_l$ as
\begin{equation} \label{R_l}
\mathcal{L}_l = \frac{1}{n_l} \sum_{i = 1}^{n_l} \mathcal{L}_{ce} \big[ \mathbf{y}_i^l, f (g_t(\mathbf{x}_i^l)) \big].
\end{equation}

Similar to $\mathcal{L}_l$, we adopt the softmax classifier $f (\cdot)$ and cross-entropy loss $\mathcal{L}_{ce} (\cdot, \cdot)$ to instantiate the empirical risk of source noise, \textit{i.e.}, $\mathcal{L}_s$. Thus, $\mathcal{L}_s$ is formulated by
\begin{equation} \label{R_s}
\mathcal{L}_s = \frac{1}{n_s} \sum_{i = 1}^{n_s} \mathcal{L}_{ce} \big[ \mathbf{y}_i^s, f(\tilde{\mathbf{x}}_i^s) \big].
\end{equation}

The distributional divergence between the source and target domains, \textit{i.e.}, $\mathcal{L}_{s, t}$, aims to quantify the discrepancy in their distributions. To this end, we adopt a simple yet effective method, \textit{Soft Maximum Mean Discrepancy} \cite{Yao2019Heterogeneous}, to instantiate $\mathcal{L}_{s, t}$, which considers both the marginal and conditional distributional divergence. Accordingly, we formulate $\mathcal{L}_{s, t}$ as
\begin{equation} \label{L_d}
\mathcal{L}_{s, t} = \frac{1}{ C + 1} \sum_{c = 0}^C \big\| \mathbf{m}_s^c - \mathbf{m}_t^c \big\|^2,
\end{equation}
where we assign all source noise and target samples to the $0$-th category in the respective domains, $\mathbf{m}_s^c$ represents the average of source noise in the $c$-th category, and $\mathbf{m}_t^c$ denotes the average of target samples for the $c$-th category. Specifically, $\mathbf{m}_s^c$ is defined as
\begin{equation} \label{m_s^c}
\mathbf{m}_s^c = \frac{1}{\sum_{i = 1}^{n_s} \mathbb{I}_c (\tilde{\mathbf{x}}_i^s)} \sum_{i = 1}^{n_s} \mathbb{I}_c (\tilde{\mathbf{x}}_i^s) \tilde{\mathbf{x}}_i^s,
\end{equation}
where $\mathbb{I}_c (\mathbf{x})$ is an indicator function that equals 1 if the sample $\mathbf{x}$ belongs to category $c$, and 0 otherwise. 
Moreover, since the target domain contains a large number of unlabeled samples, we follow \cite{Yao2019Heterogeneous} to adopt the soft-labels of unlabeled target samples provided by $f (\cdot)$ and $g_t (\cdot)$ to estimate $\mathbf{m}_t^c$. Hence, $\mathbf{m}_t^c$ is defined by
\begin{equation} \label{m_t^c}
\mathbf{m}_t^c = 
\frac{\sum_{i = 1}^{n_l} g_t (\mathbb{I}_c (\mathbf{x}_i^l) \mathbf{x}_i^l) + \sum_{i = 1}^{n_u} \hat{y}_{i, c}^u g_t (\mathbf{x}_i^u)}{\sum_{i = 1}^{n_l} \mathbb{I}_c (\mathbf{x}_i^l) + \sum_{i = 1}^{n_u} \hat{y}_{i, c}^u},
\end{equation}
where 
$\hat{y}_{i, c}^u$ stands for the predicted probability by $f (\cdot)$ and $g_t (\cdot)$ that $\mathbf{x}_i^u$ belongs to category $c$.

In the implementation, we employ a single-layer fully connected network with Leaky ReLU \cite{Maas2013Rectifier} and softmax activation functions to instantiate $g_t(\cdot)$ and $f(\cdot)$ in problem (\ref{KTF}), respectively. We empirically set hyperparameters $\beta$ and $\tau$ to be 0.1 and 0.05 for all tasks, respectively. Regarding the hyperparameter $\mu$, we empirically set it  to 0.1 for tasks with the target domain of \textbf{S}, and to 1 for tasks with the target domain of \textbf{C} ($D_{4096}$). Also, we fix the dimensionality of the common subspace to 256 and the number of iterations to 600. Moreover, we utilize the Adam optimizer \cite{Kingma2015Adam} with a learning rate of 0.001 to optimize problem~(\ref{KTF}).
 
\begin{figure*}[t]
\vspace{-1mm}
\centering
\subfloat[Target domain: \textbf{S} \label{fig:XsTS5}] {
{\includegraphics[width=0.48\columnwidth]{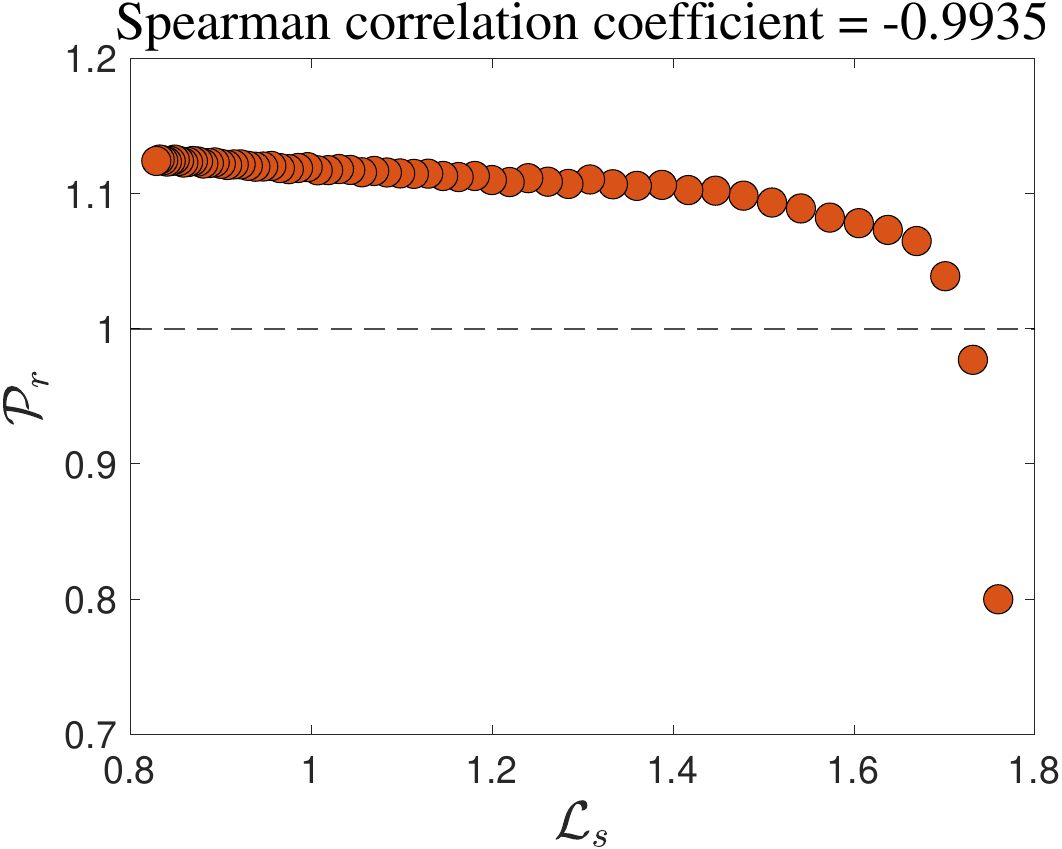}}
}
\subfloat[Target domain: \textbf{S} \label{fig:DistributionTS5}] {
{\includegraphics[width=0.48\columnwidth]{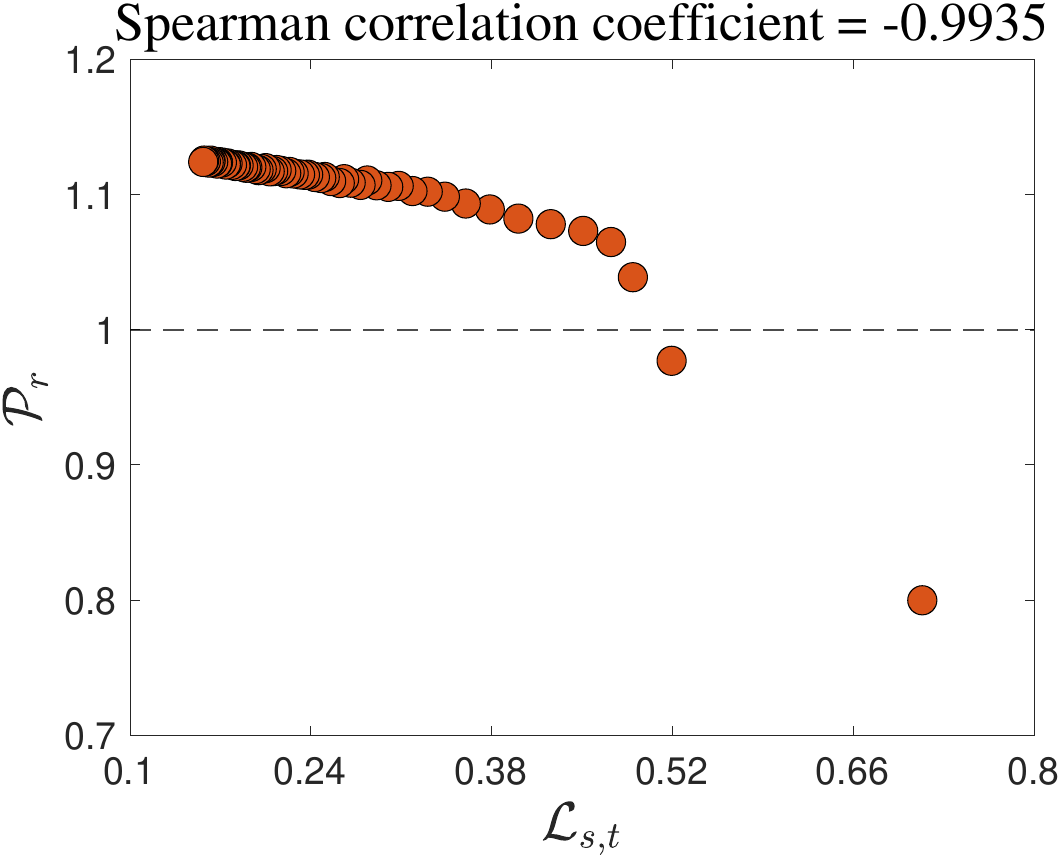}}
}
\subfloat[Target domain: \textbf{C} ($D_{4096}$) \label{fig:XsTCD}] {
{\includegraphics[width=0.48\columnwidth]{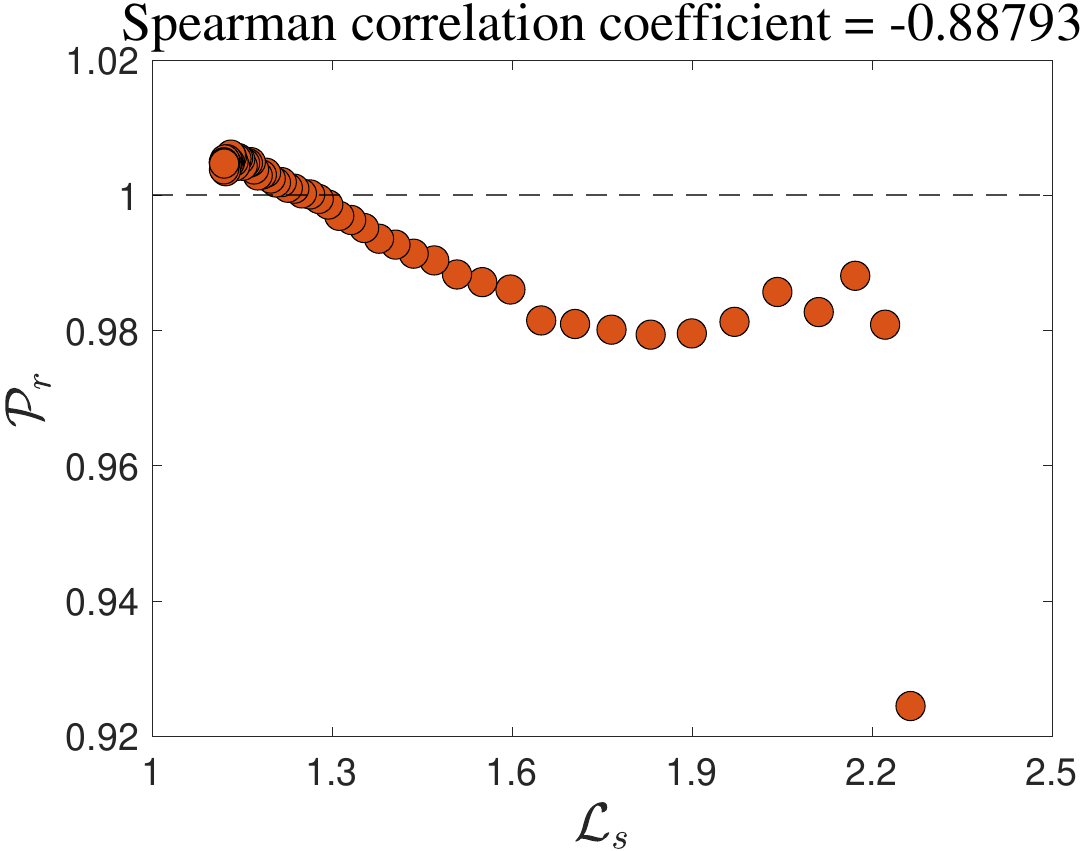}}
}
\subfloat[Target domain: \textbf{C} ($D_{4096}$) \label{fig:DistributionTCD}] {
{\includegraphics[width=0.48\columnwidth]{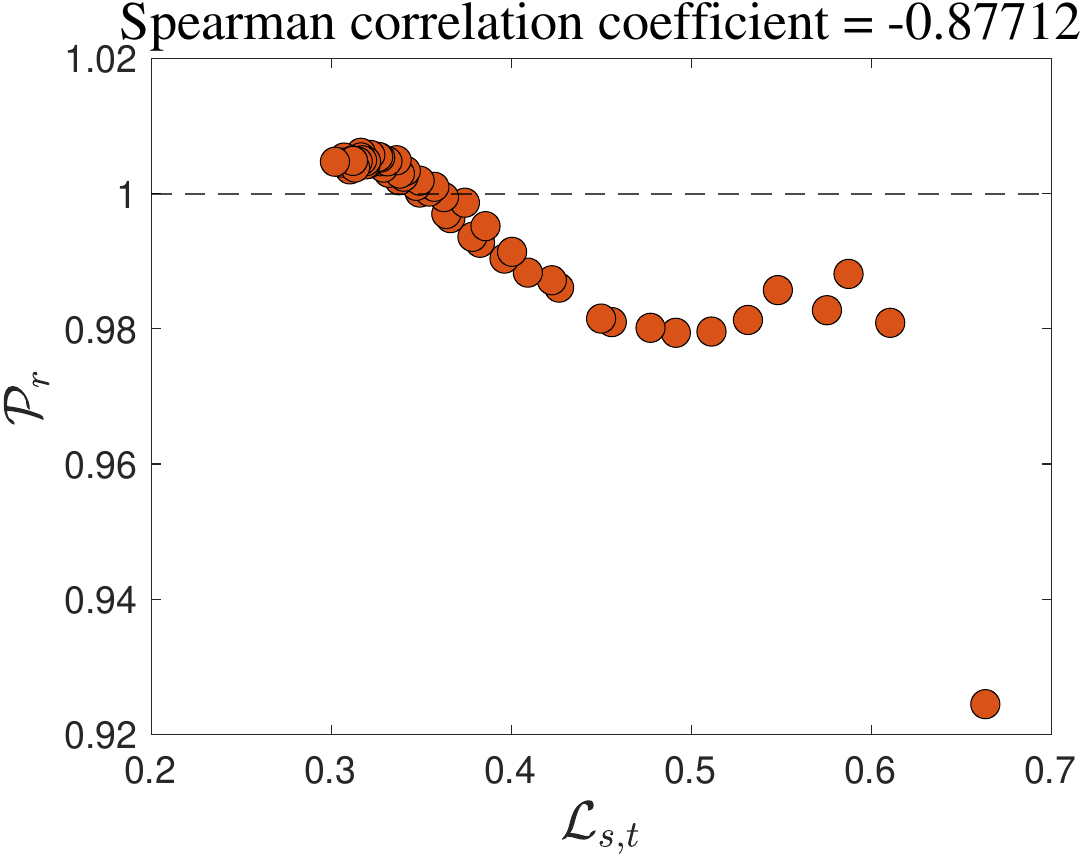}}
}
\caption{Correlation between $\mathcal{L}_s$ and $\mathcal{P}_r$, as well as between $\mathcal{L}_{s, t}$ and $\mathcal{P}_r$. Here, $\mathcal{L}_s$ represents the discriminability of the source domain, $\mathcal{L}_{s, t}$ characterizes the transferability of the source domain, and $\mathcal{P}_r$ denotes the performance improvement ratio in the target domain.}
\label{fig:spearman}
\vspace{-4ex}
\end{figure*}

\subsection{Analysis}

In this section, we utilize KTF to analyze the essence of transferable knowledge stored in source noise.

\subsubsection{Analysis on Transferable Knowledge}

Based on the objective function of KTF formulated in problem~(\ref{KTF}), we find that there are two primary factors (\textit{i.e.}, $\mathcal{L}_s$ and $\mathcal{L}_{s, t}$) closely related to the transferable knowledge. The former represents the discriminability of the source domain, while the latter characterizes the transferability of the source domain. To analyze the impact of those two factors on the performance of the target domain, we analyze the correlation between them and the performance improvement ratio in the target domain \cite{Wei2018Transfer}, respectively, where the performance improvement ratio denoted by $\mathcal{P}_r$ is defined as the ratio of the performance of KTF to that of NNt on unlabeled target samples. Hence, the larger $\mathcal{P}_r$ is, the better the transfer performance is. To assess such correlation, we employ the Spearman correlation coefficient \cite{Spearman1904The}, given its robustness to outliers and nonlinear property. The Spearman correlation coefficient ranges from $-1$ to $1$ with values close to $1$ or $-1$ indicating strong monotonic relationships, and values close to 0 indicating weak or no monotonic relationship.

We begin by choosing the \textbf{S} and \textbf{C} ($D_{4096}$) domains as the target domains, respectively. 
Then, we construct 200 noise domains, each derived from a distinct Gaussian mixture distribution. For each distribution, we generate $C$ distinct means and variances, where $C = 6$ for the \textbf{S} domain and $C = 10$ for the \textbf{C} ($D_{4096}$) domain. The means are expressed as $c \delta \cdot \bm{\mu}_c$ ($c = 1, 2, \dots, C$), and the variances as $c \delta \cdot \bm{\Sigma}_c$. Each mean $\bm{\mu}_c$ is sampled from a standard Gaussian distribution, and each variance $\bm{\Sigma}_c = \text{PSD} (\frac{\bm{\Sigma} + \bm{\Sigma}^\top}{2})$, where $\bm{\Sigma}$ is a matrix with elements drawn from a standard Gaussian distribution. 
The scaling factor $\delta$ ranges from 0.05 to 9.95 with a step size of 0.1, resulting in 100 distinct Gaussian mixture distributions for each of the \textbf{S} and \textbf{C} domains. 
Additionally, to better simulate various practical scenarios, we randomly assign the number of samples for each category in each noise domain from the range of 100 and 1000. The dimensionality of each noise in all noise domains is uniformly set to 256, aligning with the dimensionality of the common subspace in KTF. Therefore, we build 100 noise-based SHDA tasks with \textbf{S} as the target domain and another 100 with \textbf{C} ($D_{4096}$) as the target domain. For each transfer task, we record the values of $\mathcal{L}_s$, $\mathcal{L}_{s, t}$, and $\mathcal{P}_r$ every 10 iterations to alleviate correlation and capture essential trends. With 600 iterations for KTF, this results in 60 tuples $\{(\mathcal{L}_s^i, \mathcal{L}_{s, t}^i, \mathcal{P}_r^i)\}_{i = 1}^{60}$ for each transfer task. Furthermore, to capture the overall trends across different tasks within the same target domain, we average the 60 tuples generated for each transfer task to produce the final 60 tuples. Accordingly, we can analyze the correlation between $\mathcal{L}_s$ and $\mathcal{P}_r$ as well as between $\mathcal{L}_{s, t}$ and $\mathcal{P}_r$.

\cref{fig:spearman} plots the curves of $\mathcal{P}_r$ as $\mathcal{L}_s$ and $\mathcal{L}_{s, t}$ change, respectively, and provides the corresponding Spearman correlation coefficients. We can summarize several insightful observations.
\textbf{(1)} \cref{fig:XsTS5} and \cref{fig:XsTCD} show that as $\mathcal{L}_s$ increases, $\mathcal{P}_r$ gradually decreases, with the Spearman correlation coefficient as -0.9935 and -0.88793 in the target domains of \textbf{S} and \textbf{C} ($D_{4096}$), respectively. Both indicate a strong negative correlation between $\mathcal{L}_s$ and $\mathcal{P}_r$. Since a smaller $\mathcal{L}_s$ corresponds to higher discriminability of the source domain, improving the discriminability of the source domain is crucial to ensure the positive transfer from the source domain to the target domain.
\textbf{(2)} \cref{fig:DistributionTS5} and \cref{fig:DistributionTCD} illustrate that with an increase in $\mathcal{L}_{s, t}$, $\mathcal{P}_r$ decreases gradually. Also, the Spearman correlation coefficients between $\mathcal{L}_{s, t}$ and $\mathcal{P}_r$ in the target domains of \textbf{S} and \textbf{C} ($D_{4096}$) are -0.9935 and -0.87712, respectively. Those results imply that there is a strong negative correlation between $\mathcal{L}_{s, t}$ and $\mathcal{P}_r$. Since a lower $\mathcal{L}_{s, t}$ indicates stronger transferability of the source domain, it is necessary to enhance the transferability of the source domain to achieve the positive transfer.
\textbf{(3)} Based on all the above observations, we find that both the discriminability and transferability of the source domain strongly correlate with the transfer performance. Moreover, since the above experiments use randomly sampled source noise, it reveals an insightful observation: \textit{regardless of the origin of the domain of for source samples (e.g., image, text, noise), ensuring their discriminability and transferability in the common subspace can guarantee the transfer performance}. This also explains why utilizing source noise can achieve comparable performance to that of vanilla source samples.

\begin{figure*}[t]
\vspace{-1mm}
\centering
\subfloat[\small \textbf{N}$_6$ $\rightarrow$ \textbf{S}: $t = 1 $\label{fig:tsne_N2TS5_1}] {
{\includegraphics[width=0.48\columnwidth]{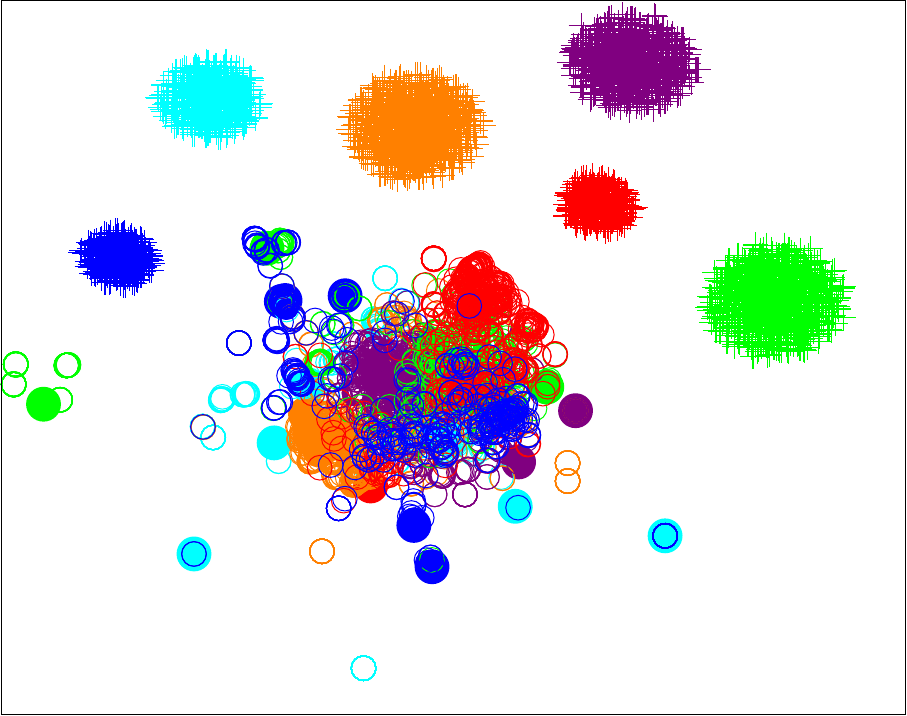}}
}
\subfloat[\small \textbf{N}$_6$ $\rightarrow$ \textbf{S}: $t = 200 $\label{fig:tsne_N2TS5_200}] {
{\includegraphics[width=0.48\columnwidth]{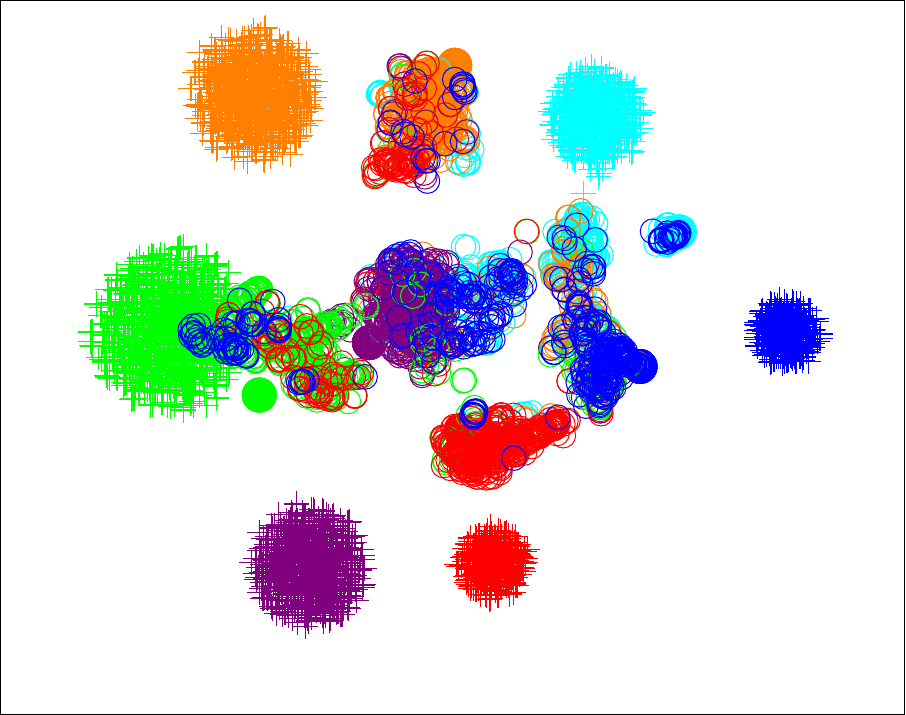}}
}
\subfloat[\small \textbf{N}$_6$ $\rightarrow$ \textbf{S}: $t = 400$ \label{fig:tsne_N2TS5_400}] {
{\includegraphics[width=0.48\columnwidth]{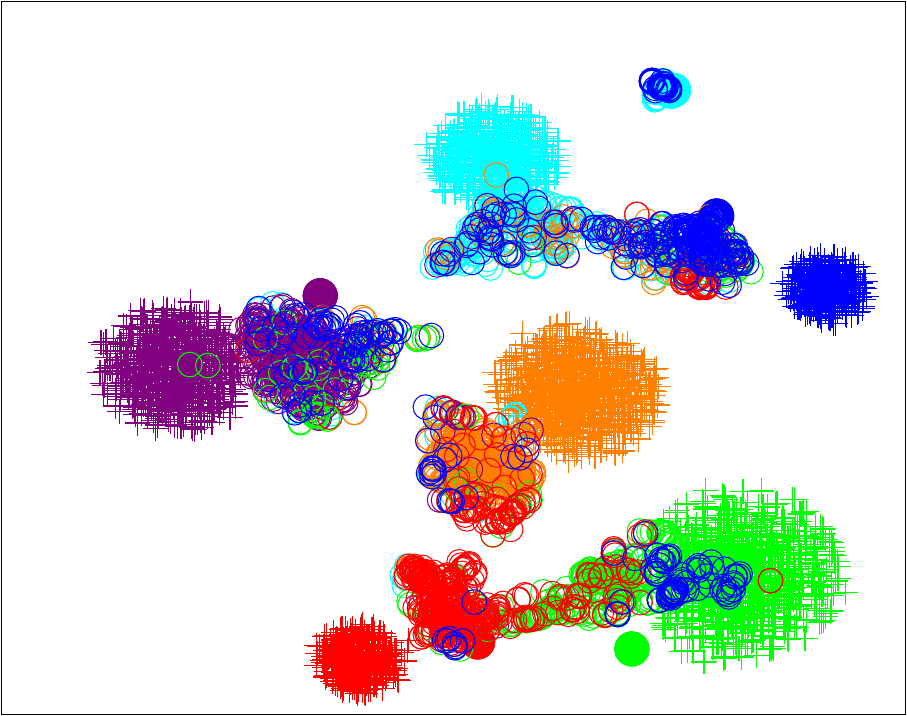}}
}
\subfloat[\small \textbf{N}$_6$ $\rightarrow$ \textbf{S}: $t = 600$\label{fig:tsne_N2TS5_600}] {
{\includegraphics[width=0.48\columnwidth]{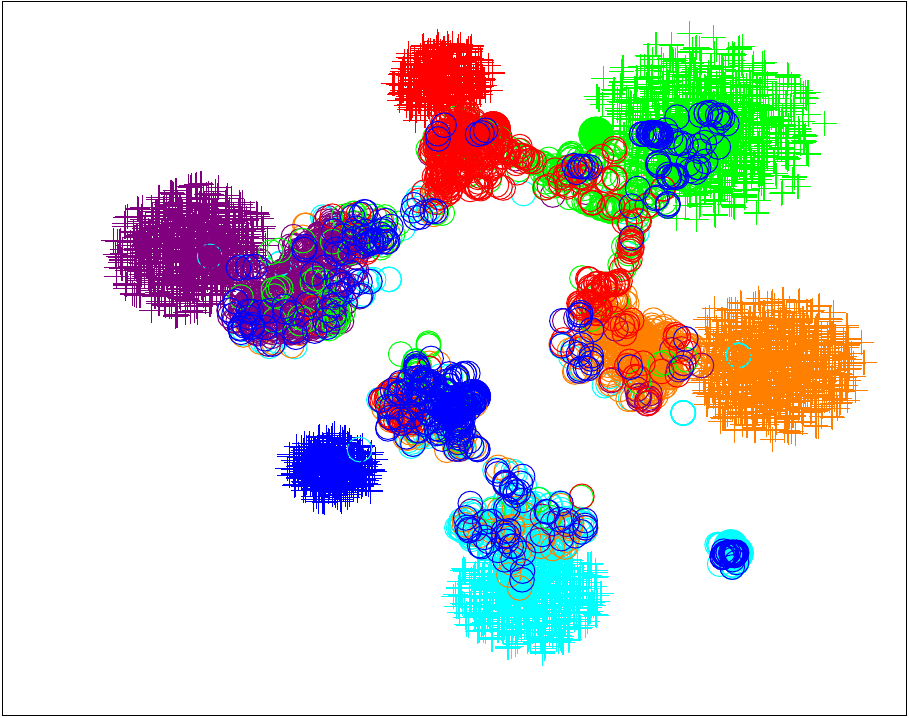}}
}
\\
\subfloat[\small \textbf{N}$_{10}$ $\rightarrow$ \textbf{C} ($D_{4096}$): $t = 1$ \label{fig:tsne_N2TCD_1}] {
{\includegraphics[width=0.48\columnwidth]{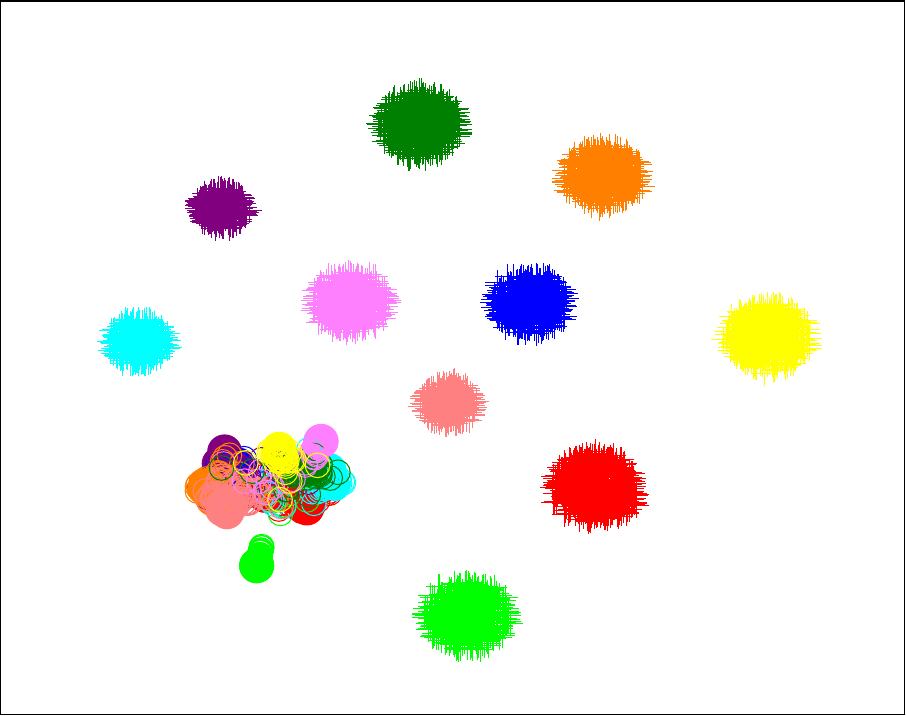}}
}
\subfloat[\small \textbf{N}$_{10}$ $\rightarrow$ \textbf{C} ($D_{4096}$): $t = 200$ \label{fig:tsne_N2TCD_200}] {
{\includegraphics[width=0.48\columnwidth]{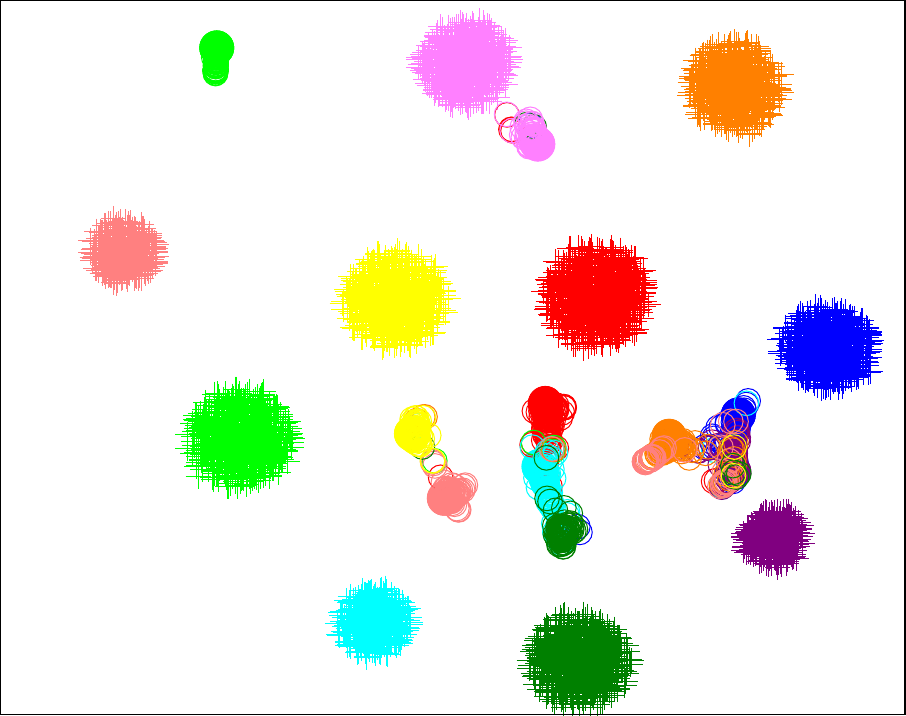}}
}
\subfloat[\small \textbf{N}$_{10}$ $\rightarrow$ \textbf{C} ($D_{4096}$): $t = 400$ \label{fig:tsne_N2TCD_400}] {
{\includegraphics[width=0.48\columnwidth]{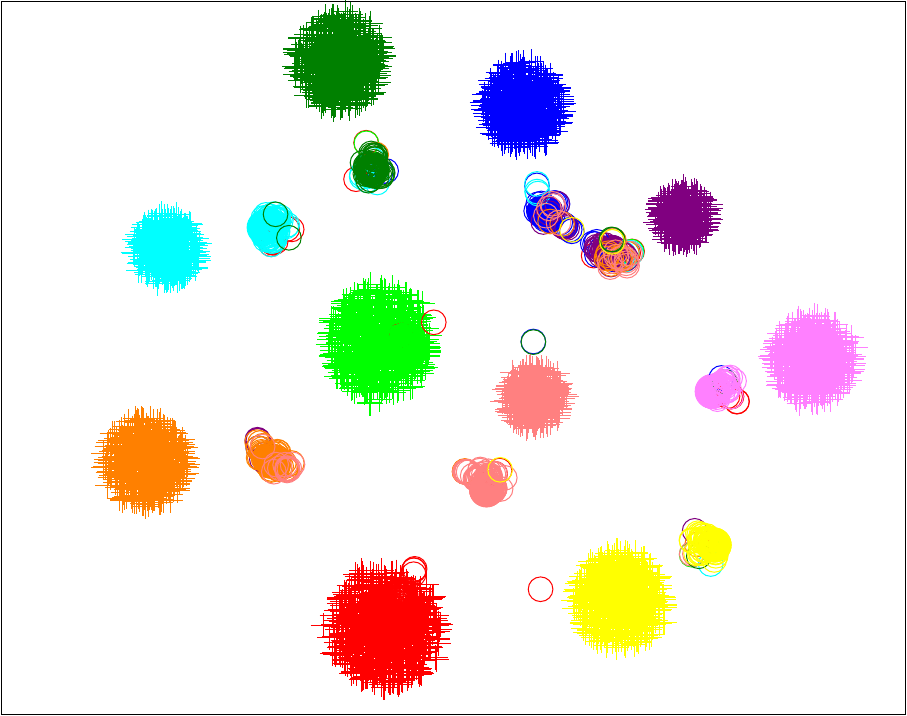}}
}
\subfloat[\small \textbf{N}$_{10}$ $\rightarrow$ \textbf{C} ($D_{4096}$): $t = 600$ \label{fig:tsne_N2TCD_600}] {
{\includegraphics[width=0.48\columnwidth]{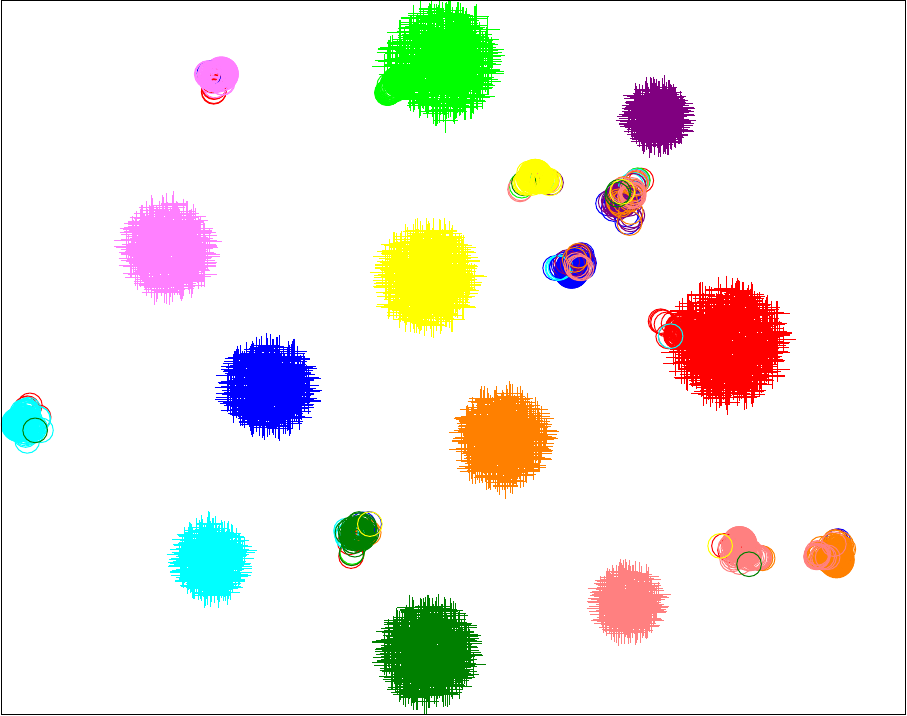}}
}
\caption{t-SNE visualization on the tasks of \textbf{N}$_6$ $\rightarrow$ \textbf{S} and \textbf{N}$_{10}$ $\rightarrow$ \textbf{C} ($D_{4096}$). Here, the `+' sign denotes a source sample, the `$\bullet$' sign represents a labeled target sample, and the `$\circ$' sign stands for an unlabeled target sample. Each color corresponds to a distinct category, and $t$ is the current number of iterations.}
\label{fig:tsneSHDA}
\vspace{-4ex}
\end{figure*}

\subsubsection{Analysis via Feature Visualization}

To intuitively understand why positive transfer occurs when the source domain exhibits good discriminability and transferability, we utilize the t-SNE technique \cite{Van2008Visualizing} to conduct feature visualization. Concretely, we first select two tasks that result in positive transfer, \textit{i.e.}, \textbf{N}$_6$ $\rightarrow$ \textbf{S} and \textbf{N}$_{10}$ $\rightarrow$ \textbf{C} ($D_{4096}$), and then visualize their transfer results when the number of iterations is set to 1, 200, 400, and 600, respectively. The visualization results are shown in \cref{fig:tsneSHDA}, which offers the following observations.
\textbf{(1)} Based on \cref{fig:tsne_N2TS5_1} and \cref{fig:tsne_N2TCD_1}, we can see that at the first iteration, source noise is well separable in the common subspace, which intuitively implies that source noise has good discriminability. Also, we find that unlabeled target samples exhibit substantial overlap and are not easy to distinguish. This is reasonable because the target projector and classifier are in the early stages of learning, leading to poor discriminability of the target domain.
\textbf{(2)} From Figs.~\ref{fig:tsne_N2TS5_200}-\ref{fig:tsne_N2TS5_600} and Figs.~\ref{fig:tsne_N2TCD_200}-\ref{fig:tsne_N2TCD_600}, we find that as the training iteration proceeds, source noise maintains high discriminability and target samples from different categories become separated progressively. Also, the distributions of both domains are gradually aligned. Those results indicate that due to the high discriminability of the source domain, it can be utilized as guidance information to enhance the discriminability of the target domain by aligning the distributions of both domains. In other words, \textit{as the transferability of the source domain continues to improve, its discriminability is gradually transferred to the target domain, thereby enhancing the discriminability of the target domain and resulting in positive transfer}.

\begin{figure*}[t]
\vspace{-2ex}
\centering
\subfloat[\small \textbf{N}$_6$ $\rightarrow$ \textbf{S}: t = 1 \label{fig:XXX}] {
{\includegraphics[width=0.48\columnwidth]{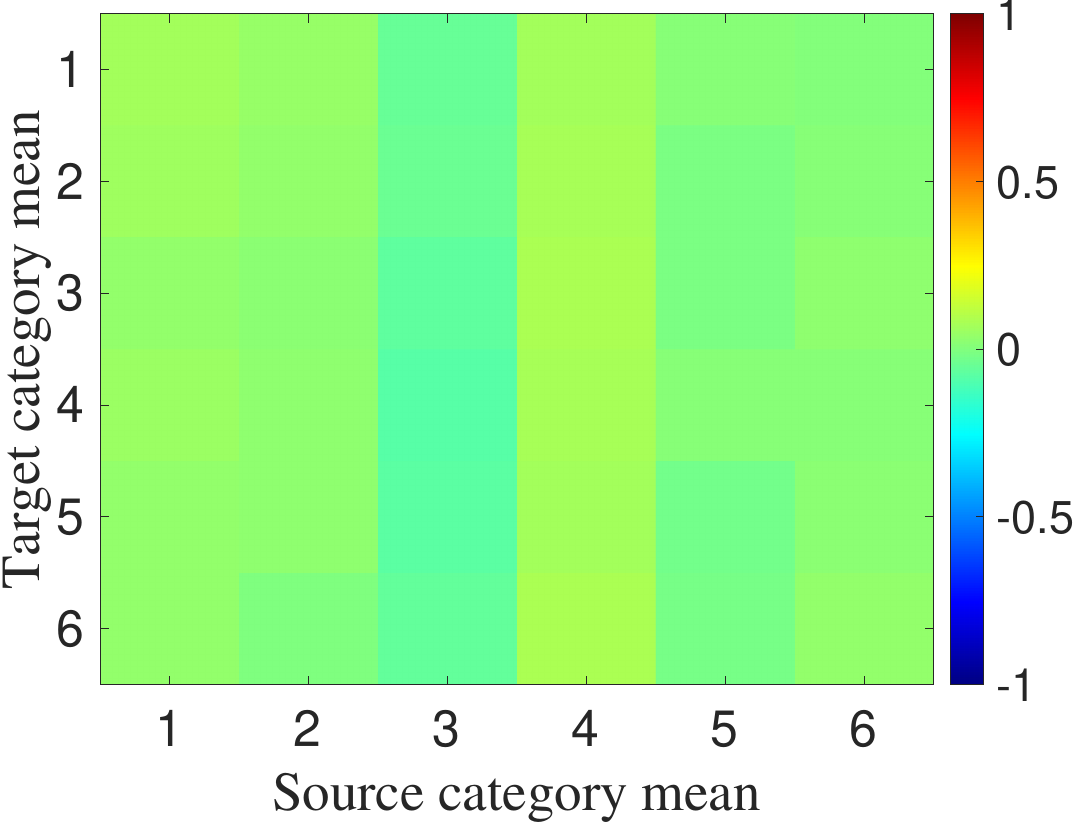}}
}
\subfloat[\small \textbf{N}$_6$ $\rightarrow$ \textbf{S}: t = 200 \label{fig:XXX}] {
{\includegraphics[width=0.48\columnwidth]{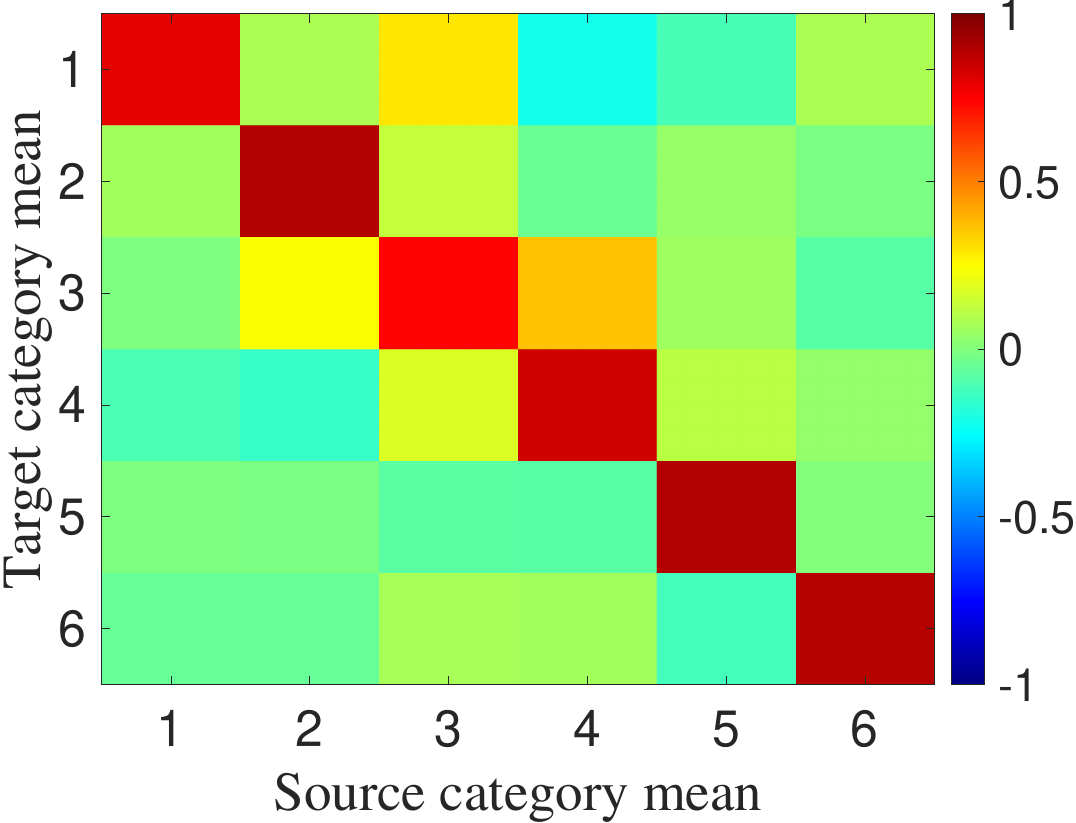}}
}
\subfloat[\small \textbf{N}$_6$ $\rightarrow$ \textbf{S}: t = 400 \label{fig:XXX}] {
{\includegraphics[width=0.48\columnwidth]{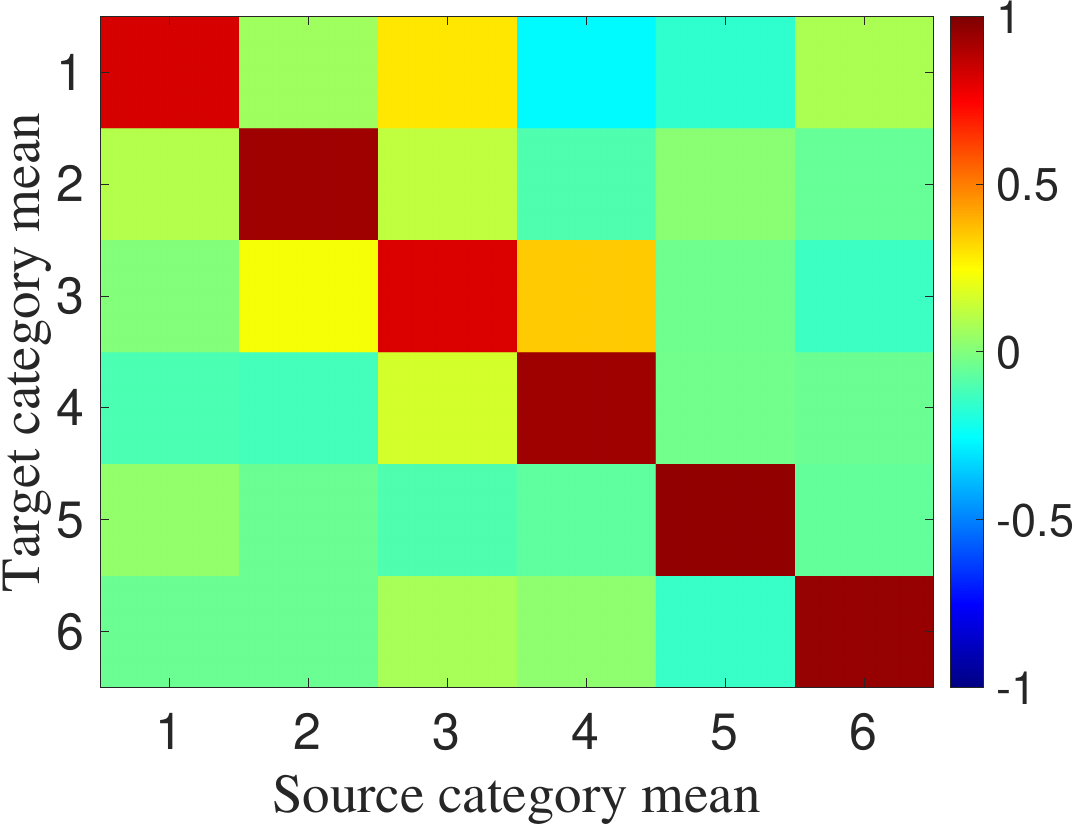}}
}
\subfloat[\small \textbf{N}$_6$ $\rightarrow$ \textbf{S}: t = 600 \label{fig:XXX}] {
{\includegraphics[width=0.48\columnwidth]{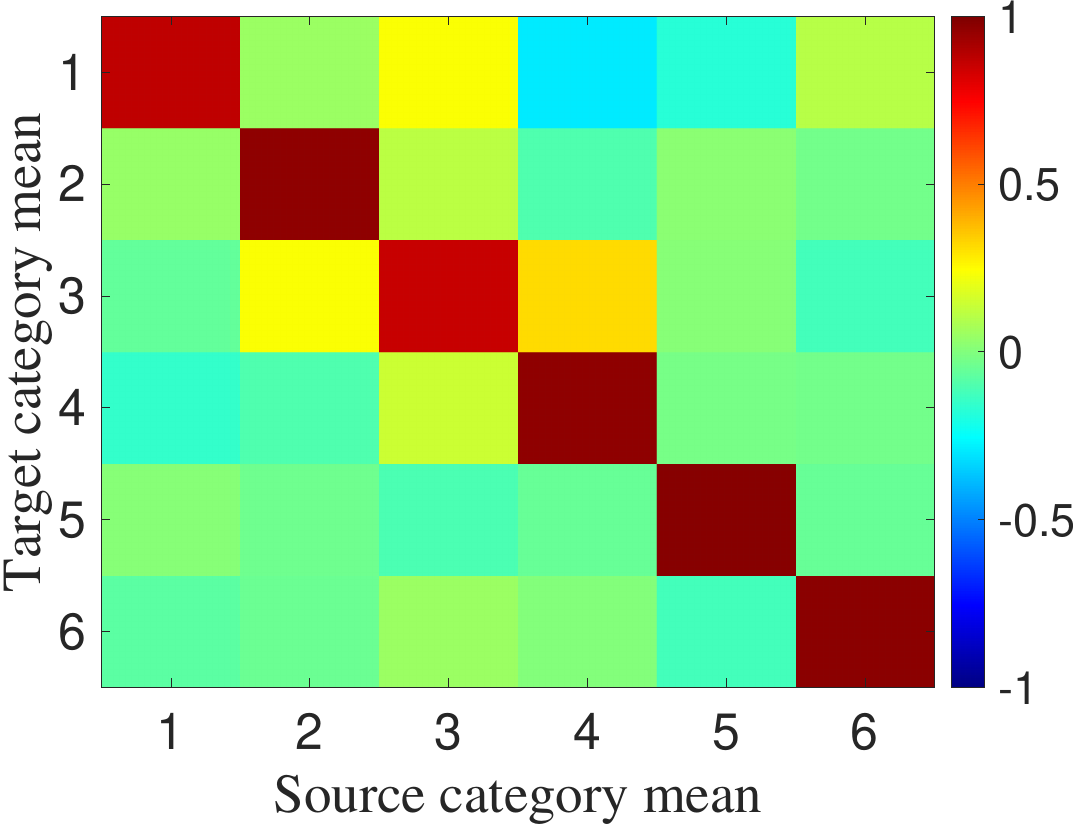}}
}
\hspace{-3.55mm}
\subfloat[\small \textbf{N}$_{10}$ $\rightarrow$ \textbf{C} ($D_{4096}$): t = 1 \label{fig:XXX}] {
{\includegraphics[width=0.48\columnwidth]{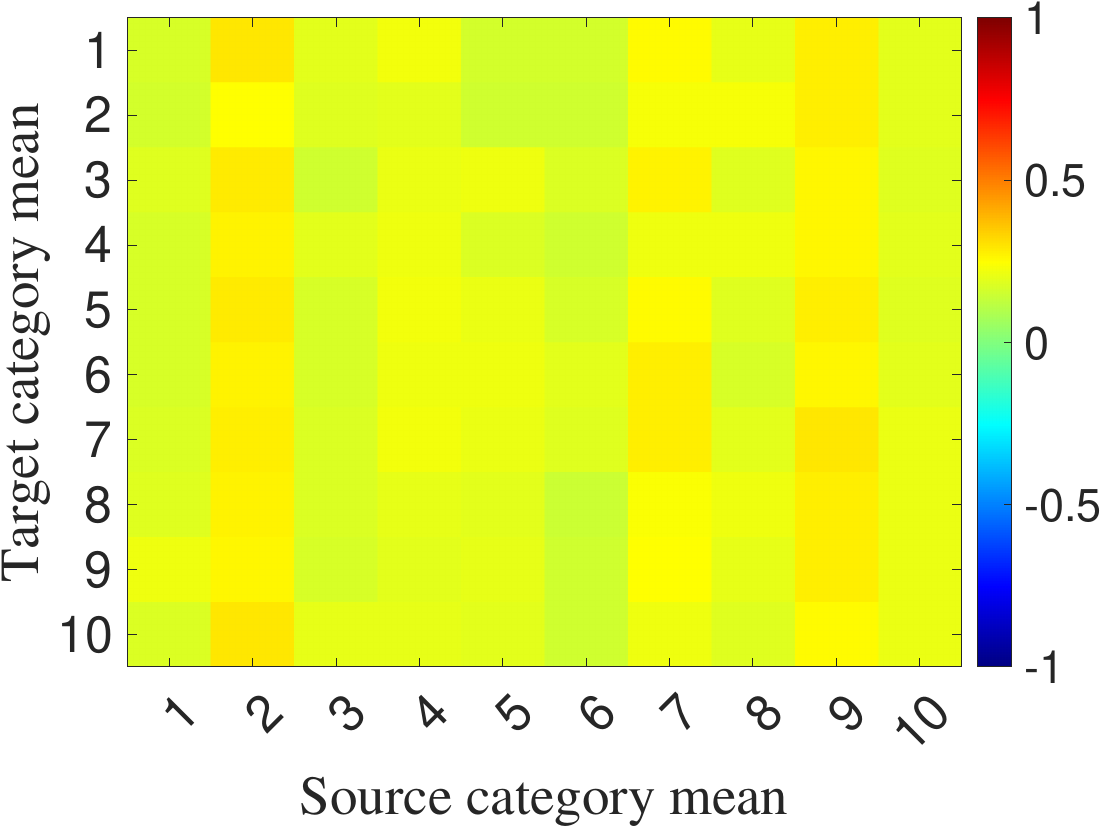}}
}
\subfloat[\small \textbf{N}$_{10}$ $\rightarrow$ \textbf{C} ($D_{4096}$): t = 200 \label{fig:XXX}] {
{\includegraphics[width=0.48\columnwidth]{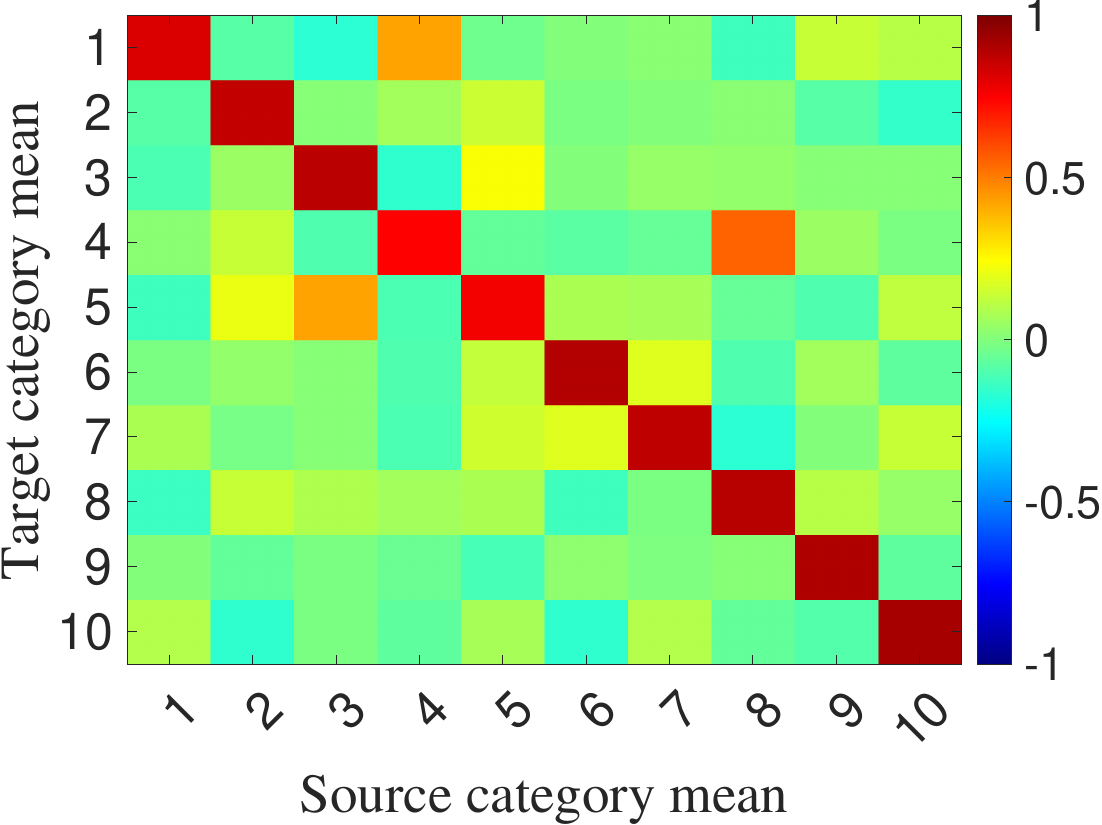}}
}
\subfloat[\small \textbf{N}$_{10}$ $\rightarrow$ \textbf{C} ($D_{4096}$): t = 400 \label{fig:XXX}] {
{\includegraphics[width=0.48\columnwidth]{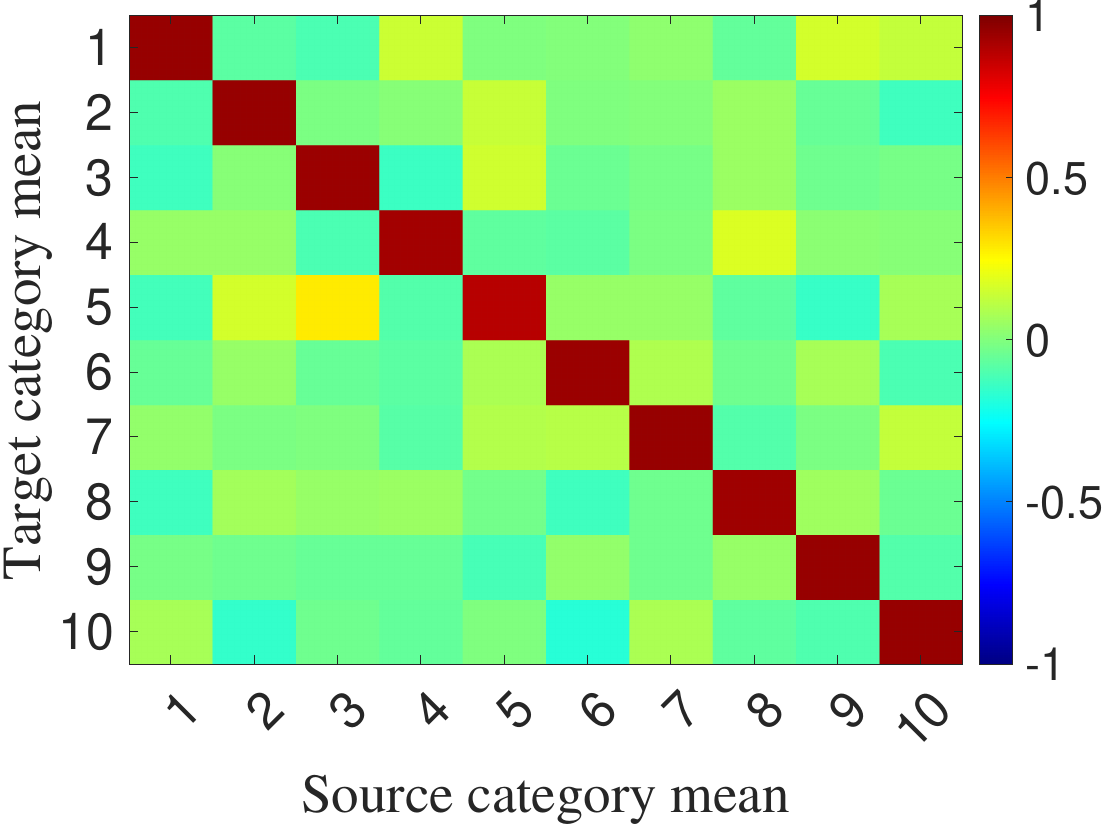}}
}
\subfloat[\small \textbf{N}$_{10}$ $\rightarrow$ \textbf{C} ($D_{4096}$): t = 600 \label{fig:XXX}] {
{\includegraphics[width=0.48\columnwidth]{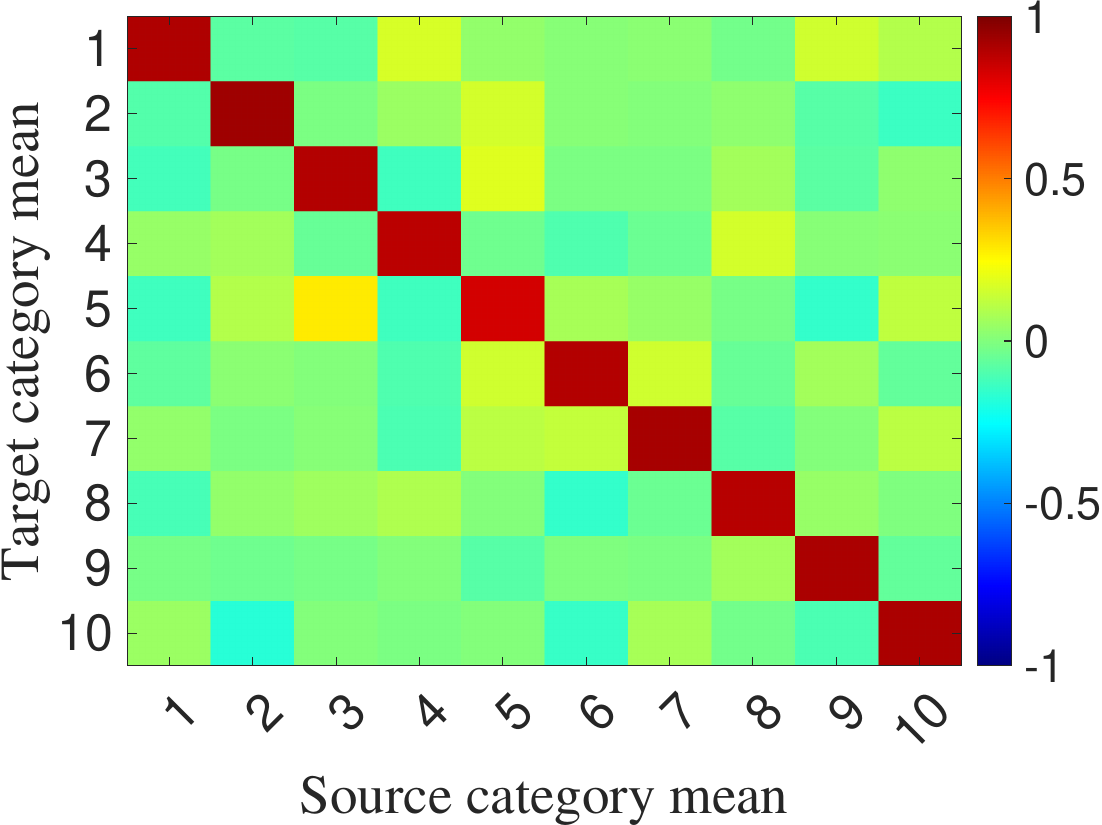}}
}
\caption{Cosine similarity between category means of source and target samples on the task of \textbf{N}$_6$ $\rightarrow$ \textbf{S} and \textbf{N}$_{10}$ $\rightarrow$ \textbf{C} ($D_{4096}$). Here, $t$ denotes the current number of iterations, and the x-axis and y-axis of each plot correspond to category means in the source and target domains, respectively.}
\label{fig:alignment}
\vspace{-3ex}
\end{figure*}

\subsubsection{Analysis via Alignment Visualization}

To gain a clear understanding of whether the discriminability of the source domain gradually transfers to the target domain as its transferability increases, we visualize the alignment process. To be specific, we utilize the cosine similarity to measure the alignment between category means of source and target samples. The cosine similarity outputs a score between $-1$ and $1$, with higher scores indicating better alignment. Specifically, the alignment score across category means of source and target samples is calculated as
\begin{align} \label{alpha_ck}
\alpha_{c, k} = \frac{\langle \mathbf{m}_t^c,  \mathbf{m}_s^k \rangle}{\| \mathbf{m}_t^c \| \| \mathbf{m}_s^k \|},
\end{align}
where $\mathbf{m}_s^k$ and $\mathbf{m}_t^c$ are defined in \eqref{m_s^c} and \eqref{m_t^c}, respectively. 
Note that we utilize the ground-truth labels of unlabeled samples to calculate $\mathbf{m}_t^c$, which can better reflect the discriminability of the target domain.

\cref{fig:alignment} visualizes the alignment processes on the tasks of \textbf{N}$_6$ $\rightarrow$ \textbf{S} and \textbf{N}$_{10}$ $\rightarrow$ \textbf{C} ($D_{4096}$), respectively. We can observe that as the number of iterations increases, the similarity between category means of source and target samples from the same category steadily improves. \textit{Those results indicate that the discriminability of the target domain is steadily approaching that of the source domain. Given the high discriminability of the source domain, the target domain inherits its discriminability to some extent, leading to positive transfer performance}. 
In a nutshell, those observations provide evidence that as the transferability of the source domain increases, the discriminability of the source domain is progressively transferred to the target one.

\subsection{Summary}

In summary, building on all the aforementioned experimental results, we can summarize them into an insightful observation.

\noindent\textbf{Observation 3:} \textit{The principal source of transferable knowledge in SHDA tasks lies in the transferability and discriminability of the source domain. Also, regardless of the domain from which source samples originate (e.g., image, text, noise), as the transferability of the source domain improves, its discriminability gradually transfers to the target domain, leading to positive transfer. Consequently, ensuring those properties in the source domain is crucial for achieving good transfer performance in SHDA}.

\section{Discussion}
\label{section:discussion}

In this section, we commence by conducting additional experiments to verify the influence of the feature projector on the performance of the target domain, aligning with the findings in \cref{subsection:categoryInformation}. Then, we discuss several studies closely related to our observations. Finally, we highlight the potential value of those observations.

\subsection{Additional Experiments on Category-permutated Homogeneous Transfer Tasks} 
\label{homogeneousExp}

\begin{figure*}[t]
\centering
\subfloat[\textbf{C$_s$} ($D_{4096}$) $\rightarrow$ \textbf{C$_t$} ($D_{4096}$) \label{fig:CDS2CDT}] 
{
{\includegraphics[width=0.5\columnwidth]{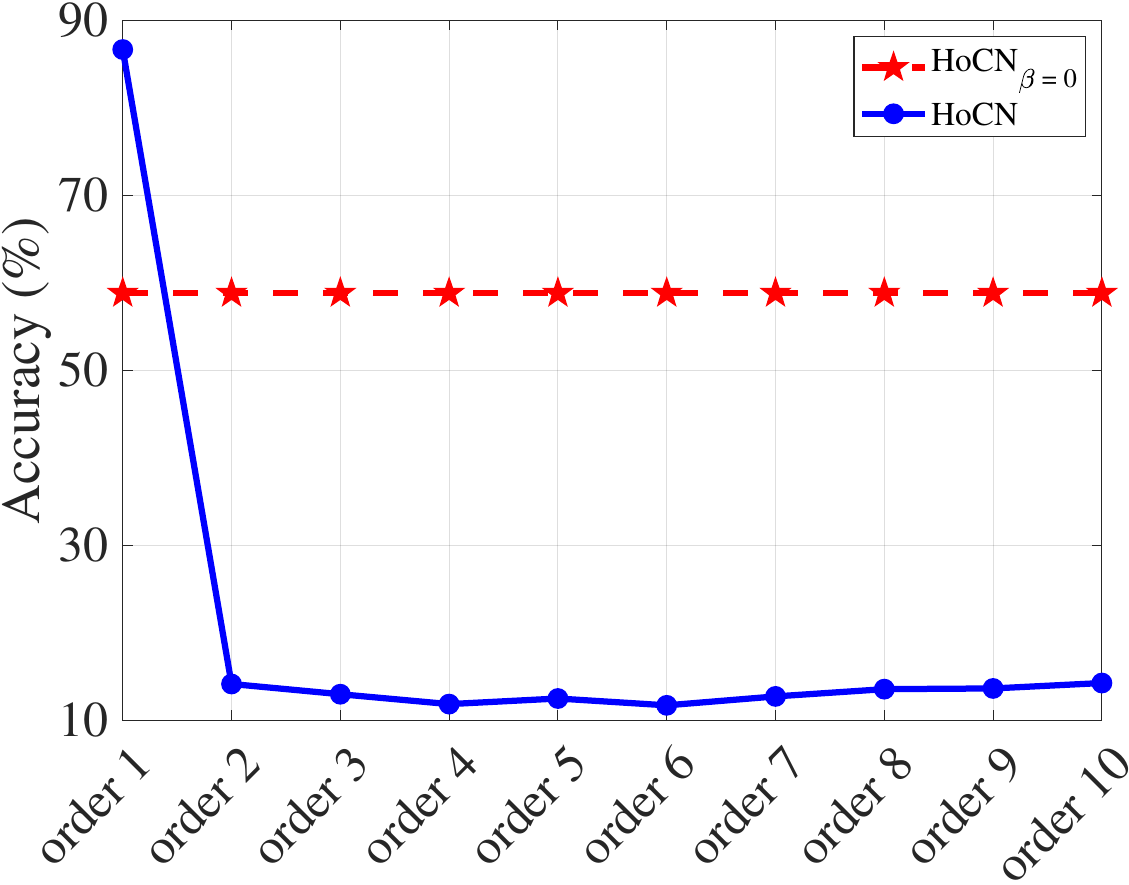}}
}
\hspace{-3.8mm}
\subfloat[\textbf{W$_s$} ($D_{4096}$) $\rightarrow$ \textbf{W$_t$} ($D_{4096}$) \label{fig:WDS2WDT}] {
{\includegraphics[width=0.5\columnwidth]{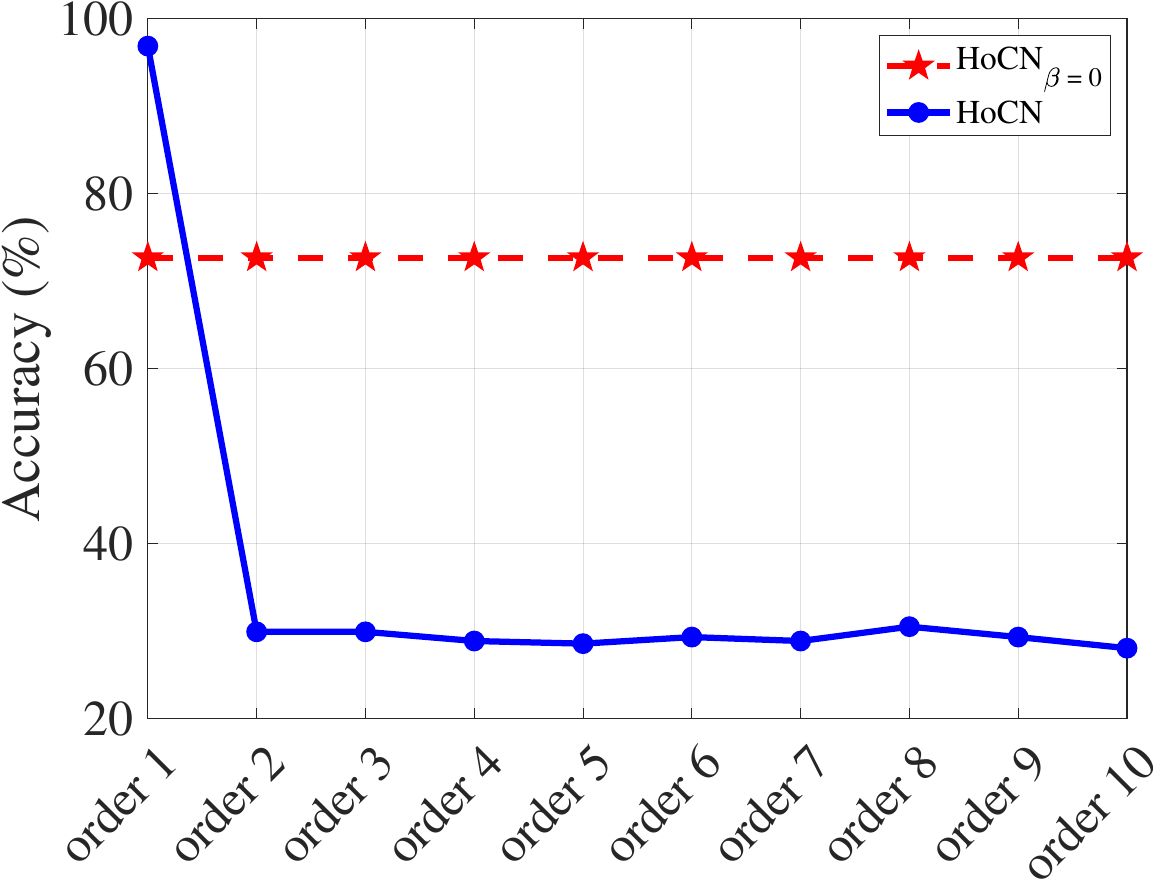}}
}
\hspace{-3.8mm}
\subfloat[\textbf{Image$_s$} $\rightarrow$ \textbf{Image$_t$} \label{fig:ImageS2ImageT}] {
{\includegraphics[width=0.5\columnwidth]{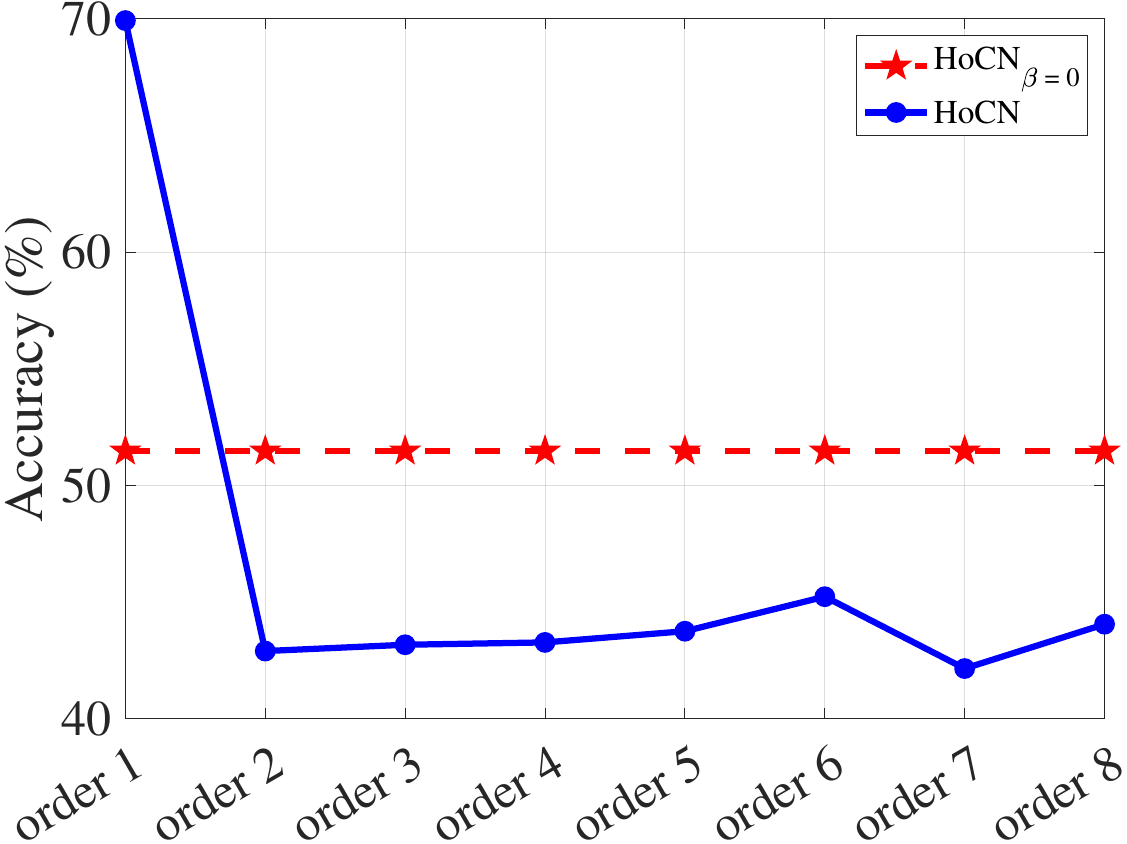}}
}
\hspace{-3.8mm}
\subfloat[\textbf{S$_s$} $\rightarrow$ \textbf{S$_t$} \label{fig:SS2ST}] {
{\includegraphics[width=0.5\columnwidth]{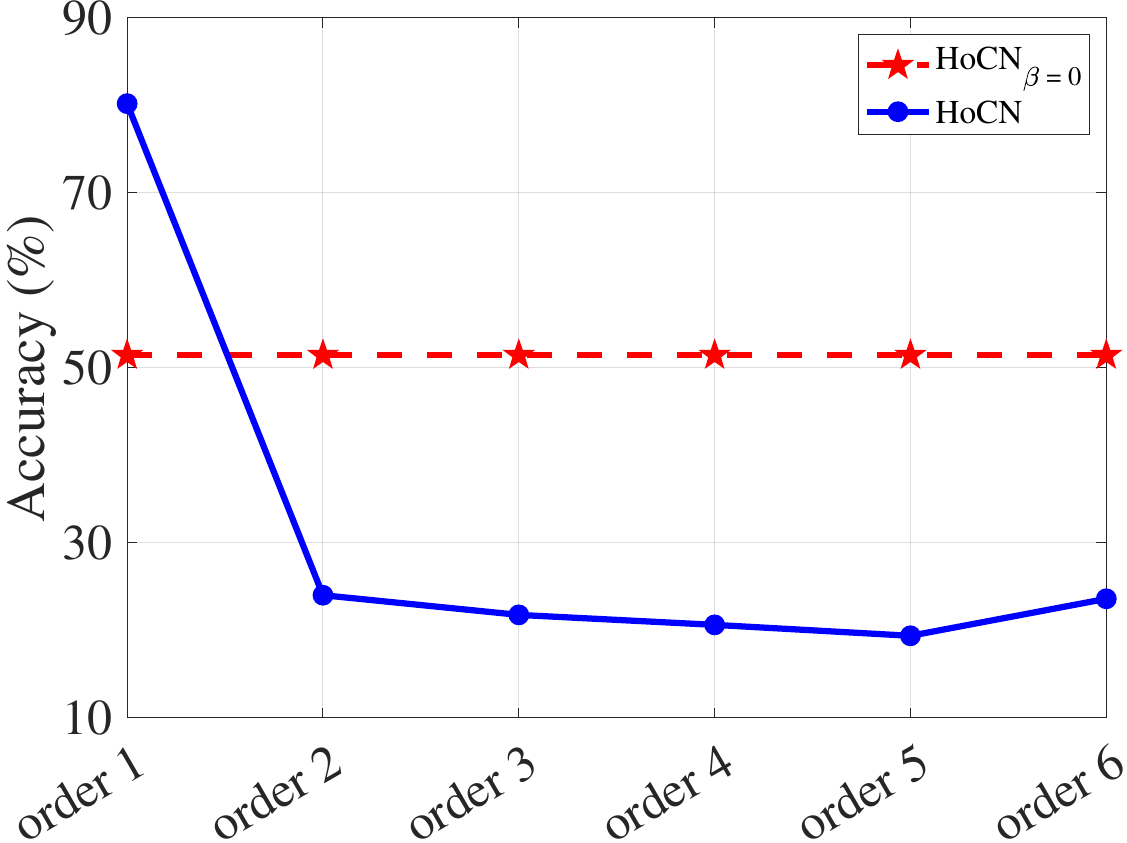}}
}
\caption{Classification accuracies (\%) with distinct orders of category indices for source samples on \textit{homogeneous} transfer tasks.}
\label{fig:Homogeneous}
\vspace{-4ex}
\end{figure*}

To assess the influence of the feature projector on the performance of the target domain, we build a series of \texttt{category-permutated homogeneous transfer tasks}. In particular, we first choose the domains of \textbf{C} ($D_{4096}$), \textbf{W} ($D_{4096}$), \textbf{Image}, and \textbf{S}. Then, we randomly and uniformly partition all samples in each domain into two parts: one for the source domain and the other for the target domain. For the source domain, we utilize all samples as labeled samples. As for the target domain, we randomly
select 1\% of the samples to be labeled, and the rest samples are considered as unlabeled ones. 
Consequently, we construct four groups of \textit{homogeneous} transfer directions: \textbf{C$_s$} ($D_{4096}$) $\rightarrow$ \textbf{C$_t$} ($D_{4096}$), \textbf{W$_s$} ($D_{4096}$) $\rightarrow$ \textbf{W$_t$} ($D_{4096}$), \textbf{Image$_s$} $\rightarrow$ \textbf{Image$_t$}, and \textbf{S$_s$} $\rightarrow$ \textbf{S$_t$}, where the subscripts (\textit{i.e.}, $s$ and $t$) denote the source and target domains, respectively.
Following the category-permutated setting detailed in \cref{subsection:categoryInformation}, we create 10 transfer tasks for the first two groups, eight for the third group, and six for the last group, based on the number of categories in each. Also, the ground-truth order is designated as order 1, while the remaining orders are permutated, leading to changes in the category information (please refer to \cref{fig:allOrder} for details).
Accordingly, we establish a total of 34 \textit{homogeneous} transfer tasks.

In addition, we develop a \textit{Homogeneous} Classification Network (HoCN) to evaluate the performance of the target domain. Concretely, HoCN projects labeled samples from both domains into a common subspace by training a \textit{domain-shared feature projector} and classifier. Thus, we formulate the objective function of HoCN as
\begin{align}  \label{HoCN}
\min_{f, g} \frac{1}{n_l} \sum_{i = 1}^{n_l} & \mathcal{L}_{ce} \big[ \mathbf{y}_i^l, f (g (\mathbf{x}_i^l)) \big]
+  \frac{\beta}{n_s} \sum_{i = 1}^{n_s} \mathcal{L}_{ce} \big[ \mathbf{y}_i^s, f (g (\mathbf{x}_i^s)) \big] \nonumber
\\ &
+ \tau \big( \left\| g \right\|^2 + \left\| f \right\|^2 \big),
\end{align}
where $g(\cdot)$ stands for a single-layer fully connected network with the Leaky ReLU activation function \cite{Maas2013Rectifier}, while $\beta$ and $\tau$ are two trade-off parameters empirically set to 0.01 and 0.005, respectively. Note that when $\beta$ is set to zero, the problem in \eqref{HoCN} degenerates into a supervised learning problem that only utilizes labeled target samples for training. We denote the optimal model for this problem as HoCN$_{\beta=0}$.

\cref{fig:Homogeneous} shows the accuracies of HoCN and HoCN$_{\beta = 0}$ \textit{w.r.t.} different orders of category indices for source samples on all the above tasks. We can summarize several insightful observations. \textbf{(1)} When the category index of source samples is the ground-truth order, \textit{i.e.}, order 1, HoCN significantly outperforms HoCN$_\beta = 0$ on all the tasks. This is reasonable because source and target samples originate from the same domain, and HoCN uses more labeled samples than HoCN$_\beta = 0$. \textbf{(2)} When the category indices of source samples do not follow the ground-truth order, HoCN yields extremely poor performance. This implies that the order of category indices for source samples is crucial in scenarios where both source and target samples share a feature projector. \textit{One important reason is that it is challenging to classify source and target samples, belonging to the same category, into different categories using a shared feature projector. This disrupts the learning of the feature projector, leading to poor performance}. Overall, all the observations provide evidence that the \textit{heterogeneity} of the source and target feature projectors is the primary cause of the phenomenon observed in \cref{fig:Categoryorder}.

\subsection{Comparison with Related Studies}

In the experiments presented in this paper, a pivotal observation is that noise may contain transferable knowledge under the SHDA setting, which seems a bit counter-intuitive. In reality, however, several studies \cite{Baradad2021Learning,Tang2022Virtual,Luo2021No} have paid attention to the value of noise for tackling distinct machine learning tasks. For instance, Baradad \textit{et al.} \cite{Baradad2021Learning} utilize noise to deal with the representation learning problem. Specifically, they pre-train a visual representation learner with a contrastive loss using noise generated from simple distributions, such as randomly initialized deep generative models. Their experiments demonstrate that the noise effectively enhances the representation ability of the visual representation learner. Another example is that Luo \textit{et al.} \cite{Luo2021No} adopt noise to handle the non-independently and identically distributed (non-i.i.d.) problem in federated learning. Specifically, they first estimate the global mean and covariance information for each category. Then, based on such information, they sample noise from a Gaussian mixture distribution to fine-tune the classifier on the server. Their experiments reveal that the noise substantially improves the classification performance. Similar to \cite{Luo2021No}, Tang \textit{et al.} \cite{Tang2022Virtual} also apply noise to tackle the non-i.i.d. issue in federated learning. In particular, they first upsample pure Gaussian noise and then align the distributions of noise and vanilla samples in each client. Their experimental results verify that federated learning could significantly benefit from the noise. Overall, those studies indicate that noise can be beneficial for several machine learning tasks, which aligns with our observation to a certain extent. Different from the above studies, we conduct comprehensive analytical experiments to delve deeper into the reason behind the effectiveness of noise for SHDA.

\subsection{Potential Value in Practical Applications}

\begin{figure}[t]
\centering
{\includegraphics[width=0.98\columnwidth]{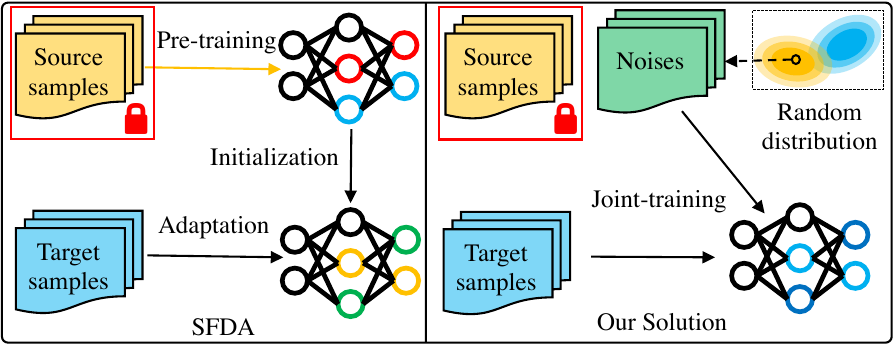}}
\caption{A comparison of the SFDA and our solution. To escape from the dilemma of unavailable access to source samples, SFDA methods rely on using a well-trained source model, whereas our solution merely requires sampling noise from a random distribution as a substitute for source samples.}
\label{fig:comparison}
\vspace{-4ex}
\end{figure}

Vanilla DA methods \cite{Pan2010A,Csurka-2017A,Zhuang2020A,Day2017A} assume that source samples are publicly available. However, in many practical applications, it is often not easy to acquire those samples due to privacy, confidentiality, and copyright issues. 
To escape from this dilemma, a potential solution, \textit{i.e.}, \textit{source-free} domain adaptation (SFDA) (see the left in \cref{fig:comparison}) \cite{Li2024A,Liang2020Do,Jing2024Visually,Luo2023Source}, has been proposed in recent years. As a rule, SFDA methods utilize a well-trained source model to initialize a target model and then adapt it using unlabeled target samples.
While source samples are not publicly accessible under the SFDA setting, the source model trained on those source samples remains necessary. However, in several practical applications with strict privacy requirements, it may be challenging to ascertain the relationship between source and target samples based solely on a public source model. This challenge hinders the further development of SFDA as we face an issue to determine which well-trained source models to be utilized for the target task. However, unlike the SFDA, our observations offer another promising solution (see the right in \cref{fig:comparison}).
It neither requires access to source samples nor a well-trained source model. Instead, 
it directly samples noise from a random distribution as source samples and then performs domain adaptation in a semi-supervised fashion.
Accordingly, it eliminates both the dependence on publicly available source samples and models.
As a result, we believe that our observations provide a new perspective to address the aforementioned dilemma, thus holding significance for various practical applications.

\section{Conclusion}
\label{section:conclusion}

This paper conducts an in-depth empirical study to investigate the transferable knowledge in SHDA. 
First, we find that the category and feature information of source samples are not the primary factors affecting the performance of the target domain.
Then, we observe that noise sampled from several simple distributions as source samples contributes to effective knowledge transfer. 
Next, we perform a series of experiments to analyze the transferable knowledge in SHDA by constructing various noise domains. 
Building on extensive experimental results, we observe that both the transferability and discriminability of the source domain are strongly correlated with the performance improvement ratio in the target domain. Accordingly, we hold an opinion that the transferability and discriminability of the source domain are the dominant factors of the transferable knowledge in SHDA. Therefore, it is vital to ensure those properties in the source domain to achieve effective knowledge transfer. One promising direction for future work is to establish theoretical foundations that support those observations.

\bibliography{SHDA}
\bibliographystyle{IEEEtran}

\vspace{-4ex}

\begin{IEEEbiography}[{\includegraphics[width=1in,height=1.25in,clip,keepaspectratio]{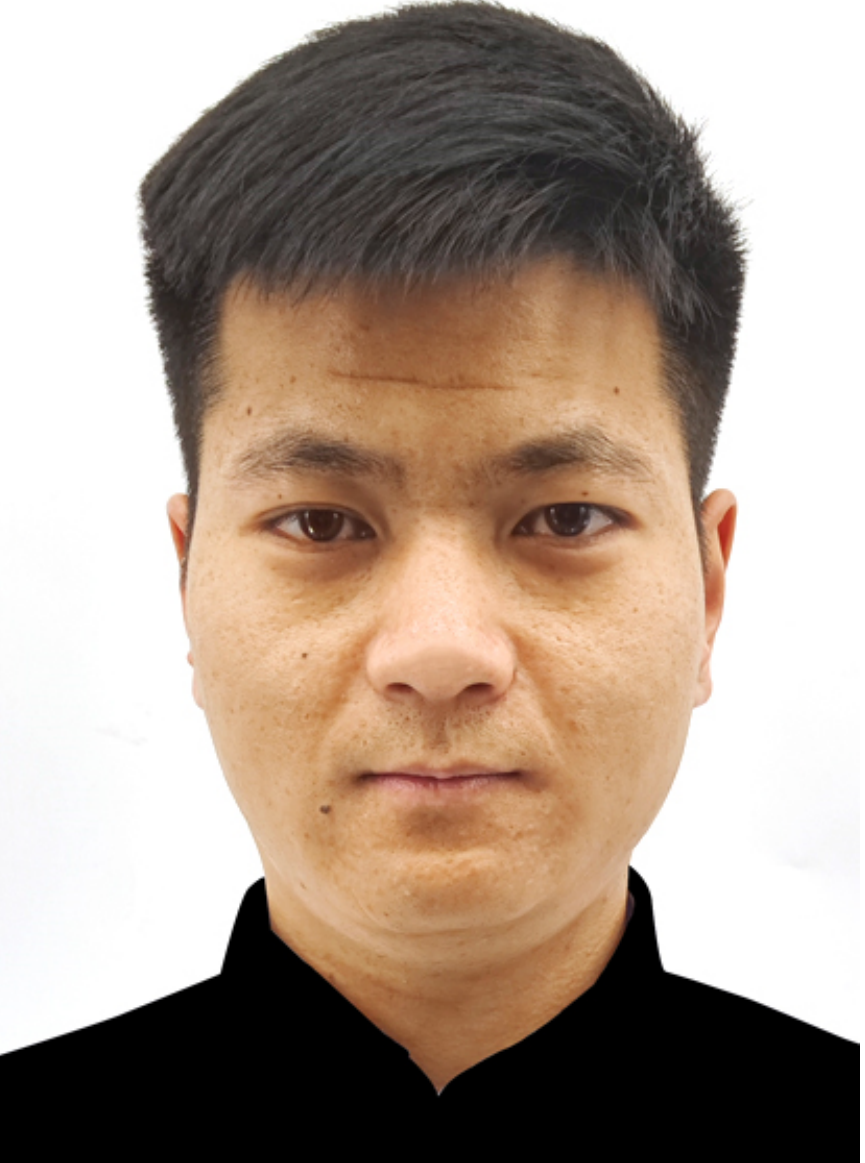}}]{Yuan Yao} received the Ph.D. degree from the Department of Computer Science and Technology, Harbin Institute of Technology, Shenzhen, China, in 2021. He previously served as a senior research and development engineer at Baidu, China. He is currently working as a researcher with the Beijing Teleinfo Technology Company Ltd., China Academy of Information and Communications Technology. His research interests include transfer learning, federated learning, and artificial intelligent watermarking. 
\end{IEEEbiography}

\vspace{-4ex}

\begin{IEEEbiography}[{\includegraphics[width=1in,height=1.25in,clip,keepaspectratio]{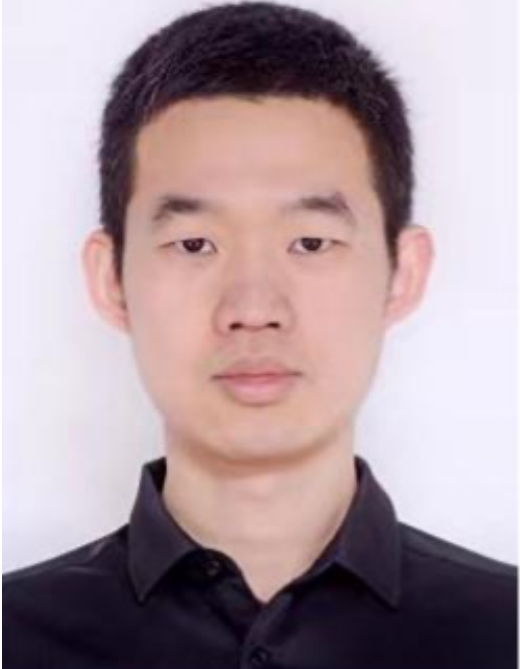}}]{Xiaopu Zhang} received the M.Eng. and Ph.D. degrees in instrumentation science and technology from Jilin University (JLU), China, in 2016 and 2019, respectively. He is currently a research fellow with Inspur Computer Technology Co., Ltd., Beijing,
China. Prior to join Inspur, he was a post-doctoral fellow with the School of Automation, Beijing Institute of Technology, Beijing, China. His research interests include machine learning and signal processing
\end{IEEEbiography}

\vspace{-4ex}

\begin{IEEEbiography}[{\includegraphics[width=1in,height=1.25in,clip,keepaspectratio]{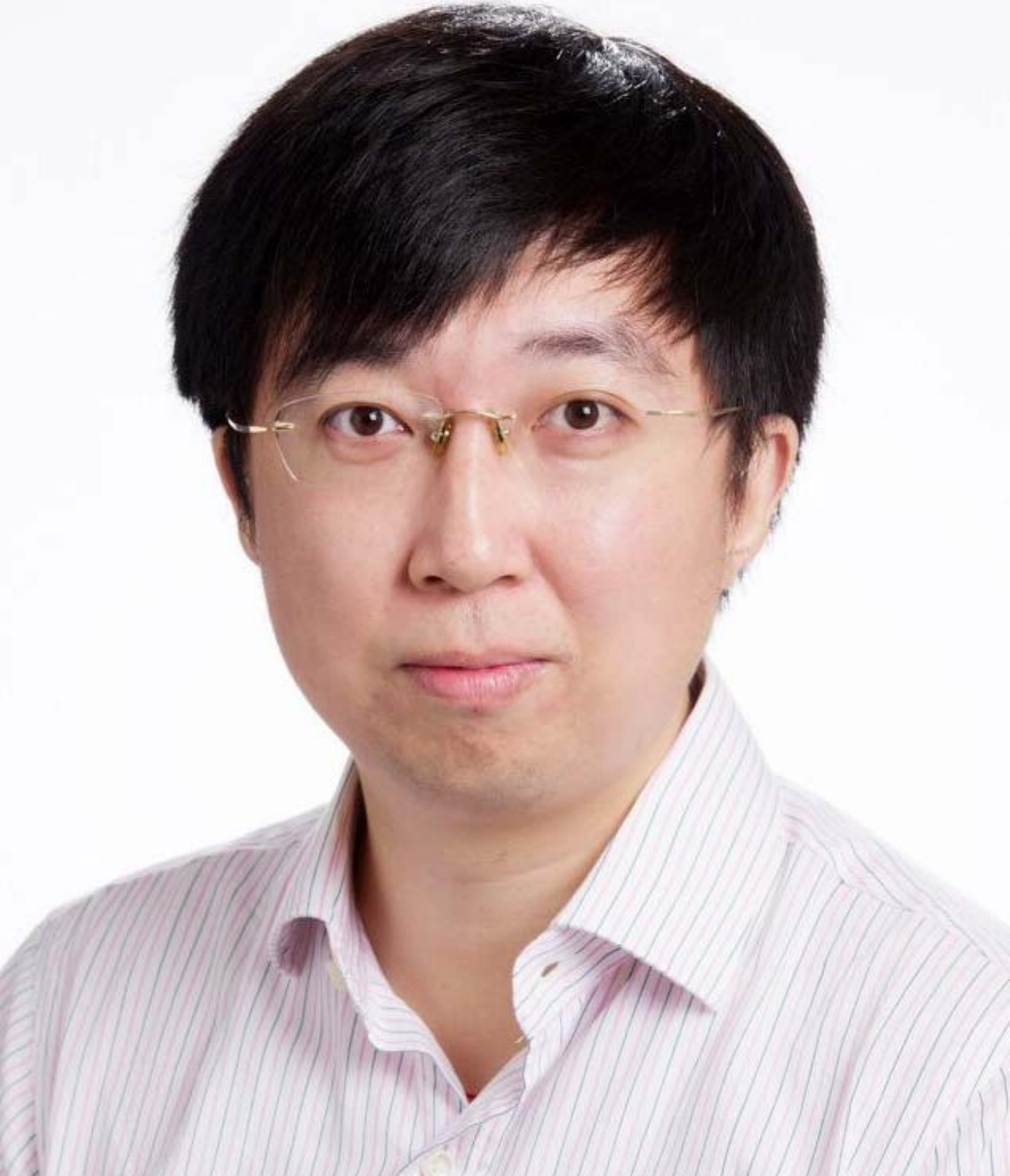}}]{Yu Zhang} (Member, IEEE) is an associate professor
with the Department of Computer Science and Engineering, Southern University of Science and Technology. His research interests mainly include artificial intelligence and machine learning, especially in multi-task learning, transfer learning, dimensionality reduction, metric learning, and semi-supervised learning. He has published a book Transfer Learning and about 80 papers on top-tier conferences and journals. He serves as a reviewer for various journals and area chairs/(senior) program committee members for several top-tier conferences. He has won the best article awards in UAI 2010 and PAKDD 2019, and the best student article award in WI 2013. 
\end{IEEEbiography}

\vfill\break

\begin{IEEEbiography}[{\includegraphics[width=1in,height=1.25in,clip,keepaspectratio]{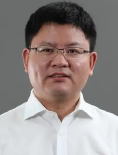}}]{Jian Jin} received the B.E. degree from Beijing Jiaotong University, Beijing, China, in 1999, and the Ph.D. degree from the University of Chinese Academy of Sciences, Beijing, in 2019. He is currently the Director of the Institute of Industrial Internet and Internet of Things, China Academy of Information and Communications Technology. His research interests include domain name, industrial Internet identity, and blockchain. 
\end{IEEEbiography}

\vspace{-3ex}

\begin{IEEEbiography}[{\includegraphics[width=1in,height=1.25in,clip,keepaspectratio]{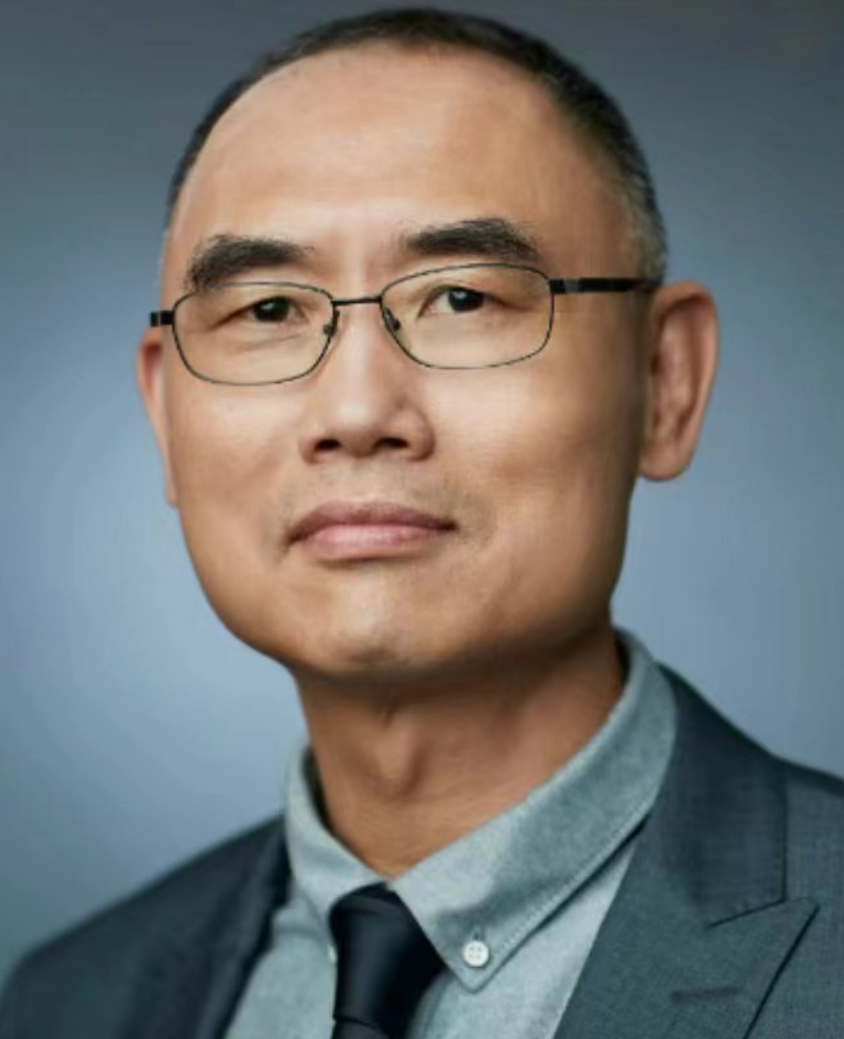}}]{Qiang Yang} (Fellow, IEEE) is a fellow of Canadian Academy of Engineering (CAE) and Royal Society of Canada (RSC), Chief Artiﬁcial Intelligence Ofﬁcer of WeBank, a chair professor of Computer Science and Engineering Department, Hong Kong University of Science and Technology (HKUST). He is the conference chair of AAAI-21, the honorary vice president of Chinese Association for Artiﬁcial Intelligence(CAAI), the president of Hong Kong Society of Artiﬁcial Intelligence and Robotics (HKSAIR) and the president of Investment Technology League (ITL). He is a fellow of AAAI, ACM, CAAI, IEEE, IAPR, AAAS. He was the founding editor in chief of the ACM Transactions on Intelligent Systems and Technology (ACM TIST) and the founding editor in chief of IEEE Transactions on Big Data (IEEE TBD). He received the ACM SIGKDD Distinguished Service Award, in 2017. He had been the founding director of the Huawei’s Noah’s Ark Research Lab between 2012 and 2015, the founding director of HKUST’s Big Data Institute, the founder of 4Paradigm and the president of IJCAI (2017-2019). His research interests are transfer learning, federated learning, artificial intelligence, machine learning, data mining and planning. 
\end{IEEEbiography}

\vfill

\end{document}